\documentclass[iicol, sn-mathphys-num]{sn-jnl}%

\usepackage{graphicx}%
\usepackage{multirow}%
\usepackage{amsmath,amssymb,amsfonts}%
\usepackage{amsthm}%
\usepackage{mathrsfs}%
\usepackage[title]{appendix}%
\usepackage{xcolor}%
\usepackage{textcomp}%
\usepackage{manyfoot}%
\usepackage{booktabs}%
\usepackage{algorithm}%
\usepackage{algorithmicx}%
\usepackage{algpseudocode}%
\usepackage{listings}%
\usepackage{placeins}
\usepackage{booktabs}\let\cline\cmidrule
\usepackage[capitalize]{cleveref}
\crefname{section}{Sec.}{Secs.}
\Crefname{section}{Section}{Sections}
\Crefname{table}{Table}{Tables}
\crefname{table}{Tab.}{Tabs.}

\theoremstyle{thmstyleone}%

\theoremstyle{thmstyletwo}%

\theoremstyle{thmstylethree}%

\begin{document}

\title[Article Title]{Assessing the Role of Datasets in the Generalization of Motion Deblurring Methods to Real Images}

\author*[1]{\fnm{Guillermo} \sur{Carbajal}}\email{carbajal@fing.edu.uy}

\author[2]{\fnm{Patricia} \sur{Vitoria}}%

\author[1]{\fnm{José} \sur{Lezama}}%

\author[1]{\fnm{Pablo} \sur{Musé}}

\affil*[1]{\orgdiv{IIE, Facultad de Ingeniería}, \orgname{Universidad de la República}, \orgaddress{\street{Herrera y Reissig 565}, \city{Montevideo}, \postcode{11500}, \country{Uruguay}}}

\affil[2]{\orgdiv{Dept. of Information \& Communication Technologies}, \orgname{Universitat Pompeu Fabra}, \orgaddress{\street{Street}, \city{Barcelona}, \postcode{08018}, \country{Spain}}}

\abstract{Successfully training end-to-end deep networks for real motion deblurring requires datasets of sharp/blurred image pairs that are realistic and diverse enough to achieve generalization to real blurred images. Obtaining such datasets remains a challenging task. In this paper, we first review the limitations of existing deblurring benchmark datasets and analyze the underlying causes for deblurring networks' lack of generalization to blurry images in the wild.  Based on this analysis, we propose an efficient procedural methodology to generate sharp/blurred image pairs based on a simple yet effective model. This allows for generating virtually unlimited diverse training pairs mimicking realistic blur properties. We demonstrate the effectiveness of the proposed dataset by training existing deblurring architectures on the simulated pairs and performing cross-dataset evaluation on three standard datasets of real blurred images. When training with the proposed method, we observed superior generalization performance for the ultimate task of deblurring real motion-blurred photos of dynamic scenes.}


\maketitle

\section{Introduction}\label{sec1}

Motion deblurring, a fundamental task in computer vision, has seen significant advancements in recent years. However, a critical challenge facing this field is the lack of robust generalization of current methods to real-world scenarios \citep{m_Tran-etal-CVPR21}. The primary issue is that modern motion deblurring techniques often fail to perform effectively beyond the specific datasets they are trained on, undermining their practical applicability. This issue, known as \textit{kernel overfitting} \cite{m_Tran-etal-CVPR21}, is particularly relevant when testing with real images. Since generating real blurry/sharp pairs is extremely challenging, deblurring networks are usually trained on synthetic training sets. Unfortunately, there is no correlation between the performance of state-of-the-art deblurring networks (MIMO-UNet+ \cite{cho2021rethinking}, MPRNet \cite{zamir2021multi}, NAFNet \cite{chen2022simple}) in the dataset they are trained on (GoPro dataset \cite{nah2017deep}) and the performance on real images (K\"{o}hler \cite{kohler2012recording}, Lai \cite{lai2016comparative}, and RealBlur \cite{rim_2020_ECCV} datasets). 

\begin{figure}[ht]
    \centering
    \includegraphics[width=0.5\textwidth]{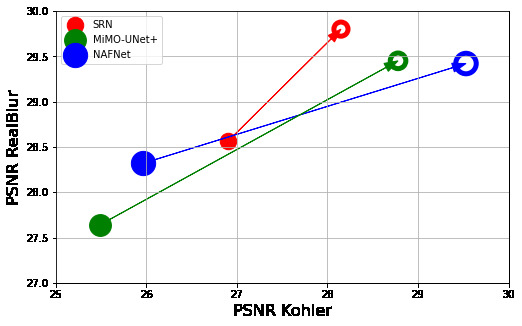}
    \caption{PSNR performance in the widely used GoPro dataset \cite{Nah_2017_CVPR} (represented by circles diameters) is not indicative of performance in real blurred photos (K\"{o}hler and RealBlur datasets).  When state-of-the art (MIMO-UNet and NAFNet) and classic (SRN) deblurring networks are trained using the GoPro dataset sharp images, but synthesizing the blurry images with the proposed procedure, the methods generalize better (filled circles to hollow circles transitions). }
    \label{fig:state_of_the_art}
\end{figure}

\begin{figure*}[ht!]
  \centering
  \setlength{\tabcolsep}{1pt}
  \small

  \begin{tabular}{p{0.4cm}*{10}{c}}%

& \multicolumn{2}{c}{Blurry} & \multicolumn{2}{c}{SRN-GoPro} &  \multicolumn{2}{c}{NAFNet-GoPro  } & \multicolumn{2}{c}{SRN-SBDD }(Ours)  & \multicolumn{2}{c}{NAFNet-SBDD(Ours) }  \\

    \multirow{2}{*}[6em]{ \vspace{0em} \rotatebox[origin=r]{90}{ K\"{o}hler \cite{kohler2012recording} }}   & \multicolumn{2}{c}{\includegraphics[trim=0 150 0 10, clip,width=0.18\textwidth]{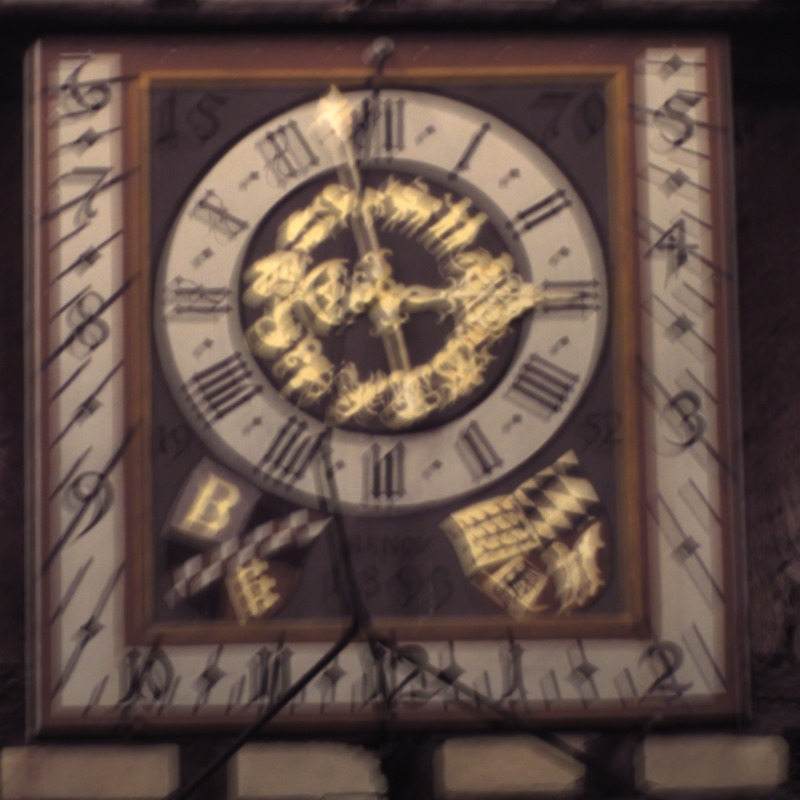}}   & 
    \multicolumn{2}{c}{\includegraphics[trim=0 150 0 10, clip,width=0.18\textwidth]{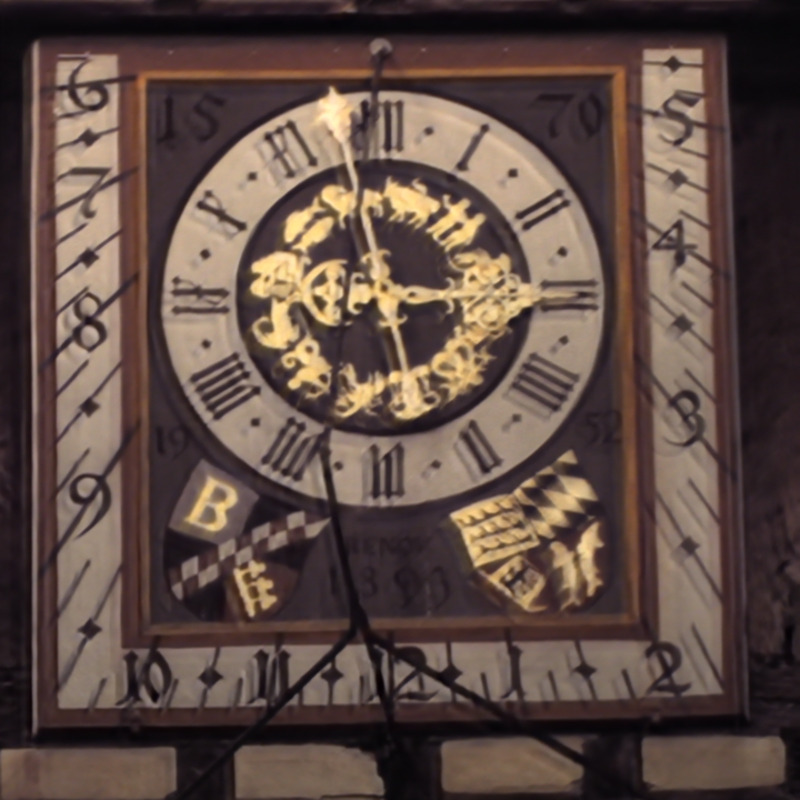}}   &
    \multicolumn{2}{c}{\includegraphics[trim=0 150 0 10, clip,width=0.18\textwidth]{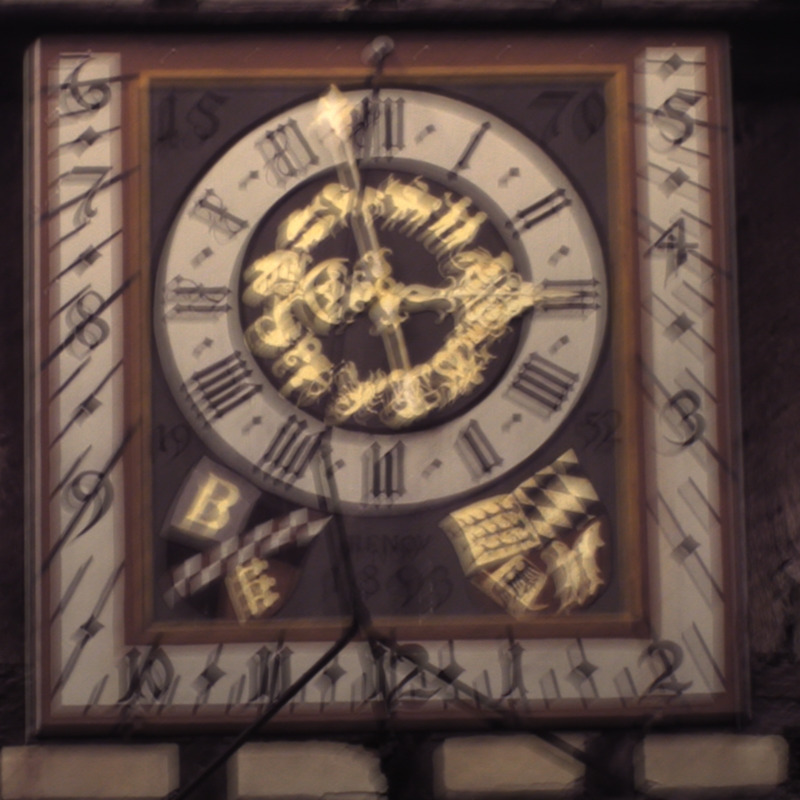}} &
    \multicolumn{2}{c}{\includegraphics[trim=0 150 0 10, clip,width=0.18\textwidth]{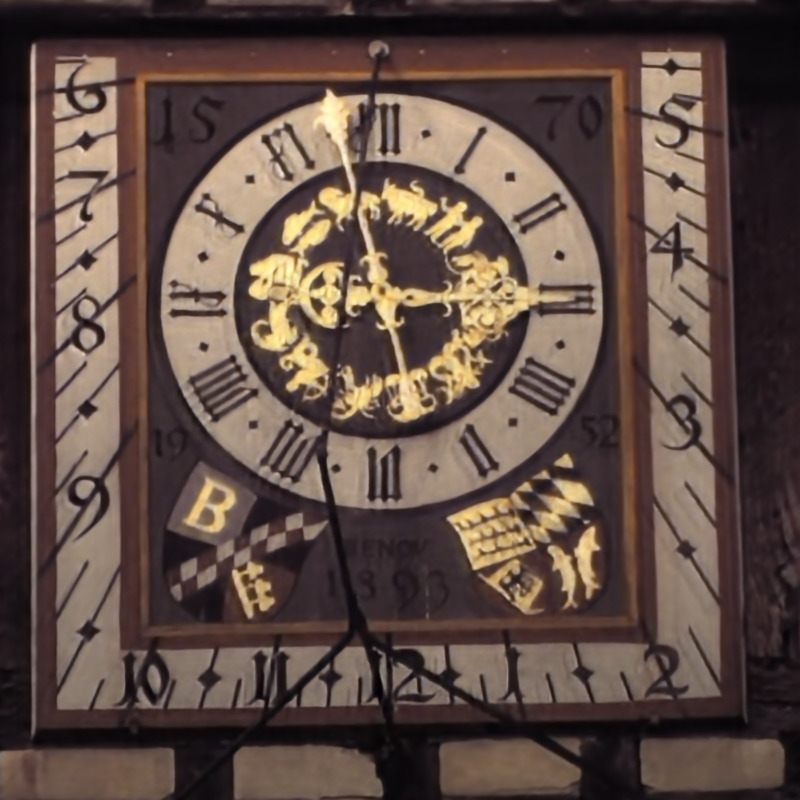}} &
    \multicolumn{2}{c}{\includegraphics[trim=0 150 0 10, clip,width=0.18\textwidth]{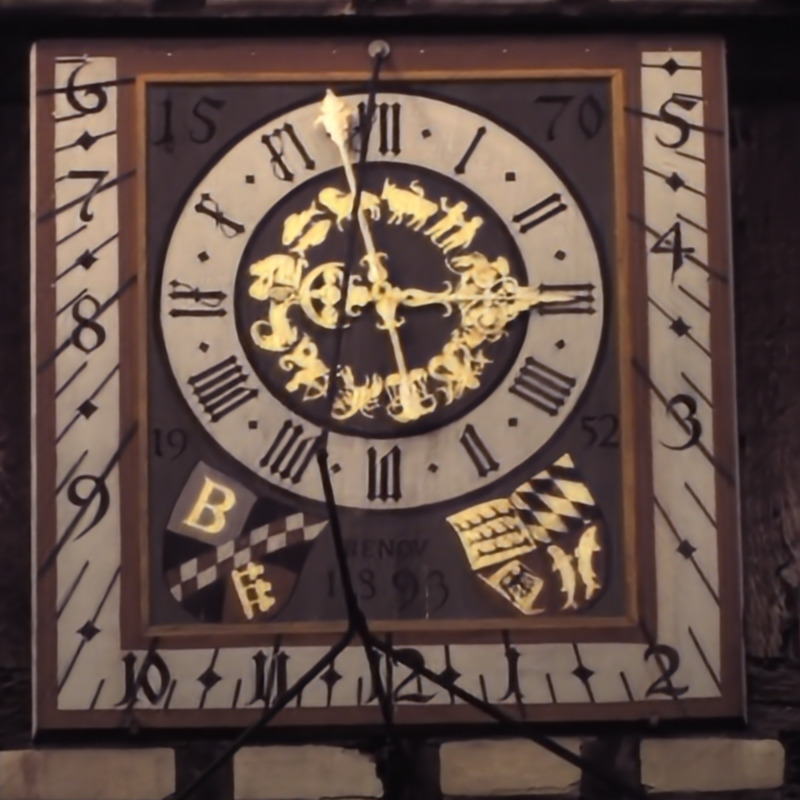}}   \\
 & \includegraphics[trim=280 270 300 350, clip,width=0.09\textwidth]{imgs/Blurry/Blurry2_1.jpg} & \includegraphics[trim=450 406 170 246,   clip,width=0.09\textwidth]{imgs/Blurry/Blurry2_1.jpg} & 
    \includegraphics[trim=280 270 300 350, clip,width=0.09\textwidth]{imgs/SRN/Blurry2_1.jpg} &
    \includegraphics[trim=450 406 170 246, clip,width=0.09\textwidth]{imgs/SRN/Blurry2_1.jpg}  &
    \includegraphics[trim=280 270 300 350, clip,width=0.09\textwidth]{imgs/NAFNet_with_GoPro/Blurry2_1.jpg} & \includegraphics[trim=450 406 170 246,                    clip,width=0.09\textwidth]{imgs/NAFNet_with_GoPro/Blurry2_1.jpg} & 
    \includegraphics[trim=280 270 300 350, clip,width=0.09\textwidth]{imgs/SRN_with_GoPro_non_uniform_mob5_ks65_texp05_F1000_ill_aug_2up_n10_gf1/Blurry2_1.jpg} & \includegraphics[trim=450 406 170 246, clip,width=0.09\textwidth]{imgs/SRN_with_GoPro_non_uniform_mob5_ks65_texp05_F1000_ill_aug_2up_n10_gf1/Blurry2_1.jpg} & 
    \includegraphics[trim=280 270 300 350, clip,width=0.09\textwidth]{imgs/NAFNet_with_GoPro_non_uniform_mob5_ks65_texp05_F1000_ill_aug_2up_n10_gf1/Blurry2_1.jpg} & 
    \includegraphics[trim=450 406 170 246, clip,width=0.09\textwidth]{imgs/NAFNet_with_GoPro_non_uniform_mob5_ks65_texp05_F1000_ill_aug_2up_n10_gf1/Blurry2_1.jpg}   \\

    \multirow{2}{*}[7em]{ \vspace{0em} \rotatebox[origin=r]{90}{ RealBlur \cite{rim_2020_ECCV} }}   & \multicolumn{2}{c}{\includegraphics[trim=0 150 0 10, clip,width=0.18\textwidth]{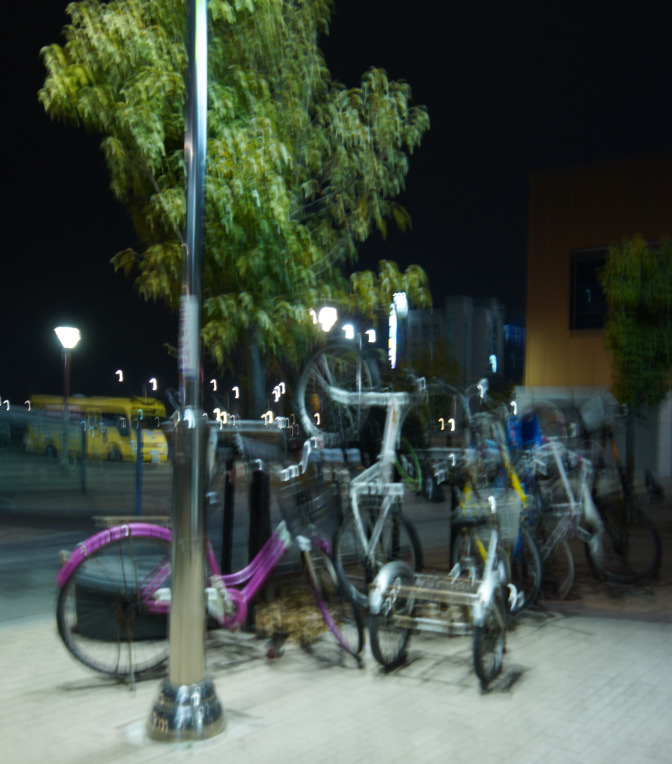}}   & 
    \multicolumn{2}{c}{\includegraphics[trim=0 150 0 10, clip,width=0.18\textwidth]{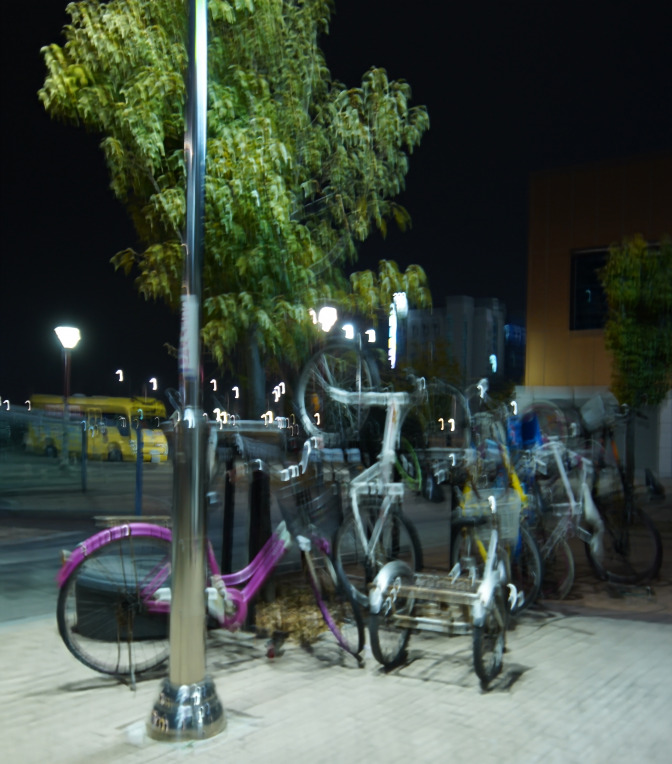}}  &
    \multicolumn{2}{c}{\includegraphics[trim=0 150 0 10, clip,width=0.18\textwidth]{imgs/NAFNet_with_GoPro/scene129_blur_19}} &
    \multicolumn{2}{c}{\includegraphics[trim=0 150 0 10, clip,width=0.18\textwidth]{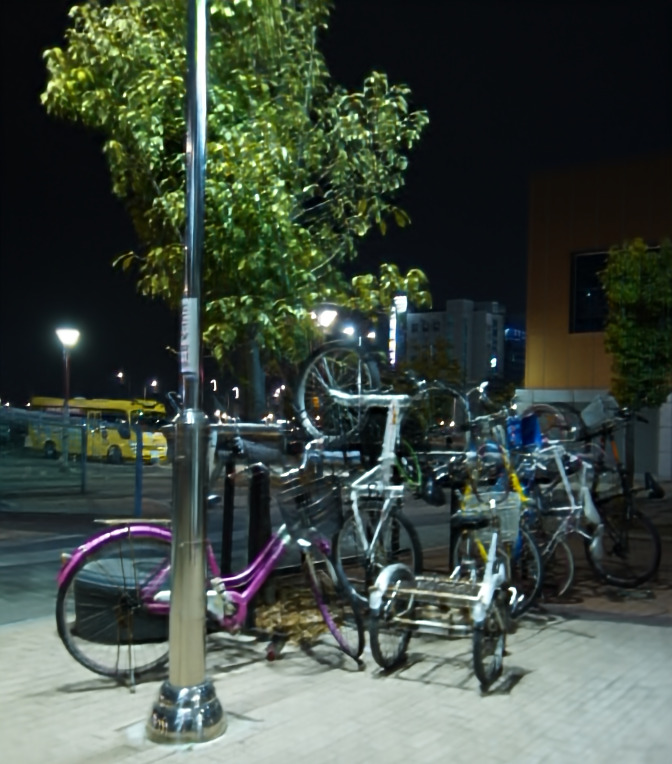}} &
    \multicolumn{2}{c}{\includegraphics[trim=0 150 0 10, clip,width=0.18\textwidth]{imgs/NAFNet_with_GoPro_non_uniform_mob5_ks65_texp05_F1000_ill_aug_2up_n10_gf1/scene129_blur_19}}   \\
 & \includegraphics[trim=100 280 400 340, clip,width=0.09\textwidth]{imgs/Blurry/scene129_blur_19.jpg} & \includegraphics[trim=180 460 310 152,   clip,width=0.09\textwidth]{imgs/Blurry/scene129_blur_19} & 
    \includegraphics[trim=100 280 400 340, clip,width=0.09\textwidth]{imgs/SRN/scene129_blur_19} &
    \includegraphics[trim=180 460 310 152, clip,width=0.09\textwidth]{imgs/SRN/scene129_blur_19}  &
    \includegraphics[trim=100 280 400 340, clip,width=0.09\textwidth]{imgs/NAFNet_with_GoPro/scene129_blur_19} & \includegraphics[trim=180 460 310 152,                    clip,width=0.09\textwidth]{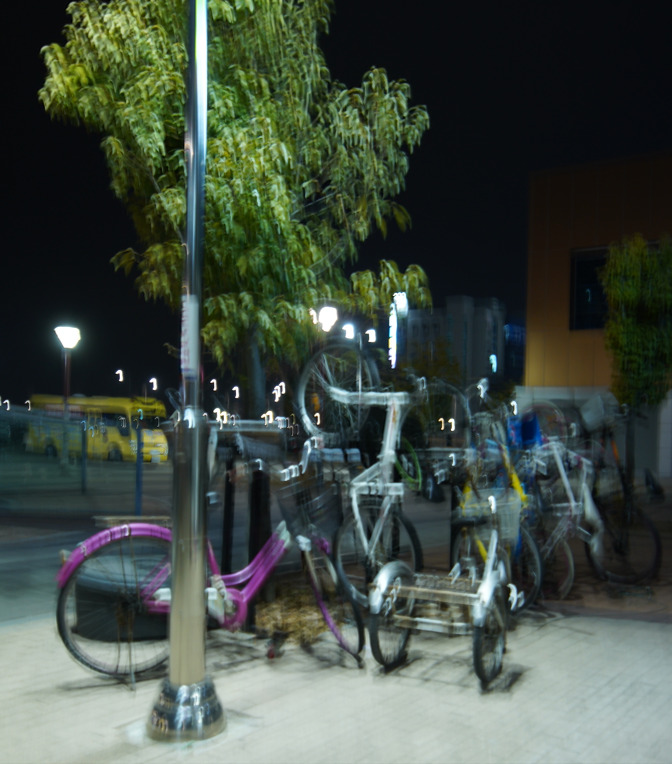} & 
    \includegraphics[trim=100 280 400 340, clip,width=0.09\textwidth]{imgs/SRN_with_GoPro_non_uniform_mob5_ks65_texp05_F1000_ill_aug_2up_n10_gf1/scene129_blur_19} & \includegraphics[trim=180 460 310 152, clip,width=0.09\textwidth]{imgs/SRN_with_GoPro_non_uniform_mob5_ks65_texp05_F1000_ill_aug_2up_n10_gf1/scene129_blur_19} & 
    \includegraphics[trim=100 280 400 340, clip,width=0.09\textwidth]{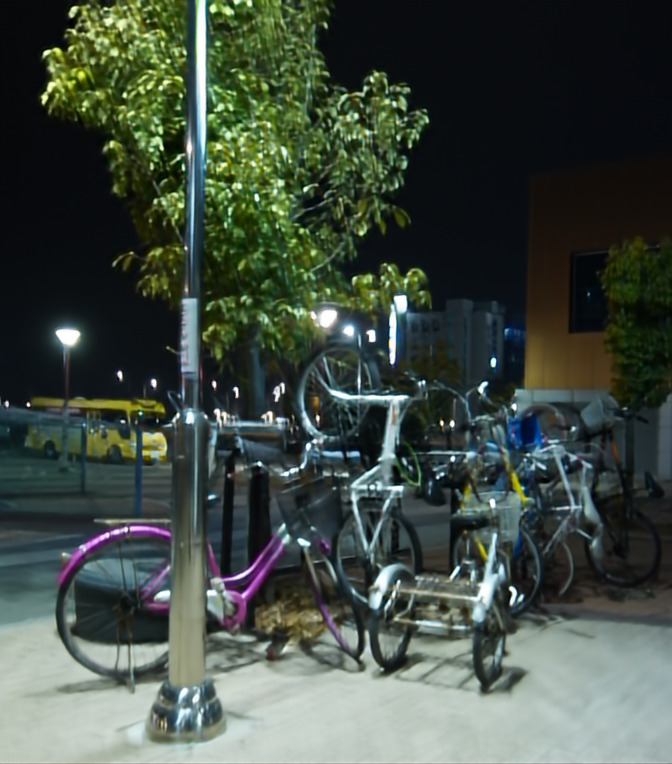} & 
    \includegraphics[trim=180 460 310 152, clip,width=0.09\textwidth]{imgs/NAFNet_with_GoPro_non_uniform_mob5_ks65_texp05_F1000_ill_aug_2up_n10_gf1/scene129_blur_19.jpg}   \\

    \multirow{2}{*}[5em]{ \vspace{0em} \rotatebox[origin=r]{90}{ Lai \cite{lai2016comparative} }}   & \multicolumn{2}{c}{\includegraphics[trim=0 150 0 10, clip,width=0.18\textwidth]{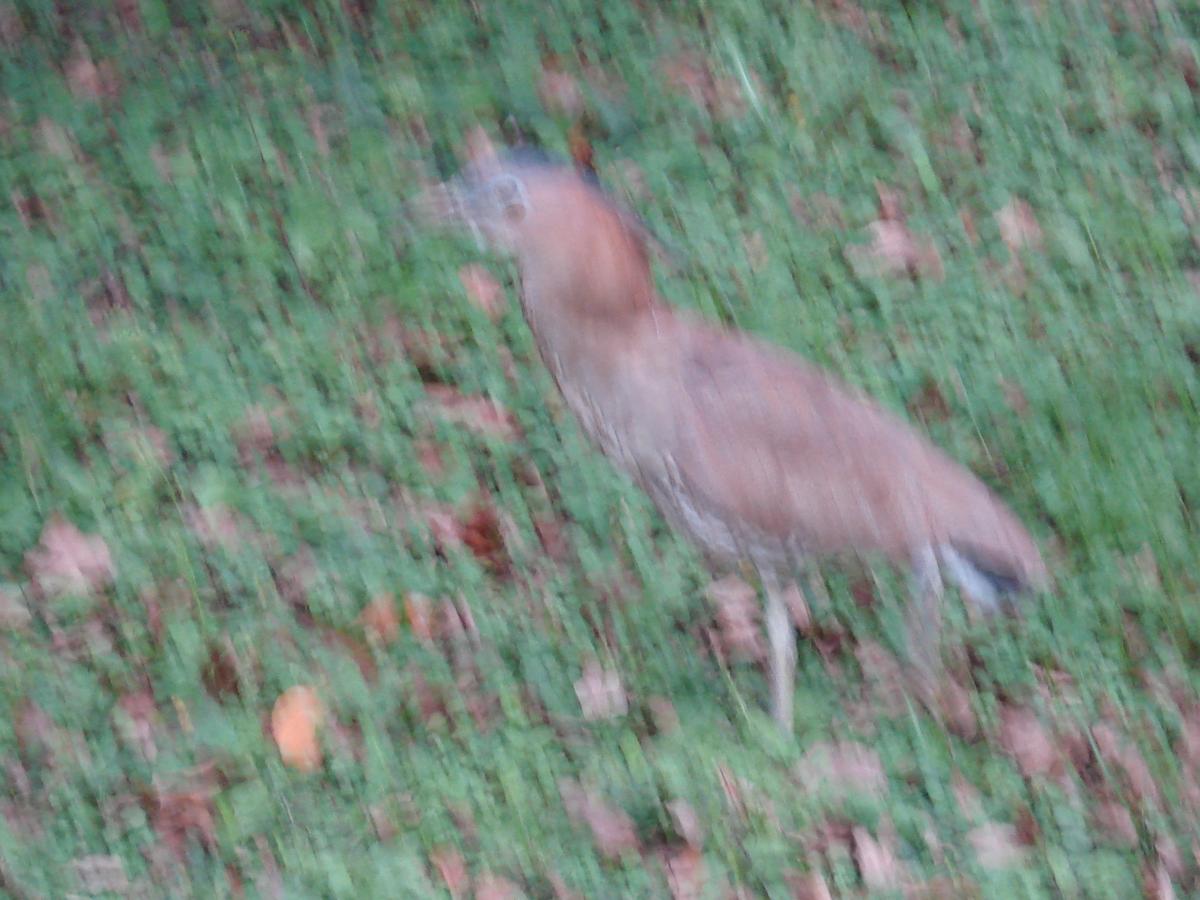}}   & 
    \multicolumn{2}{c}{\includegraphics[trim=0 150 0 10, clip,width=0.18\textwidth]{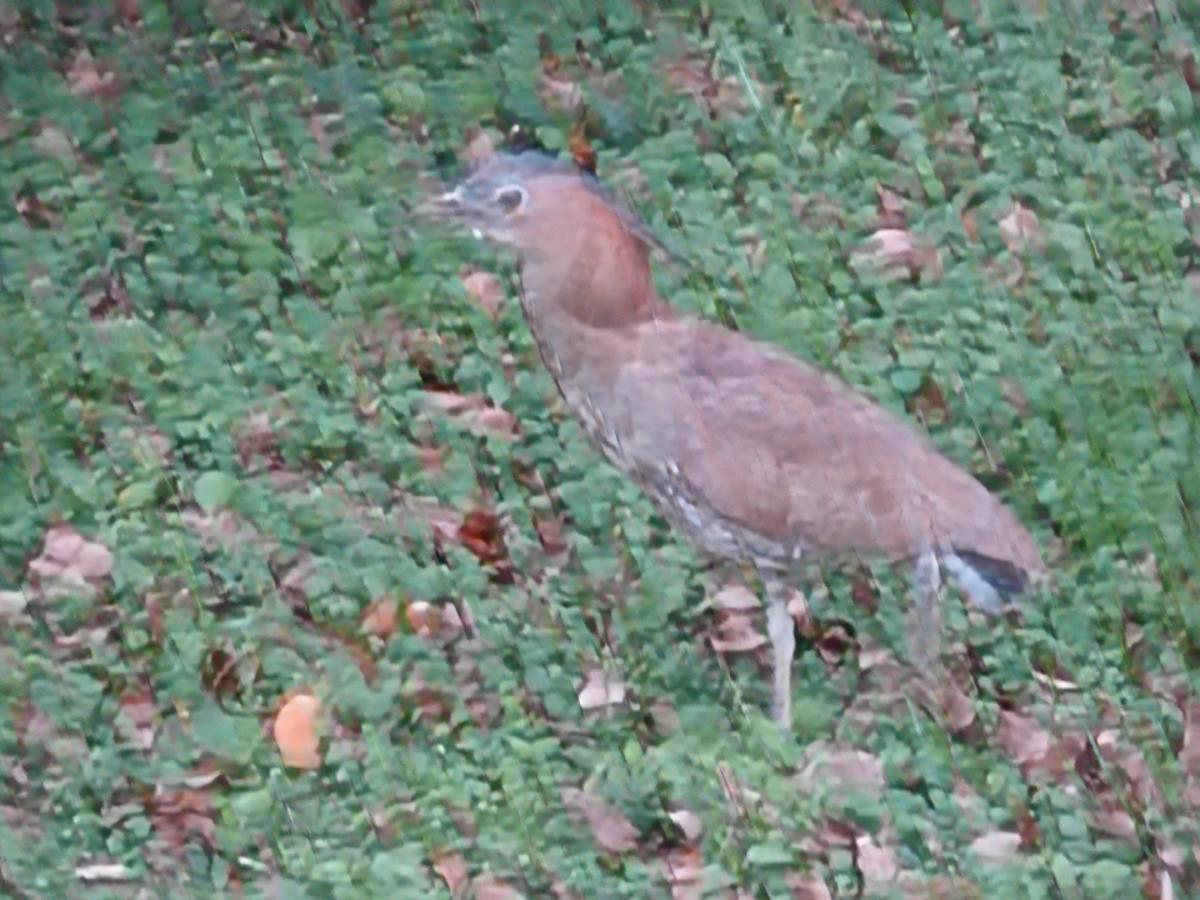}}  &
    \multicolumn{2}{c}{\includegraphics[trim=0 150 0 10, clip,width=0.18\textwidth]{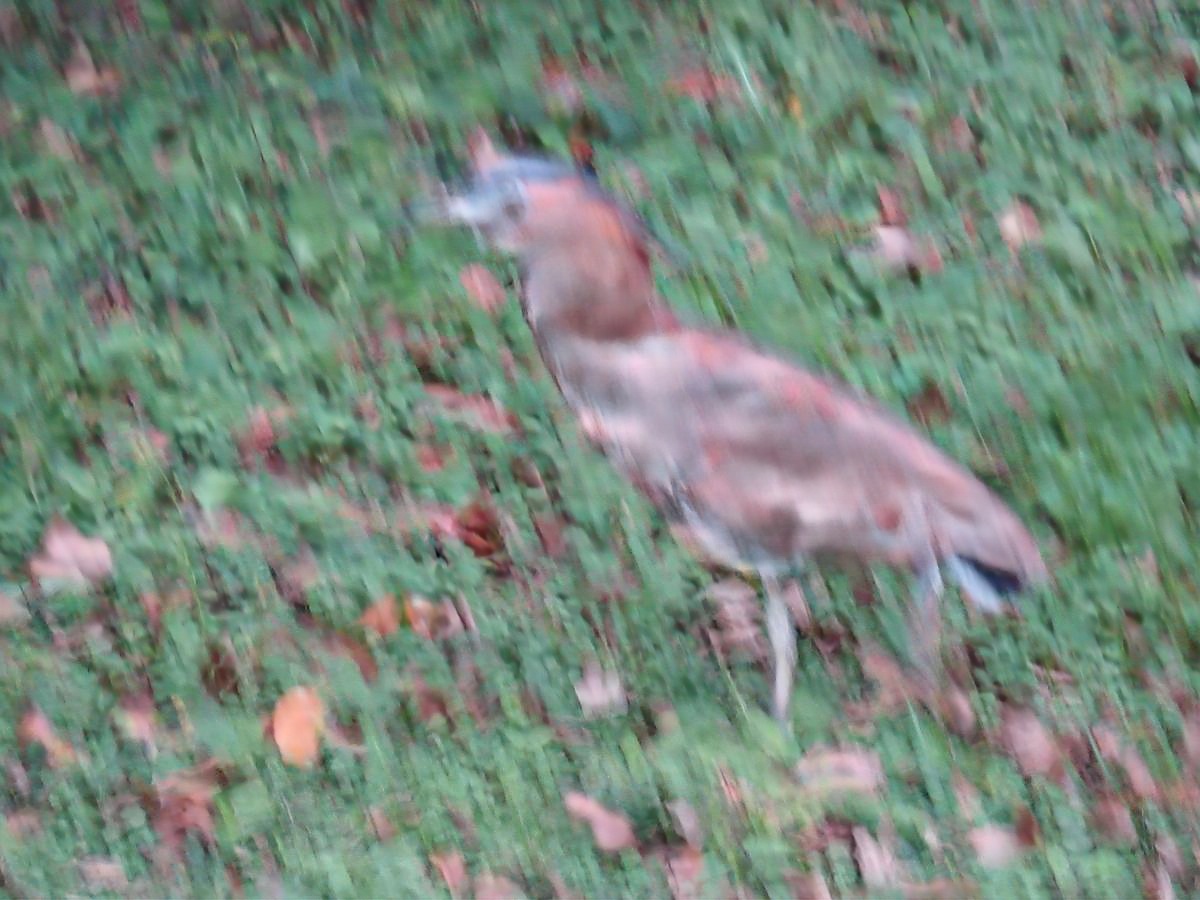}} &
    \multicolumn{2}{c}{\includegraphics[trim=0 150 0 10, clip,width=0.18\textwidth]{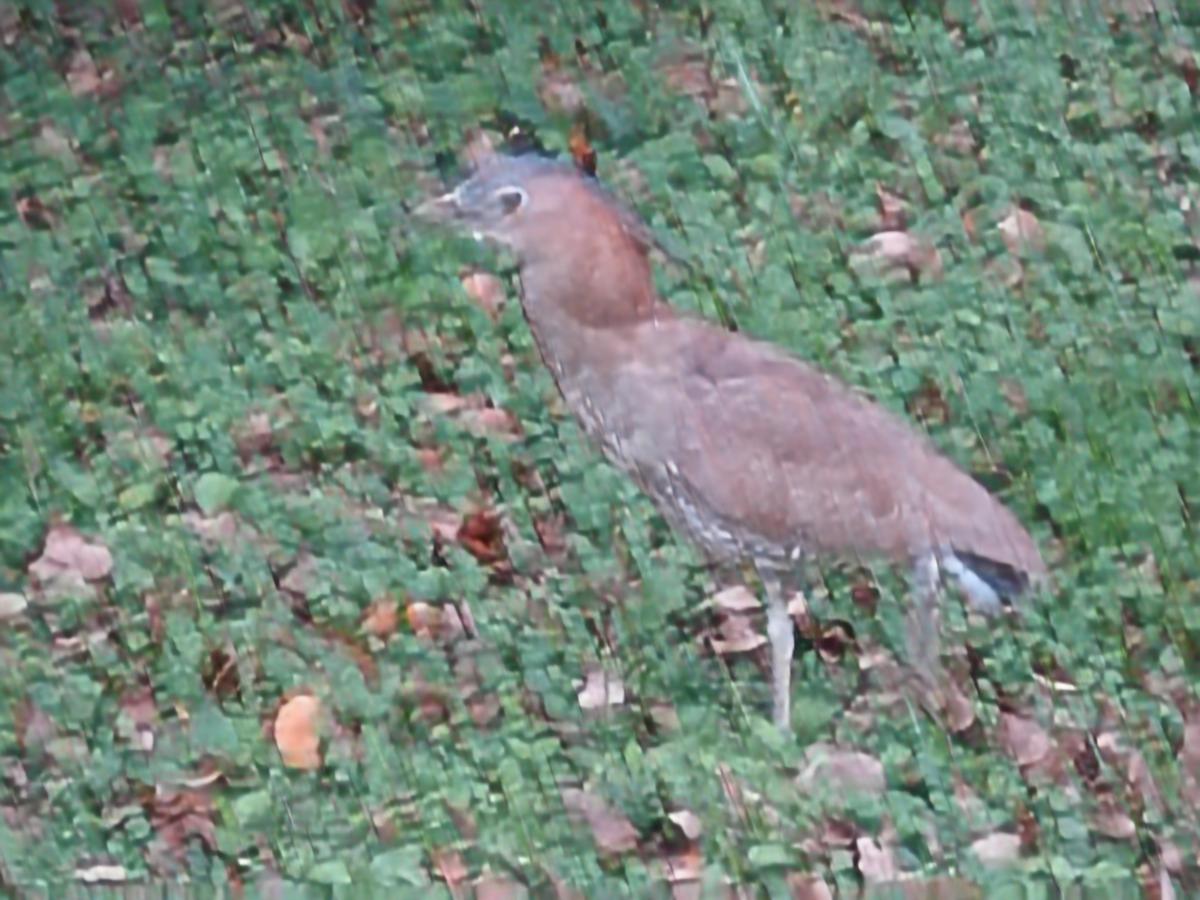}}  &
    \multicolumn{2}{c}{\includegraphics[trim=0 150 0 10, clip,width=0.18\textwidth]{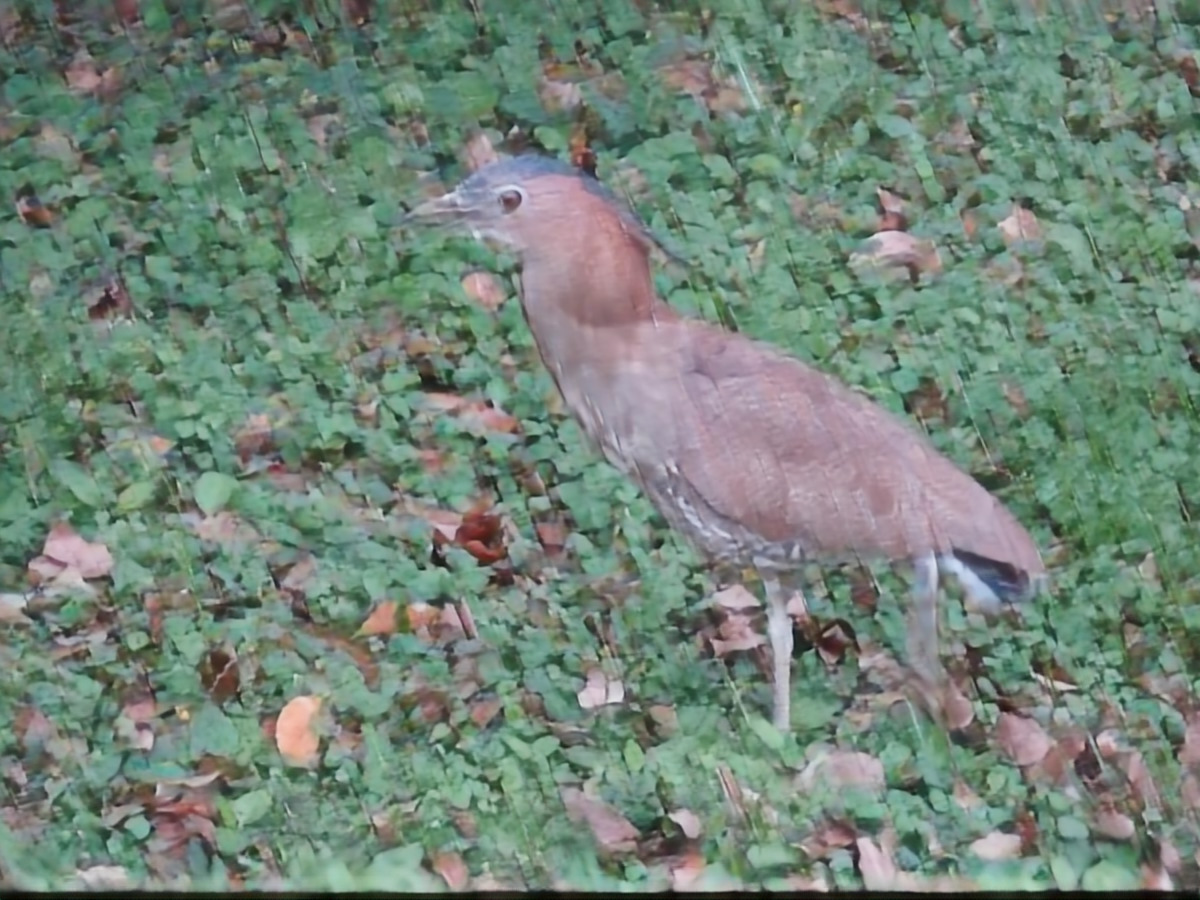}}   \\
 & \includegraphics[trim=350 480 490 80, clip,width=0.09\textwidth]{imgs/Blurry/bird} & \includegraphics[trim=680 440 320 270,   clip,width=0.09\textwidth]{imgs/Blurry/bird} & 
    \includegraphics[trim=350 480 490 80, clip,width=0.09\textwidth]{imgs/SRN/bird} &
    \includegraphics[trim=680 440 320 270, clip,width=0.09\textwidth]{imgs/SRN/bird}  &
    \includegraphics[trim=350 480 490 80, clip,width=0.09\textwidth]{imgs/NAFNet_with_GoPro/bird} & \includegraphics[trim=680 440 320 270, clip,width=0.09\textwidth]{imgs/NAFNet_with_GoPro/bird} & 
    \includegraphics[trim=350 480 490 80, clip,width=0.09\textwidth]{imgs/SRN_with_GoPro_non_uniform_mob5_ks65_texp05_F1000_ill_aug_2up_n10_gf1/bird} & \includegraphics[trim=680 440 320 270, clip,width=0.09\textwidth]{imgs/SRN_with_GoPro_non_uniform_mob5_ks65_texp05_F1000_ill_aug_2up_n10_gf1/bird} & 
    \includegraphics[trim=350 480 490 80, clip,width=0.09\textwidth]{imgs/NAFNet_with_GoPro_non_uniform_mob5_ks65_texp05_F1000_ill_aug_2up_n10_ef5/bird} & 
    \includegraphics[trim=680 440 320 270, clip,width=0.09\textwidth]{imgs/NAFNet_with_GoPro_non_uniform_mob5_ks65_texp05_F1000_ill_aug_2up_n10_ef5/bird}   \\

  \end{tabular}
  \caption{State-of-the-art deblurring neural networks achieve spectacular restorations whithin the dataset they are trained on, but generalize poorly to real blurred images. We conjecture that this is due to a discrepancy between the training set underlying degradation model and that of the actual blurred photographs of dynamic scenes. We propose a new model-based methodology for generating training pairs, that improves model performance in the ultimate task of deblurring real blurred images  (right column)}
  \label{fig:GoPro_vs_SBDD}
\end{figure*}

One plausible explanation for the lack of generalization in motion deblurring is that existing datasets \citep{Nah_2017_CVPR, Nah_2019_CVPR_Workshops_REDS, su2017deep, rim_2020_ECCV} may not faithfully represent the complexities of real-world motion blur processes. Compared to other image restoration tasks, such as denoising or superresolution, generating vast amounts of realistic source/target image pairs for training is much more cumbersome, and therefore, training sets used in practice are limited in diversity.

 Alternatively, the issue could be due to the architectural limitations of neural networks. Instead of focusing on generalization, much of the literature optimizes performance metrics on benchmark datasets. This trend has led to the development of larger and more complex neural architectures \citep{cho2021rethinking, chen2022simple, zhang2020deblurring}. 

\cref{fig:state_of_the_art} shows that classic networks like the Scale-Recurrent Network (SRN)\citep{tao2018scale} demonstrate superior performance on real images, challenging the assumption that larger models inherently outperform their counterparts in terms of both benchmark performance and real-world applicability. This may indicate that current deblurring networks are overfitted to benchmark datasets. By changing the training set, we improve the performance of both state-of-the-art and classic networks on real images. \cref{fig:GoPro_vs_SBDD} shows some examples. More importantly, we shed light on the essential factors that hinder or obstruct the path to effective generalization in real-world motion deblurring scenarios. We aim to unravel the underlying issues through empirical analyses, contributing to a more in-depth understanding of motion deblurring generalization.

\subsection{Background}

The image formation process of real motion-blurred photos of dynamic scenes is extremely complex and, therefore, hard to model. The unprecedented progress in learning approaches achieved by deep learning in vision tasks has shifted the attention of the motion deblurring community from classic model-based methods toward supervised learning approaches. Recently, spectacular results have been reported by end-to-end deep neural networks~\citep{tao2018scale, nah2017deep, Zhang_2019_CVPR, kupyn2018deblurgan, kupyn2019deblurgan, Chen_2021_CVPR, cho2021rethinking, zhang2020deblurring, chen2022simple}. These supervised learning approaches are trained on pairs of sharp and motion-blurred images obtained by leveraging high-speed cameras~\citep{agrawal2009optimal,su2017deep,wieschollek2017learning,kim2017dynamic,nah2017deep,nah2019ntire,shen2019human}, beamsplitter-based setups~\citep{rim_2020_ECCV, rim_2022_ECCV, zhong2023real}, or synthetically blurred images~\citep{kaufman2020deblurring}. However, these datasets are bounded in their diversity, making models trained on them limited in their ability to generalize to real ``in the wild'' motion-blurred images \citep{zhang2022deep}. Indeed, as pointed out by~\cite{m_Tran-etal-CVPR21}, many deep-learning models degenerate to nearly an identity map when tested on out-of-domain blur operators. %

In the context of the approximation-generalization trade-off, current research has shown that CNNs excel in deblurring images with different amounts and types of blur within a dataset.  A natural step forward is to design methods with good out-of-distribution performance, capable of performing well in real scenarios. 
In this work, instead of focusing on developing new architectures, we take a data-centric approach by investigating the properties in the synthetic training set that cause the generalization gap. 

\subsection{Contributions}
We first present a thorough review of the three common strategies for generating deblurring training pairs used in the literature, namely: 1) Frame averaging from high-speed cameras, 2) Dual-camera setups with different exposure configurations, and 3) Synthetically blurred images by convolving a sharp image with a given kernel.  Following this analysis, we conjecture about the reasons behind the lack of generalization of networks trained with the first two approaches. Then, we propose a synthetic dataset generation methodology based on the third strategy, that generates training pairs inducing powerful generalization in existing deblurring models, as demonstrated on three standard datasets of real blurred images \citep{kohler2012recording,lai2016comparative,rim_2020_ECCV}.

Compared to existing approaches, our generation methodology naturally allows for more diverse training pairs and takes into account saturated pixels and multiple object motions in the scene. The proposed dataset generation procedure is simple yet effective and allows distilling the impact in the network's performance of modeling aspects like $\gamma$-correction, blur kernel distribution, non-uniformity, and pixel saturation. Experimentally, we found that those elements are crucial for obtaining good generalization to real images, as they allow the network to see smoothly varying motion blur kernels within objects and, simultaneously, to learn to restore patches where multiple motion blurs and saturation coexist. The source code to generate synthetic datasets is publicly available at \url{https://github.com/GuillermoCarbajal/SBDD}.

\section{Ill-posedness of motion deblurring:  intrinsic and extrinsic factors}

In this section, we briefly discuss the motion blur formation model. The objective is to identify the major scene-related and camera-related factors that control motion blur in images and then analyze how the different database generation strategies take them into account.

Motion blur occurs when relative motion exists between the
camera and the scene during exposure time. 
As a result, the camera sensor at each pixel receives and accumulates light from different sources, producing a blurry output. The blurring process is modeled as follows:
\begin{equation}
    \mathbf{v} = g \left( \frac{1}{T} \int_{t=0}^{T} \tilde{\mathbf{u}}(t) dt \right),  \label{eq:motion_blur_model}
\end{equation}
where $T$ and $\tilde{\mathbf{u}}(t)$ denote the exposure time and the sensor signal of a sharp image at time $t$, respectively. The function $g$ is the  Camera Response Function (CRF) that maps the number of photons accumulated during the exposure time $T$ into an observed intensity image. 

From \cref{eq:motion_blur_model}, it can be seen that for dynamic scenes, the problem of associating a single sharp frame to a given blurry image is not well-defined.  This ambiguity also extends to static scenes observed by a moving camera, particularly in scenarios where objects may undergo occlusion or appear during the exposure period due to variations in scene depth. A well-defined scenario occurs when the scene is planar, thereby allowing us to consider $\mathbf{\tilde{u}}(t)$ as equivalent to $\mathbf{\tilde{u}}(t+\Delta t)$. It is worth noting, however, that even in the planar case, the scene may be affected by illumination changes during the exposure time, and therefore, the assumption may not be valid. Consequently, the problem of recovering a single sharp image from a blurry one is highly ill-posed.    

\begin{figure}[ht]
\setlength{\tabcolsep}{1pt}
    \centering
    \begin{tabular}{cc}
    Sharp pattern &  Blurry pattern ($\gamma$=1) \\
    \fbox{\includegraphics[width=0.22\textwidth]{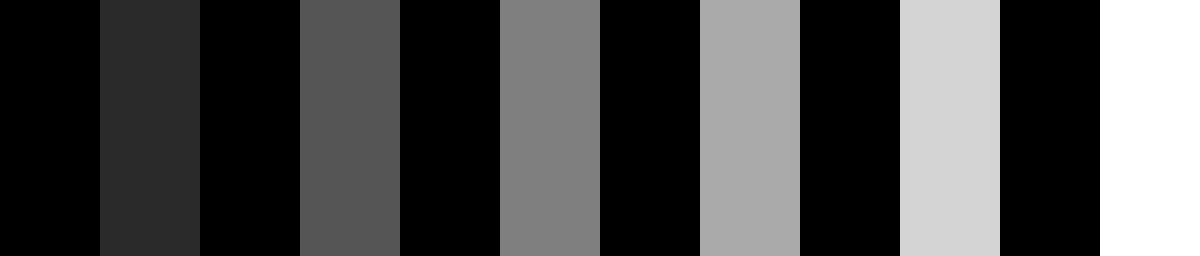}}& 
    \fbox{\includegraphics[width=0.22\textwidth]{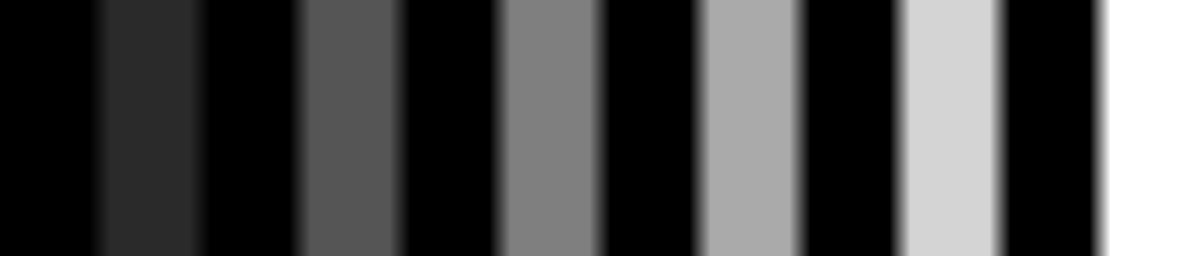}} \\
    Blurry pattern ($\gamma$=2.2) & Blurry pattern ($\gamma$=4.0) \\
    \fbox{\includegraphics[width=0.22\textwidth]{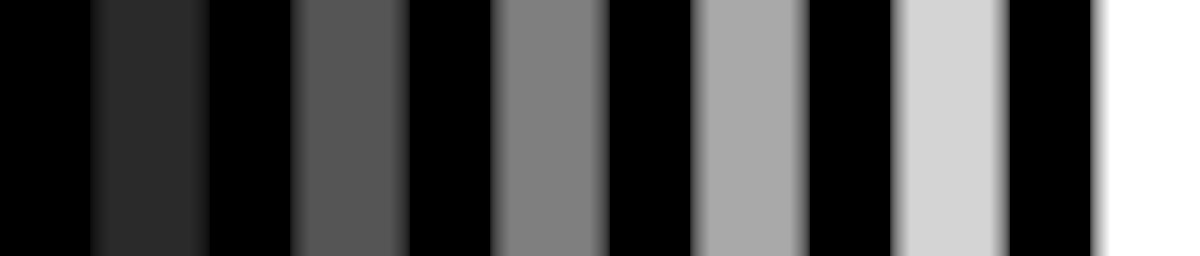}} &
    \fbox{\includegraphics[width=0.22\textwidth]{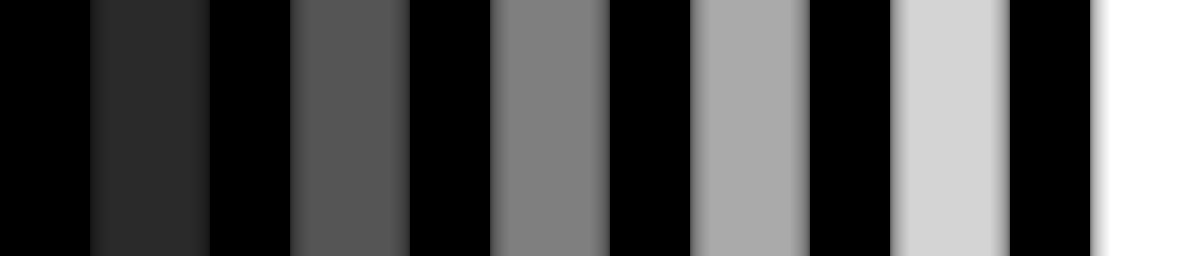}} \\
    \end{tabular}
    \caption{A synthetic pattern with varying height steps was generated to show the effect of the Camera Response Function (CRF) on the blurry borders of an image. The CRF is modeled as $g({x})={x}^{1/\gamma}$. As the gamma value increases, the light steps look wider than the dark ones.  \label{fig:step_patterns} }
\end{figure}

In addition, the limited dynamic range of camera sensors makes the problem even more difficult. Furthermore, the transformation of scene irradiance into pixel values is camera-dependent and varies depending on factors such as white balance, tone mapping, and post-processing. All these factors also affect blurry images, making it challenging to disentangle them from the motion blur. For example, nonlinear CRFs strongly impact the blur within the image \citep{tai2013nonlinear}, causing a spatially invariant blur to behave as a spatially varying blur. Moreover, the same camera motion can yield visually different borders depending on the CRFs in play.  While a pre-calibrated CRF is undoubtedly the optimal solution, it may not always be available, especially when the camera operates in ``auto'' mode, where the CRF varies in response to the scene's characteristics. \cref{fig:step_patterns} illustrates, on synthetic edges, how the effects of CRF and motion blur are strongly intertwined.

\begin{figure}[ht]
\setlength{\tabcolsep}{0pt}
\small 
\begin{tabular}{ccc}
     \hspace{3.8em} Blurry \hspace{3.4em} & Blurry($\gamma$-corrected) \hspace{1.5em} & Sharp  \\
    \multicolumn{3}{c}{\includegraphics[width=0.5\textwidth]{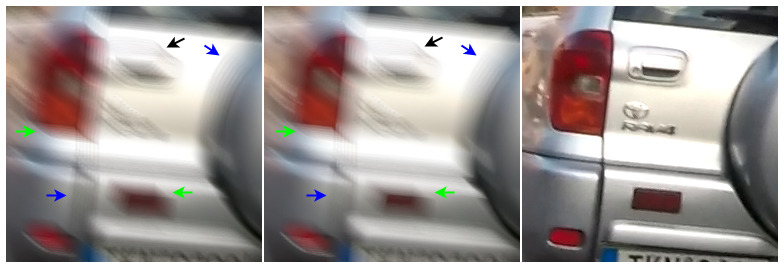}} \\
\end{tabular}
\caption{Blurry/sharp pair crop sourced from the GoPro dataset for the linear and $\gamma$-corrected cases. The crops correspond to a saturated case. The blue arrows indicate the ghosting effect, while the green arrows show that the $\gamma$-corrected image has ``whiter'' transitions. The black arrow indicates the door opener borders present in the synthetic blurry images. However, these borders would not have been visible in a real blurry image since the surrounding pixels are saturated. Best viewed in electronic format. \label{fig:gopro_training_images}}
\end{figure}

\begin{figure}[ht]
    \setlength{\tabcolsep}{0.5pt}
    \begin{tabular}{cc}
     Blurry & Sharp \\
     \includegraphics[width=0.24\textwidth,clip,trim=370 457 240 262]{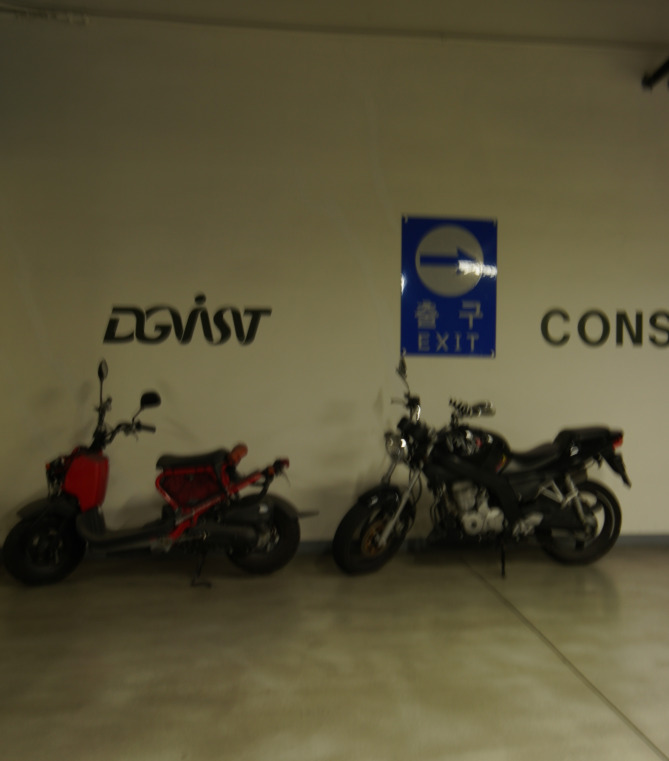} & 
    \includegraphics[width=0.24\textwidth,clip,trim=370 457 240 262]{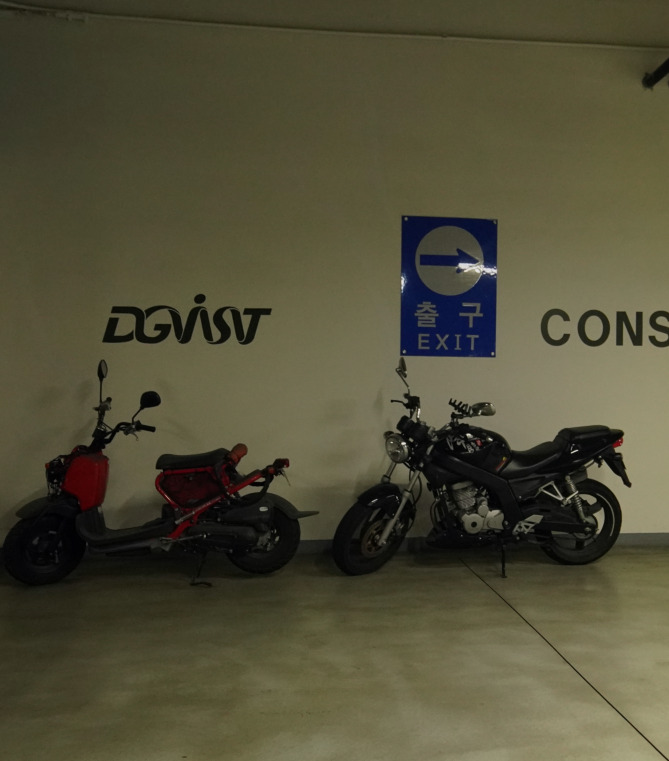}
     \\ 
     \includegraphics[width=0.24\textwidth,clip,trim=300 350 280 350]{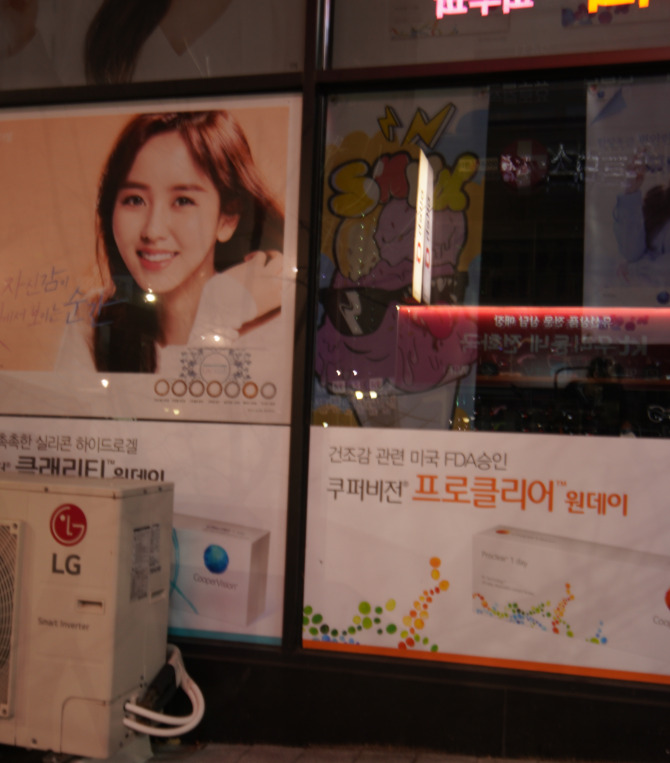}  &
    \includegraphics[width=0.24\textwidth,clip,trim=300 350 280 350]{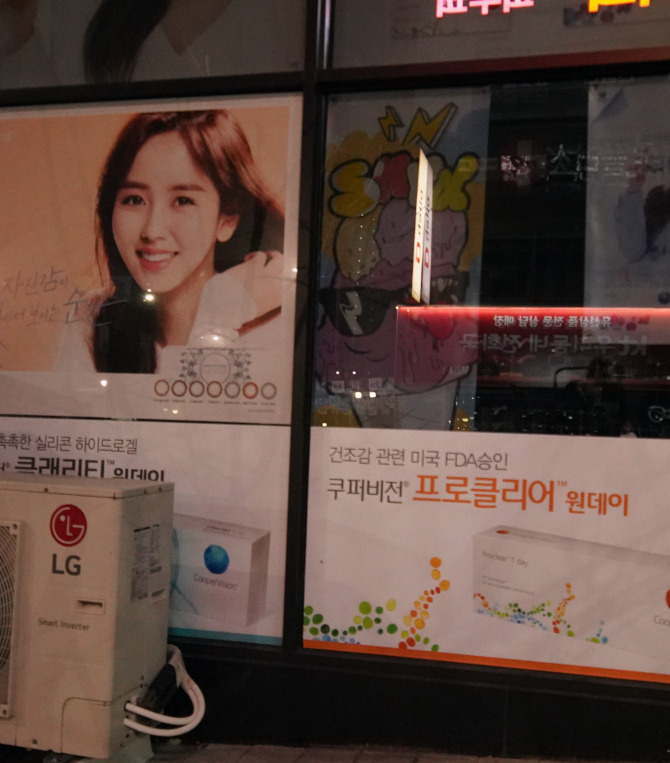} \\
    \end{tabular}
    \caption{Blurry/sharp pair crop sourced from the RealBlur dataset It is observed that saturated regions in the blurry image are larger, with higher pixel values than the corresponding sharp image, contradicting the motion blur model. Best viewed in electronic format.}
    \label{fig:realblur_training_images}
\end{figure}

\section{Analysis of strategies for generating motion deblurring benchmarks}

In this section, we present the approaches commonly used to produce motion-deblurring benchmark datasets and analyze why the performance on these datasets does not reflect the efficacy of deblurring methods on real images.

\subsection{High-Speed Camera-based Generation}

This approach has been used to produce extremely popular datasets in the motion deblurring community~\citep{agrawal2009optimal,su2017deep,wieschollek2017learning,kim2017dynamic,nah2017deep,Nah_2019_CVPR_Workshops_REDS,shen2019human}. The ground truth pairs of blurry/sharp images are extracted from short video sequences acquired with high-speed video cameras. For each sequence, the middle frame is defined as the sharp image, while the motion-blurred image is synthesized by averaging all the frames in the sequence. 
This blur generation procedure is the discrete approximation of the blur generation model given by \cref{eq:motion_blur_model}, that is,
\begin{equation}
    \mathbf{v} \simeq g \left( \frac{1}{M}  \sum_{i=0}^{M-1} \tilde{\mathbf{u}}[i] \right),
    \label{eq:high_speed_blur_model}
\end{equation}
where $M$ and $\tilde{\mathbf{u}}[i]$ are the number of sampled frames and the \textit{i}-th sharp frame signal captured during the exposure time, respectively. A common practice is to approximate the CRF as a gamma curve with $\gamma = 2.2$, 
\begin{equation}
    g(x)=x^{1/\gamma},
    \label{eq:gamma_function}
\end{equation}
as it is an approximated average of known CRFs \citep{tai2013nonlinear}. Thus, by correcting for the gamma function, the latent frame signal $\tilde{\mathbf{u}}[i]$ is obtained from the observed image $\mathbf{u}[i]$, i.e. $\tilde{\mathbf{u}}[i] = (\mathbf{u}[i])^\gamma$. 
It is worth noticing that the GoPro dataset \citep{Nah_2017_CVPR} has two versions, one which assumes a linear CRF ($\gamma$=1) and another that applies $\gamma$-correction with $\gamma=2.2$. We call the latter GoPro($\gamma$=2.2). Similar to the GoPro dataset, DVD \citep{su2017deep} does not account for gamma correction but incorporates alignment between frames and interpolation. REDS \citep{Nah_2019_CVPR_Workshops_REDS} is a much larger dataset composed of more diverse scenes. The blurry images were generated using a calibrated CRF.

\paragraph{Limitations} %
The literature on motion deblurring has identified two primary factors that prevent networks trained using high-speed cameras to generalize to real images \citep{rim_2022_ECCV, zhong2023real, nah2021ntire}. Firstly, the unique characteristics of the recording devices employed for each dataset can hinder the ability of algorithms to generalize to other cameras. Secondly, in contrast to real motion blur kernels, the implicit kernels utilized in this process are prone to be discrete (``dotted kernels'') since cameras operate with a duty cycle that prevents continuous acquisition. Consequently, an undesirable ghosting effect \textcolor{blue}{(c.f. \cref{fig:gopro_training_images}) } becomes evident in the presence of fast-moving objects,  which deviates significantly from the appearance of genuine motion-blurred images. One approach to mitigate ghosting involves using cameras with even higher frame rates. However,  this comes with its challenges, as the ``sharp'' frames captured in such scenarios may not be suitable due to their low signal-to-noise ratio. The REDS dataset \citep{Nah_2019_CVPR_Workshops_REDS} performs inter-frame interpolation to reduce ghosting while keeping the noise level reasonable. While REDS proved effective when training large networks such as HiNet \cite{Chen_2021_CVPR} or NAFNet \cite{chen2022simple}, notably in outdoor environments with natural light such as the dataset provided by \citep{zhang2020deblurring}, its widespread adoption remains limited. This is likely due to its size, which is considerably larger than GoPro and demands more resources for training, and because most of the generalization shortcomings still need to be addressed.

In this study, we present empirical evidence challenging the common assertion in the literature that ``dotted kernels'', which are sequences of discrete delta functions resulting from undersampling rapid motion in video, are the main cause of poor generalization \citep{nah2019ntire, su2017deep,rim_2022_ECCV,zhao2023representing}. While rarely encountered in practice, these kernels can indeed train neural networks. Instead, we posit other influential factors. Firstly, the approximation assumption stated in \cref{eq:high_speed_blur_model} does not hold due to saturated pixels within the sharp images. This leads to two noteworthy consequences: i) Saturated pixels whose values are below their value on ideal conditions. When the loss of information is significant, the image has saturated regions that hinder the motion inference from it. ii) Since sharp images generate blurry counterparts, blurry pixels influenced by saturated pixels exhibit values lower than they should. In an extreme case, though not entirely uncommon, the synthetic blurry image may contain structures that would remain invisible in a genuine image due to substantial information loss. \cref{fig:gopro_training_images} illustrate these situations.  
Secondly, as the ``sharp'' frame is typically identified as the middle frame of the sequence, certain moving objects may appear only on the blurry frame. When training with these pairs, the network must not only learn to deblur but also to discern and discard structures that are absent in the target image. The problem becomes more challenging with longer time windows, as it becomes increasingly ill-posed, and therefore, interpolating frames and keeping the number of averaged frames constant help mitigate this problem. However, datasets employing frame interpolation tend to produce smooth results, which may be related to the interpolation. Investigating the impact of frame interpolation on deblurring outcomes is an intriguing research avenue, although it falls outside the scope of this article. Our intuition suggests that frame interpolation helps to pose the deblurring problem better by constraining the motion kernel space. As a side effect, the kernel space mainly comprises straight line-shaped motion kernels, while other shapes are underrepresented.

\subsection{Beam Splitter-based datasets}

Another strategy, proposed by \citet{rim_2020_ECCV}, relies on an image acquisition system that consists of two cameras built into a beamsplitter setup that simultaneously captures pairs of blurred and sharp images using different exposure times. Using this setup, \citet{rim_2020_ECCV} built RealBlur, the first large-scale dataset of real-world blurred images suitable for training image deblurring deep networks. The dataset, collected in
low-light environments, consists of two subsets sharing the same image contents, one generated from camera raw images and the other from JPEG images processed by the camera Image Signal Processor (ISP). In both cases, the captured images were also post-processed for noise reduction as well as for geometric and photometric alignment.  
More recently, the same authors proposed RSBlur~\citep{rim_2022_ECCV}, a dataset of real blurry images and a sequence of sharp frames associated with each blurry frame. \citet{zhong2023real} followed a similar procedure and generated a blurry/sharp video clip dataset.

\paragraph{Limitations} As pointed out by \citet{rim_2022_ECCV}, models trained on datasets generated following this procedure tend to exhibit poor performance when applied to other datasets. A plausible explanation, mentioned by \citet{rim_2022_ECCV}, is that these datasets are acquired using a \textit{single camera pair}. Expanding the scope and diversity of these datasets poses significant challenges since collecting them requires a specially designed camera, which is tremendously laborious. Additionally, despite diligent efforts to obtain high-quality blurry/sharp pairs, these datasets often exhibit illumination inconsistencies, as exemplified in \cref{fig:realblur_training_images}. Notably, these datasets encompass images captured with varying exposure times, leading to a scenario where blurry/sharp pairs do not depict the same scene under identical conditions. Further inconsistencies may arise from scene-dependent camera ISP procedures. In our view, those deviations from the motion blur formation model limit their generalization capabilities. Interestingly, we devised a dataset of blurry/sharp pairs following a motion blur degradation model with images taken by a single camera. Remarkably, our approach achieves good generalization properties compared with these methods.

\subsection{Synthetic kernel-based datasets}

A third strategy, commonly employed in the case of spatially uniform motion blur, involves the convolution of sharp images with synthesized blur kernels. Here, the real scene is typically assumed to be static and planar, and the blur operator is uniform across the entire image. When working with such data, the generalization ability of CNNs is superior among static and planar scenes \citep{kaufman2020deblurring, rim_2020_ECCV}. 
\paragraph{Limitations}
This approach is not widely adopted due to its limitations, particularly when dealing with scenarios where motion blur kernels vary rapidly in adjacent regions.

\begin{figure}[t!]
    \centering
    \includegraphics[width=0.48\textwidth]{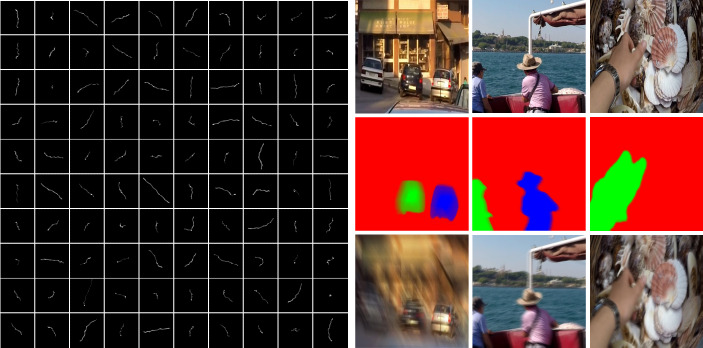}
    \caption{Examples of synthetic blurred images generated by the proposed procedure. From left to right: camera-shake motion kernels from physiological tremor model \citep{gavant2011physiological}; sharp image from the GoPro dataset~\citep{Nah_2017_CVPR}; segmentation masks generated with Mask-CNN \citep{wei2018mask}; resulting motion blurred image, obtained by convolving each region in the image with a different blur kernel. %
    }
    \label{fig:dataset_generation}
\end{figure}

\paragraph{Overview of the proposed approach}
In this work, we propose extending this procedure to face the challenges posed by saturated and non-uniformly motion-blurred images. Before further describing the proposed procedure for generating blurry/sharp image pairs, it is interesting to delve deeper into the fundamental difference between a convolution-based procedure and the other two approaches presented above. Indeed, these are two radically different paradigms. On the one hand, the high-speed camera-based and the beam-splitter-based generation procedures build on the rationale of acquiring as faithful as possible to reality. However, they fail to associate a single sharp image with a blurred image since the blurry image is the average of an entire sequence of sharp images. On the other hand, a method based on convolutional synthesis assumes a planar approximation of the real scene, which may seem simplistic but holds locally in most cases and has the advantage that a single well-defined sharp image generates the blurry image. We claim, and support with experiments, that this latter approach generalizes better to real motion-blurred images. More importantly, this approach enables us to identify the primary factors contributing to the limited generalization observed in deblurring networks.

\FloatBarrier

\section{Segmentation-Based Generation of Motion Deblurring Training Pairs}

In this section, we propose a simple yet effective method to generate blurry/sharp image pairs accounting for non-uniform motion blur, that overcomes several limitations of the existing datasets. Using this procedure, we generate a new dataset, called \emph{SBDD} (Segmentation-Based Deblurring Dataset)\footnote{Dataset and source code for generation are publicly available at \url{https://github.com/GuillermoCarbajal/SBDD}}. %

\subsection{The Non-Uniform Motion Blur Degradation Model} \label{subsec:model}

Non-uniform motion blur can be modeled as the local per-pixel convolution of a sharp image with a spatially varying filter, the \emph{motion blur field}. Given a sharp image $\mathbf{u}$ of size $H\times W$, and a set of per-pixel blur kernels $\mathbf{k}_i$ of size $K\times K$, the observed blurry image $\mathbf{v}$ is generated as
\begin{equation}
    {v}_i = \langle \mathbf{u}_{nn(i)}, \mathbf{k}_i \rangle + n_i,
        \label{eq:model}
\end{equation}
where $\mathbf{u}_{nn(i)}$ is a window of size $K \times K$ around pixel $i$ in image $\mathbf{u}$ and $n_{i}$ is additive noise. We assume non-negative kernels (no negative light) of area one (conservation of energy).

By considering the sensor saturation and the CRF $g$, the model becomes
\begin{equation}
    v_i = g\left( R \left( \right \langle g^{-1} \left(\mathbf{u}_{nn(i)}\right), \mathbf{k}_i  \rangle + n_i )\right) ,
      \label{eq:model_sat}
\end{equation}
where  $R(\cdot)$ is the pixel saturation operator that clips image values $v_{i}$ which are larger than 1 (the dynamic range is normalized to $[0,1]$). 

Simulating realistic blurry images with pixel-wise non-uniform kernels is extremely hard. On the other hand, assuming uniform blur across the image is unrealistic in most cases, as illustrated in ~\cref{fig:non_uniform_eval_GoPro}. An intermediate approach that allows to extend the capacity of deblurring networks to deal with non-uniform blur is to assume piece-wise constant blur:

\begin{equation}
    v_i = g \left( R \left( \right \langle g^{-1} \left(\mathbf{u}_{nn(i)}\right), \sum_{b=1}^{B}\mathbf{k}^b m^b_i \rangle + n_i ) \right) ,
      \label{eq:generation_model}
\end{equation}
where $B$ is the number of uniform blur regions, $m_i^b$ is the mask that defines the region $b$, and $\mathbf{k}^b$ is the motion kernel associated with that region. 

The non-uniform motion blur model represented by \cref{eq:generation_model} uses a spatially varying blur kernel, where the kernel applied to each pixel is determined by the object (or background) to which that pixel belongs. This degradation model reflects that moving objects in a dynamic scene naturally produce different blur kernels due to variations in their speed and direction.

\subsection{Synthetic Dataset Generation}\label{subsec:DatasetGen}

We generate a set of blurry/sharp image pairs using the degradation model defined by~\eqref{eq:generation_model}, and a generator of realistic camera-shake trajectories~\citep{gavant2011physiological,delbracio2015removing}. We refer to the Supplementary Material for details on the trajectories and kernels generation process.

\subsubsection{Procedure}\label{subsubsection:generation_procedure} 

Given a random sharp image $\mathbf{u}$ and corresponding objects' segmentation masks, we apply separate kernels to the background and to each segmented object (cf. \cref{fig:dataset_generation}). To ensure smooth transitions between regions, we convolve each object mask with its corresponding kernel before combining them. This blending process creates a more realistic soft transition, allowing multiple objects to contribute to a single pixel's value, unlike the unnatural abrupt changes of hard transitions. %
This procedure aims to provide the network with examples of patches containing multiple motion blurs while avoiding hard transition artifacts at the objects' boundaries. 

Prior to convolution with their respective kernels, we initiate the process by transforming the image to the photon domain. Additionally, we incorporate an illumination augmentation step, which accounts for sensor saturation and variations in lighting conditions. Subsequently, we convolve each image segment with a randomly chosen kernel, followed by a weighted fusion of these segments based on their associated masks. Finally, we convert the result back to the pixel domain. The pseudo-code of the generation procedure is presented in \cref{alg:SyntheticDataset}. 

By implementing this methodology, we generated diverse training datasets, and conducted a thorough analysis to evaluate the influence of the following factors on the outcomes.

\paragraph{``Dotted'' kernels} 

\begin{figure}[t!]
\centering
\small 
\setlength{\tabcolsep}{2pt}

  \begin{tabular}{cccc}
    1000 points & 25 points   &  15 points & 8 points \\   %
    \includegraphics[width=0.11\textwidth]{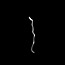} &
    \includegraphics[width=0.11\textwidth]{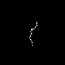} &
    \includegraphics[width=0.11\textwidth]{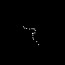}   &
    \includegraphics[width=0.11\textwidth]{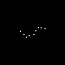} \\
    \hline
    \includegraphics[width=0.11\textwidth]{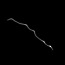} &
    \includegraphics[width=0.11\textwidth]{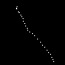} &
    \includegraphics[width=0.11\textwidth]{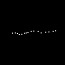}   &
    \includegraphics[width=0.11\textwidth]{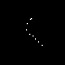} \\
    \hline
    \includegraphics[width=0.11\textwidth]{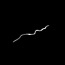} &
    \includegraphics[width=0.11\textwidth]{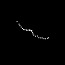} &
    \includegraphics[width=0.11\textwidth]{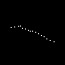}   &
    \includegraphics[width=0.11\textwidth]{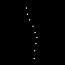} \\
    \end{tabular}
  \caption{The kernel generator ~\citep{gavant2011physiological,delbracio2015removing} produces camera shake trajectories that can be subsampled  to mimic the motion blur kernels responsible for the \textit{ghosting effect} present in synthetic blurry images generated from high-speed videos. \label{fig:discontinuos_kernels}}
\end{figure}

The limited generalization observed in deblurring networks trained using the GoPro dataset is often attributed to the discontinuities present in the blur kernels, resulting in a \textit{ghosting effect} within the blurry images \citep{m_Tran-etal-CVPR21, rim_2022_ECCV, zhao2023representing}.  This raises the question of whether these networks can accurately restore motion kernels that were not encountered during their training. To answer this question, we generated different sets of discontinuous kernels by subsampling 3D motion trajectories provided by the camera trajectories generator \cite{gavant2011physiological}. Examples of the kernels obtained are shown in \cref{fig:discontinuos_kernels}.

\paragraph{Non-uniformity of the blur kernel field} One of the main features of our data generation methodology involves modeling non-uniform blur, wherein distinct motion blurs are applied to individual objects within segmented scenes. The scenes were segmented using Mask-CNN \citep{wei2018mask}.

\begin{figure}
    \centering
\setlength{\tabcolsep}{1pt}
\begin{tabular}{cccc}
& $t_{exp}=0.25s$ & $t_{exp}=0.5s$ & $t_{exp}=1s$  \\
\rotatebox{90}{\hspace{0.2cm} $F=500$} & \includegraphics[width=0.3\linewidth]{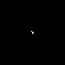} &
\includegraphics[width=0.3\linewidth]{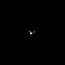} &
\includegraphics[width=0.3\linewidth]{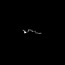} \\
\rotatebox{90}{\hspace{0.2cm} $F=1000$} & \includegraphics[width=0.3\linewidth]{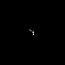} &
\includegraphics[width=0.3\linewidth]{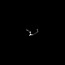} &
\includegraphics[width=0.3\linewidth]{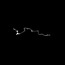} \\
\end{tabular}
    \caption{The generated kernels' shape and length are a function of exposure time ($t_{exp}$) and focal length ($F$), which together determine how the motion path is projected onto the image. Exposure time dictates the overall length of the path. Focal length, for a given exposure time, controls the magnification of that path; higher focal lengths (greater zoom) reveal finer details.}
    \label{fig:kernels_texp_F}
\end{figure}

\paragraph{Kernel size and shape} Intuitively, the larger the kernel support, the more cases the network can handle. However, as the degradation becomes more severe, the restoration problem becomes increasingly complex due to the destruction of most image structures. The blur kernel generator is governed by two parameters: the exposure time (texp) and the image focal length (F). A higher exposure time creates more prominent and curved kernels. 
Large kernels that are not as serpentine can be generated by increasing the focal length while maintaining a constant exposure time. \cref{fig:kernels_texp_F} illustrates exposure time, and the focal length influence in the generated kernels.

\paragraph{Camera Response Function}

While the influence of the Camera Response Function (CRF) on deblurring performance was extensively explored by \cite{tai2013nonlinear, anger2018modeling} prior to the advent of deep learning, it has largely been overlooked in recent times. In our study, we thoroughly investigate the significance of the CRF to gain insights into the generalization performance of deblurring networks. In particular, we asses two families of CRF: $\gamma$-functions  described in \cref{eq:gamma_function} and the exponential function defined as
\begin{equation}
    g(x)=1-\exp(-ax),
    \label{eq:crf_exponential}
\end{equation}
for an exponential decay factor $a > 0$. 

\paragraph{Saturated pixels} We compare several \textit{data augmentation} procedures to asses how end-to-end motion deblurring networks cope with this major challenge. 

\begin{itemize}
    \item[] \textit{Multiplicative augmentation.} In this approach, sharp image pixels undergo multiplication by a random factor within the [1, 3] range prior to applying the convolutions. This augmentation simulates an increase in scene luminance. Rather than directly multiplying each color channel by this factor, we initially convert the image from the \textit{rgb} color space to the \textit{hsv} space and multiply the luminance channel. Subsequently, we convert it back to the \textit{rgb} space. While both alternatives yield similar quantitative results, multiplying the luminance channels generates more realistic images.
    \item[] \textit{Multiplicative clipped augmentation.} This augmentation is similar to the previous one, the only difference being that we clip the scene values to the [0, 1] range before convolving with the blur kernels. The clipping simulates the loss of information that occurs when generating blurry images through frame averaging.
    \item[] \textit{Random Streaks.}  This technique simulates streaks that appear in blurry images when pinpoint lights are present in the scene. We randomly select small areas within the image and multiply the pixel values in those regions by a random factor within the [0, 8] range. 
    \item[] \textit{Saturation streaks.} In contrast to \textit{multiplicative augmentation}, this method exclusively multiplies saturated pixels by a random factor in the range [1,5], leaving non-saturated pixels unaltered.
\end{itemize}

\paragraph{Dataset diversity} The proposed approach offers the distinct advantage of generating a virtually limitless array of image pairs, setting it apart from presently employed training datasets that exhibit constraints in terms of both size and diversity. Nonetheless, to underscore the simplicity and practicality of the proposed procedure, we confine ourselves to generating blurry images exclusively from the GoPro dataset. %
In the experiments, we generated ten blurry images for each sharp image to investigate the dataset's generalization performance.

\begin{algorithm}
\caption{Dataset Generation}
\begin{algorithmic}
\State \textbf{Input}
\State $\mathbf{u},\{\mathbf{k}\}$, $\gamma$ \Comment{Sharp Image, dataset of kernels and $\gamma$ factor}
\Procedure{BlurImage}{$\mathbf{u},\{\mathbf{k}\}, \gamma$}
\State $\mathbf{k}_u, \mathbf{m}_u = $ [\,],[\,] \Comment{List of kernels and masks initialization}
\State $\mathbf{k}_u.append(Random(\{\mathbf{k}\}))$\Comment{Background kernel added to the list}
\State $\mathbf{m}_u.append(ones(size(\mathbf{u})))$\Comment{Background mask initialization}
\For{$\mathbf{m}\in SegmentedObjectMasks(\mathbf{u})$}
    \State $\mathbf{k} =Random(\{\mathbf{k}\})$ \Comment{Object kernel} 
    \State $\mathbf{m} = \mathbf{m}*\mathbf{k} $\Comment{Smooth mask}
    \State $\mathbf{m}_u[0] = \mathbf{m}_u[0] -\mathbf{m}$ \Comment{Update background mask}
    \State $\mathbf{k}_u.append(\mathbf{k})$ ;  $\mathbf{m}_u.append(\mathbf{m})$
\EndFor
\State $\mathbf{u}_{ph} = \mathbf{u}^{\gamma}$   \Comment{Convert to photons}
\State $\mathbf{u}_{ph}$ = $illum\_augment(\mathbf{u}_{ph}$)  %
\State $\mathbf{v}_{ph}=zeros(size(\mathbf{u}))$\Comment{Initialize blurry image}
\For{$\mathbf{k},\mathbf{m} \in \mathbf{k}_u,\mathbf{m}_u$}
    \State $\mathbf{v}_{ph} =  \mathbf{v}_{ph} + \mathbf{m}(\mathbf{k} * \mathbf{u}_{ph})$
\EndFor

\State $\mathbf{u} = \mathbf{u}_{ph}^{1/\gamma}$; \Comment{sharp is converted-back to pixels} 
\State $\mathbf{v} = \mathbf{v}_{ph}^{1/\gamma}$  \Comment{blurry is converted-back to pixels}  
\State \textbf{return $\mathbf{u}$, $\mathbf{v}$} \Comment{Blurred-Sharp pairs}
\EndProcedure
\end{algorithmic}
\label{alg:SyntheticDataset}
\end{algorithm}

\subsection{Limitations of the generated dataset}

Our degradation model assumes that the transitions between the objects and the background are smooth, even though this assumption may not be entirely realistic.  Although more realistic models are available for handling transitions, e.g.  \citep{hasinoff2007layer}, these methods require prior knowledge of the scene's depth, adding complexity to the simulation procedure. In any case, this boundary effect is so local that, in practice, a straightforward smooth transition suffices in most practical scenarios. Moreover, we show quantitatively and through examples in the Supplementary Material that the CRF mismatch effect has a far more substantial impact on achieving accurate deblurring along the image edges, including regions with occlusions, compared to the influence of mask smoothing.   

Our model does not explicitly describe motion blur due to camera movements, such as rotations, zoom-in, or zoom-out. However, experimentally, we have observed that it is feasible to deblur images affected by those movements. We attribute this behavior in part to the convolution being a local operation. Nonetheless, it is worth pointing out that the proposed model exhibits improved performance compared to the scenario where a uniform blur across the entire image is assumed.

\FloatBarrier

\section{Experiments}
\label{sec:experiments}

The primary objective of the experiments reported in this section is to investigate how the training dataset's composition influences deblurring networks' generalization. To this end, we conduct quantitative assessments of their cross-dataset performance and examine the restoration of real images qualitatively in the absence of ground truth data. Our study involves a comparative analysis between well-established training datasets, namely GoPro, RealBlur, and REDS, and several instances of our synthesis procedure. We crafted these instances to isolate and thoroughly examine the impact of specific factors, such as different models of sensor saturation, kernel discontinuities, uniformity of blur, and the Camera Response Function (CRF) effect.  

In line with recent research endeavors~\citep{rim_2022_ECCV, zhong2023real} that prioritize the application of deblurring networks to real-world images over simply achieving superior performance on benchmark datasets, we opt for the SRN architecture proposed by \citet{tao2018scale} as our reference architecture. This choice is guided by its advantages, including faster training, resource efficiency, and superior generalization capabilities compared to most state-of-the-art networks.

Since experiments conducted with the SRN architecture may not be directly transferable to other architectures, for selected instances of our synthesis procedure, we extend our investigations to include state-of-the-art networks such as MIMO-UNet+ \citep{cho2021rethinking} and NAFNet \citep{chen2022simple}. We emphasize that we intend not to compare the networks but to dissect the factors contributing to their generalization performance in real-world motion blur.

\subsection{Benchmark datasets}

\subsubsection{Synthetic datasets}

We analyze the cross-dataset performance on datasets generated by averaging frames from high-speed video cameras. Datasets such as GoPro, GoPro ($\gamma=2.2$), and REDS differ in the original video frame rate, in whether frame interpolation is applied, and in the CRF that is used. In addition, we evaluate the performance on the 100 non-uniformly blurred images from Lai's dataset. To synthesize the blurry images, \citet{lai2016comparative} follow the Projective Motion Blur Model (PMBM), a homography-based model for camera shakes that assumes a rotating camera or planar scene. %

\subsubsection{Real datasets}

The K\"{o}hler dataset~\cite{kohler2012recording}  comprises 48 images of 4 posters placed on a wall, taken by a high-precision robot arm. The trajectory followed by the robot arm during the blurry image acquisition corresponds to that of 12 human shakes previously recorded. For each blurry image, 167 images were taken as ground-truth candidates by playing back the trajectory in 167 steps and taking one sharp photograph per step. The K\"{o}hler dataset quantifies deblurring algorithms' performance in the simplest case, almost uniform blur and planar scene. Performing well on this dataset is not guaranteed to perform well on more challenging scenarios, but being unable to deblur these images is a clear sign of overfitting.

The RealBlur dataset~\cite{rim_2020_ECCV} is a good benchmark for evaluating the performance of the algorithms in low-light conditions with saturated pixels, which is the most typical scenario for unintended motion blur. Additionally, we show results on a standard dataset of real blurred images without ground truth correspondences proposed by \citet{lai2016comparative}, to analyze the generalization performance of classical motion deblurring algorithms on real images. This dataset comprises 100 blurred images, each captured using different cameras and deliberately selected to encompass a wide array of scenes (both indoor and outdoor), subjects (including objects, faces, text, and landscapes), blur types (including uniform and non-uniform blur resulting from camera shake or object motion), and varying illumination conditions.

\subsection{Influence of the training dataset conformation on the motion deblurring results}

\subsubsection{``Dotted'' kernels}

\begin{figure*}[ht!]
  \centering
  \small
  \setlength{\tabcolsep}{1pt}
  \setlength{\fboxrule}{1pt}
  \setlength{\fboxsep}{0pt} %
  \begin{tabular}{*{20}{c}}
    Blurry & Original GoPro & SBDD\_U(8 points)  & SBDD\_U(15 points) &   SBDD\_U(1000 points) \\
    \includegraphics[trim=280 160 0 0, clip,width=0.19\textwidth]{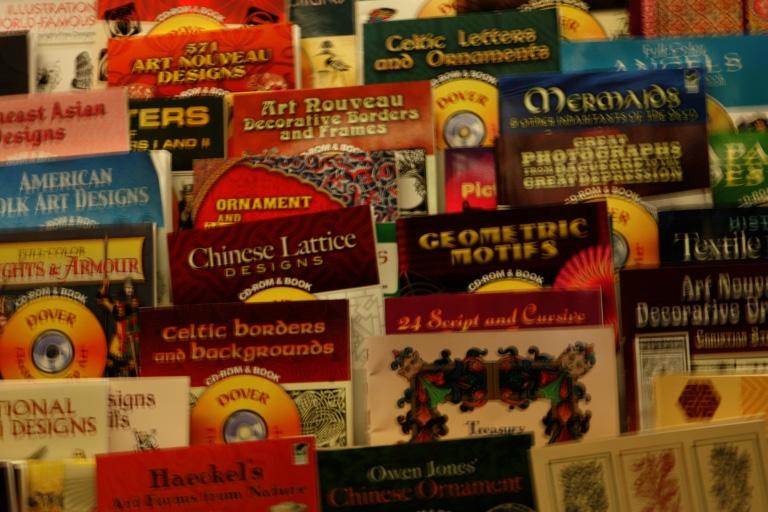} &
       \includegraphics[trim=280 160 0 0, clip,width=0.19\textwidth]{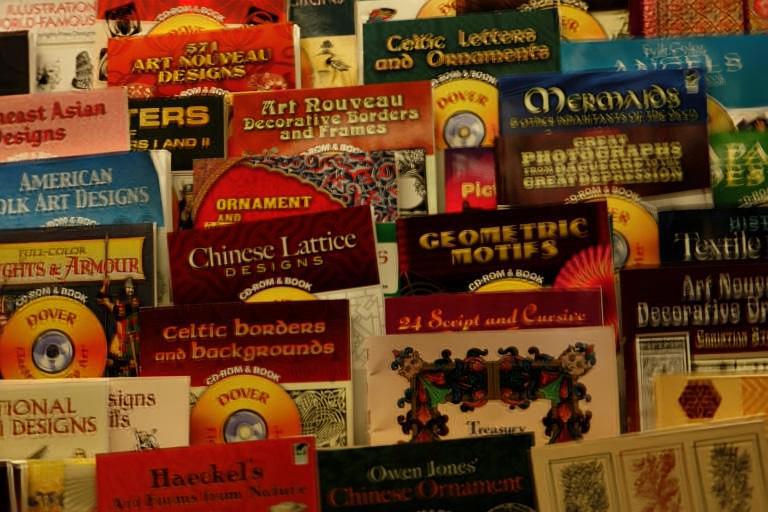} &
    \includegraphics[trim=280 160 0 0, clip,width=0.19\textwidth]{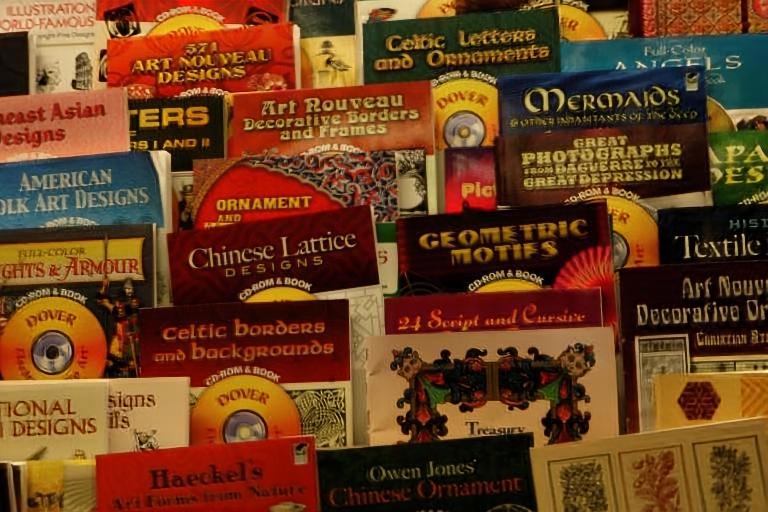} 
    & 
    \includegraphics[trim=280 160 0 0, clip,width=0.19\textwidth]{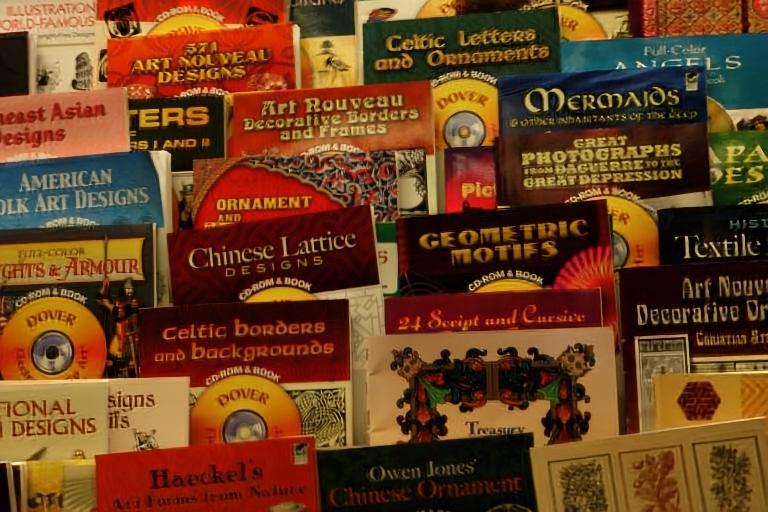}
    &
    \includegraphics[trim=280 160 0 0, clip,width=0.19\textwidth]{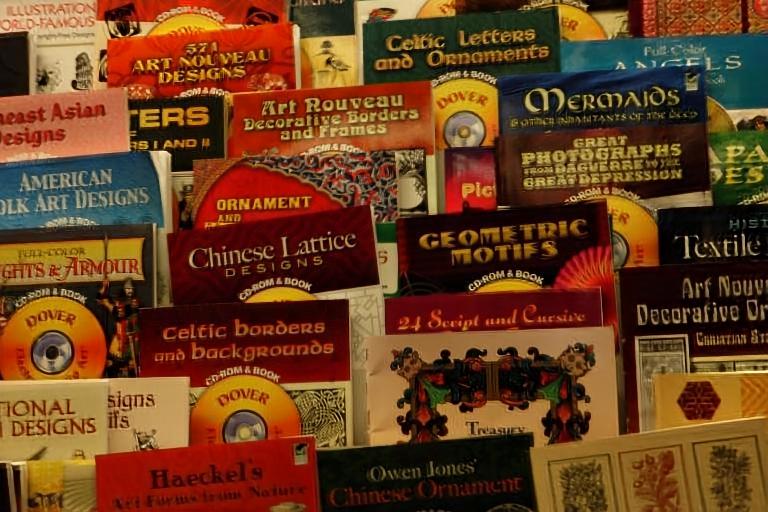}   \\
    \includegraphics[trim=250 100 300 140, clip,width=0.19\textwidth]{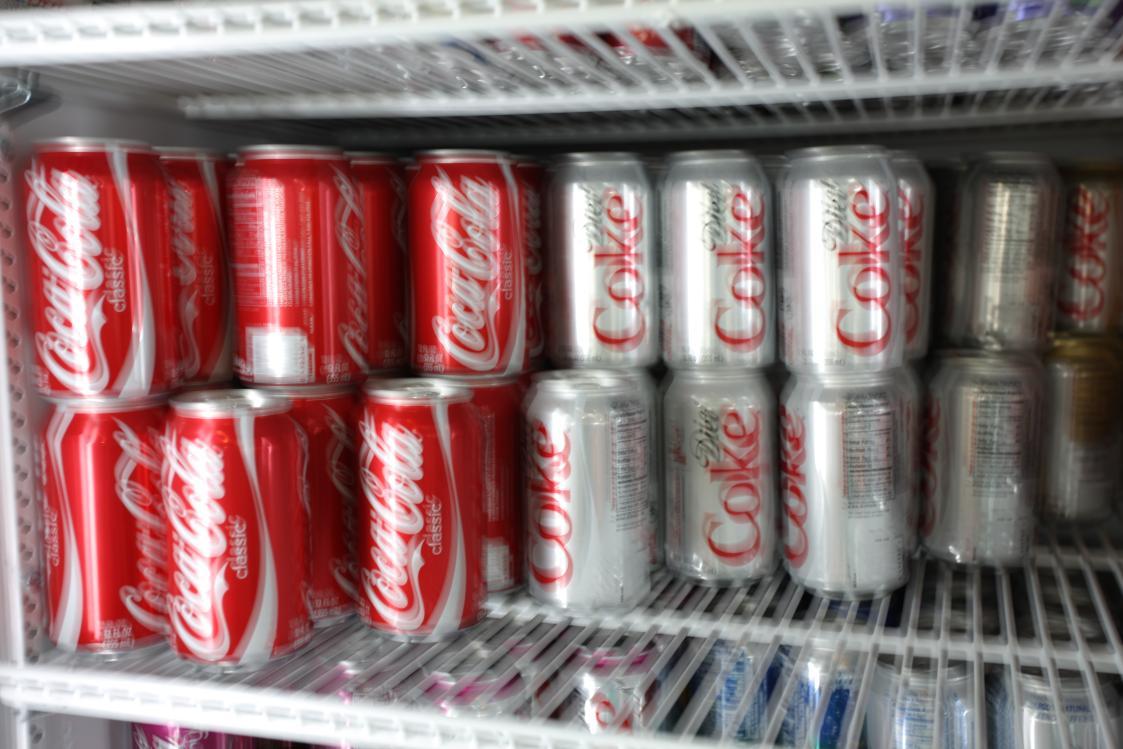} &
       \includegraphics[trim=250 100 300 140, clip,width=0.19\textwidth]{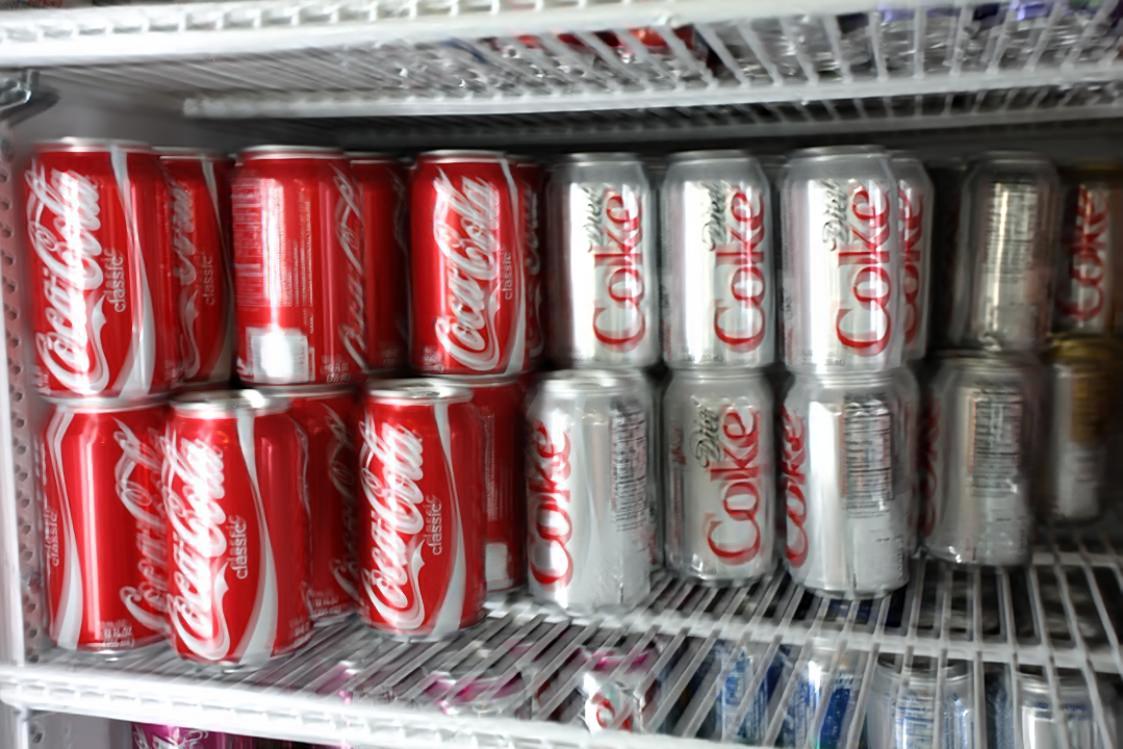} &
    \includegraphics[trim=250 100 300 140, clip,width=0.19\textwidth]{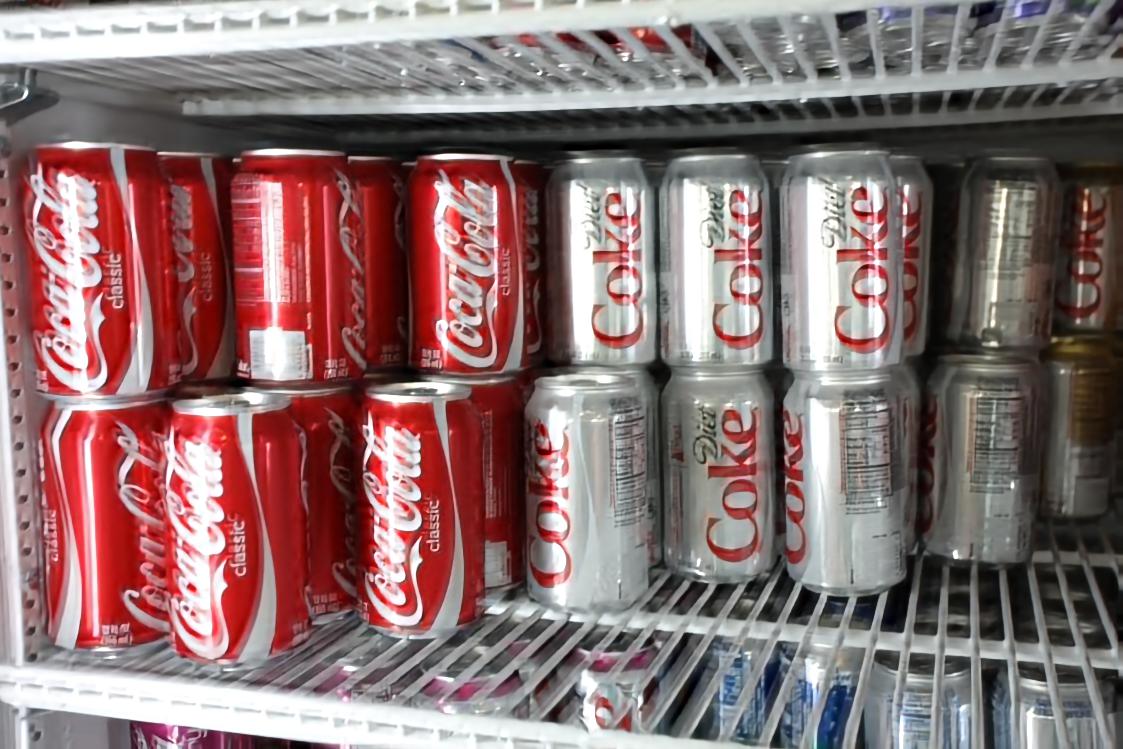} 
    & 
    \includegraphics[trim=250 100 300 140, clip,width=0.19\textwidth]{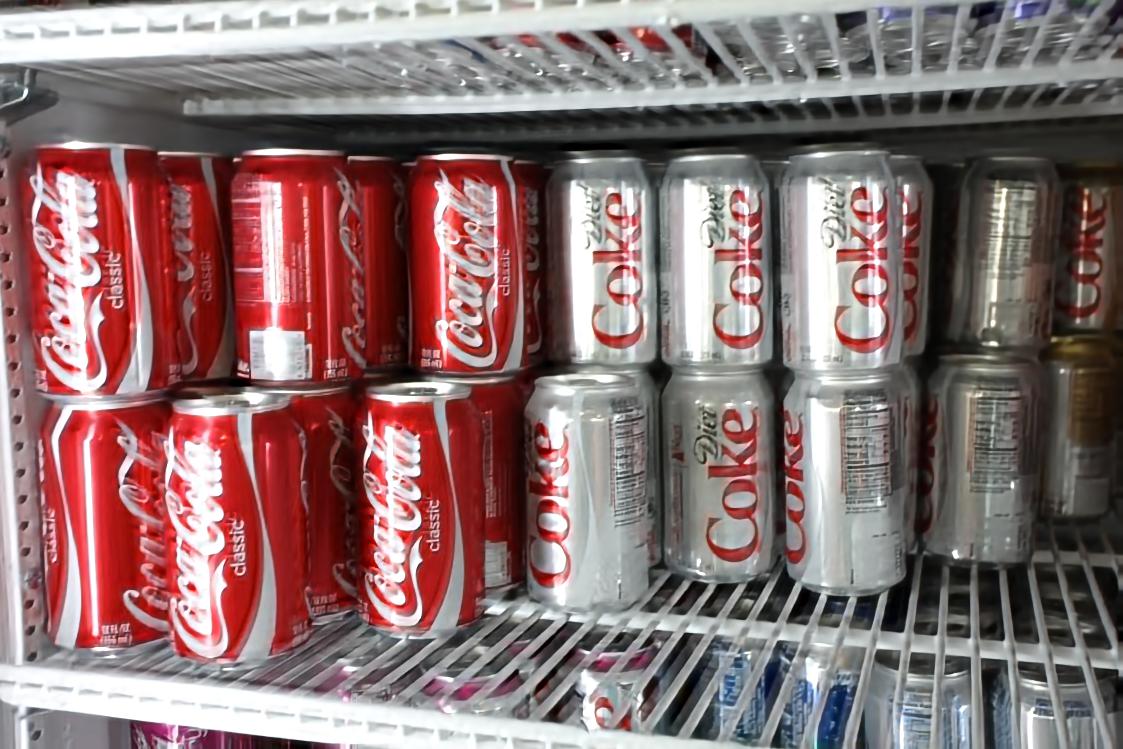}
    &
    \includegraphics[trim=250 100 300 140, clip,width=0.19\textwidth]{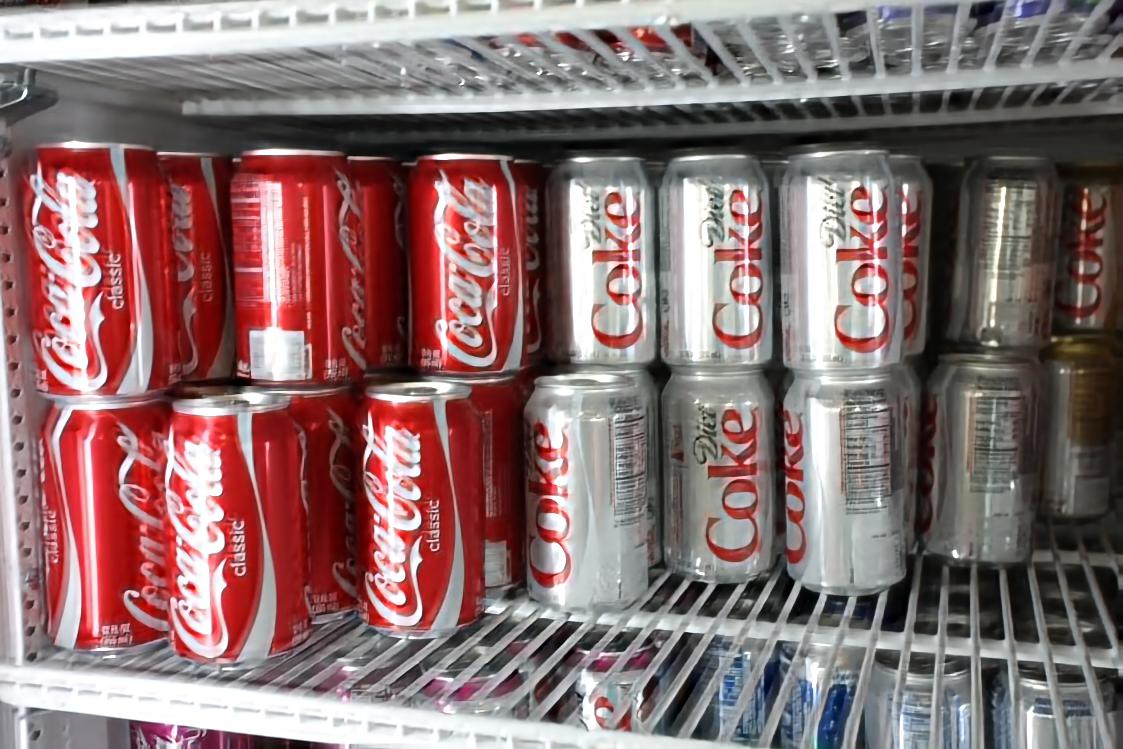}   \\
    \includegraphics[trim=200 0 40 540, clip,width=0.19\textwidth]{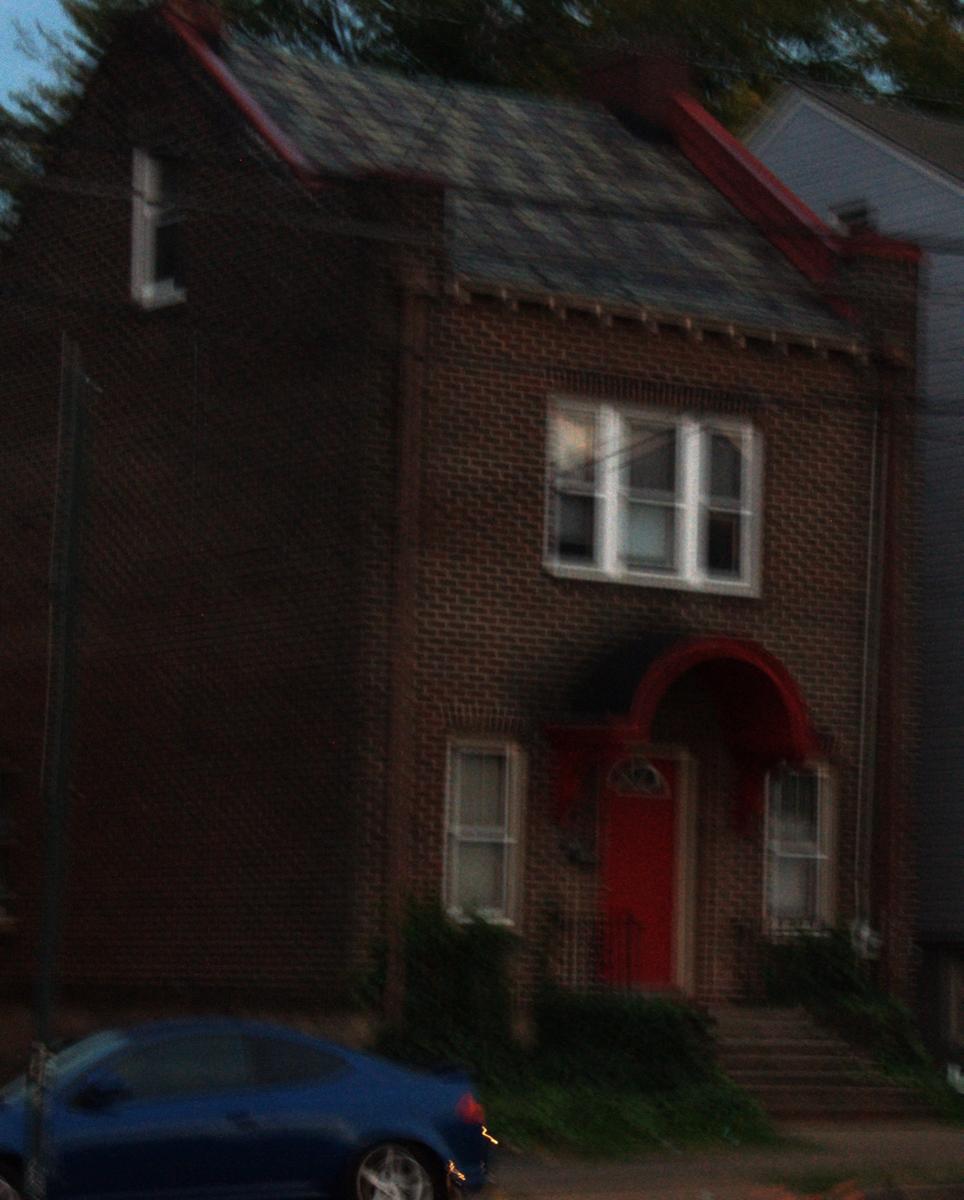} &
    \includegraphics[trim=200 0 40 540, clip,width=0.19\textwidth]{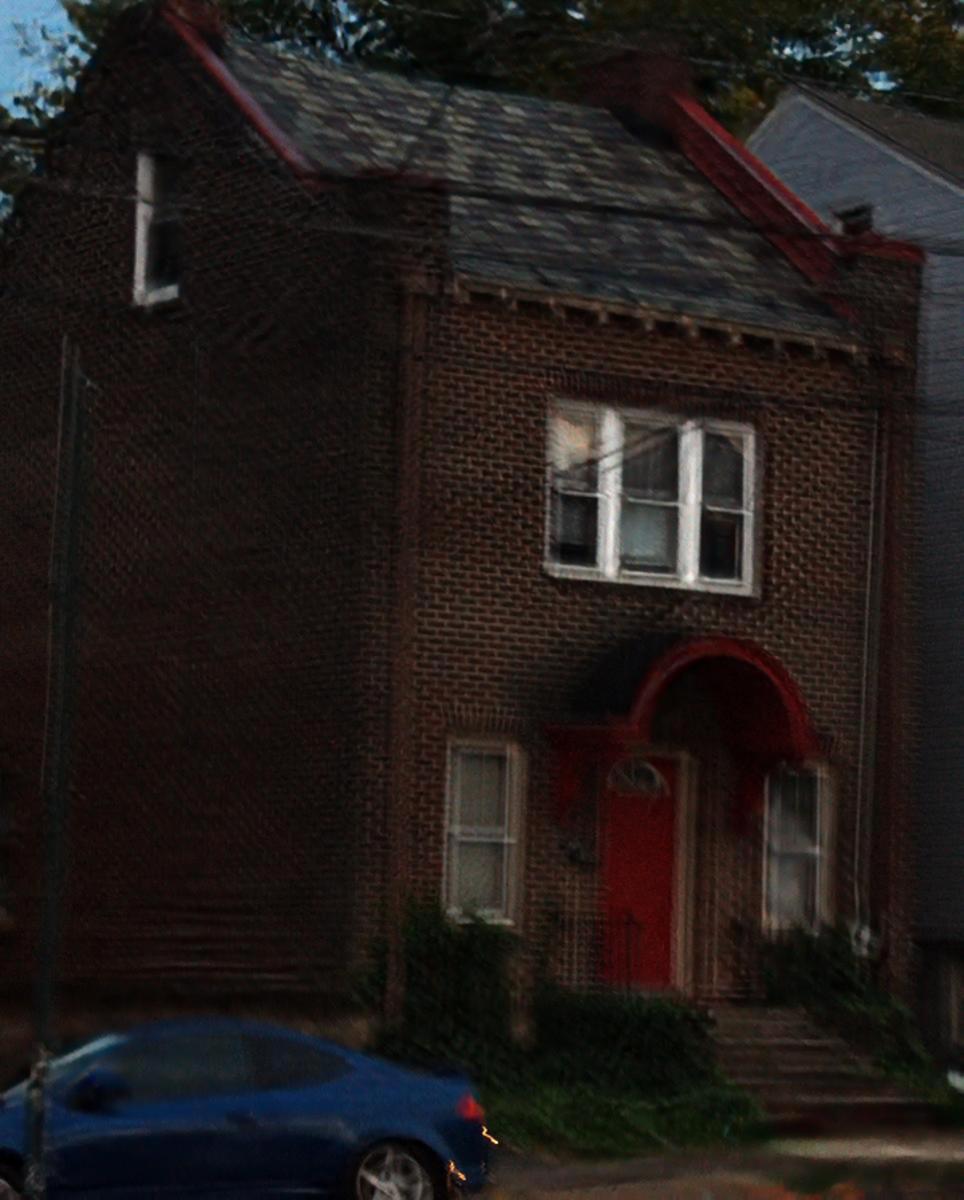} &
    \includegraphics[trim=200 0 40 540, clip,width=0.19\textwidth]{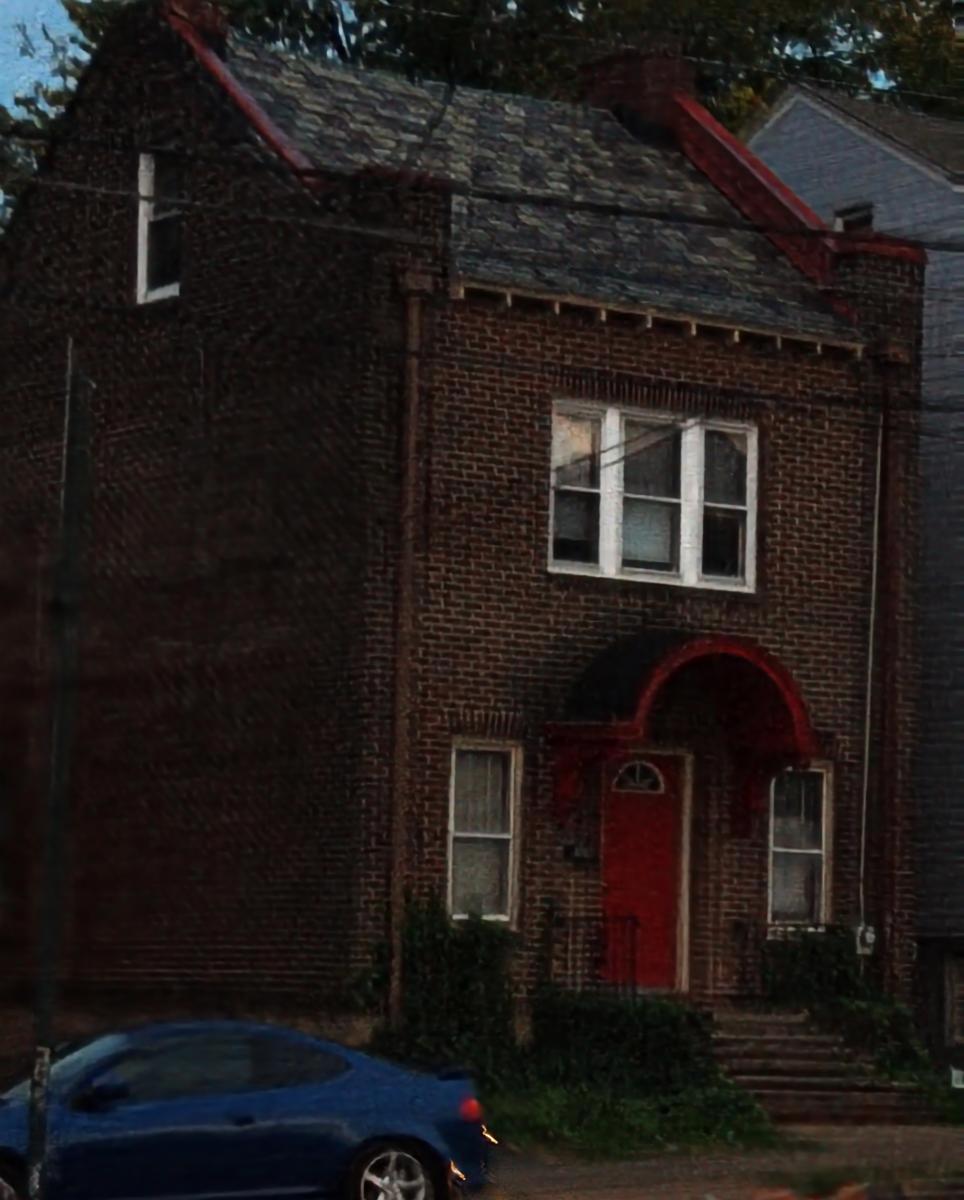} 
    & 
    \includegraphics[trim=200 0 40 540, clip,width=0.19\textwidth]{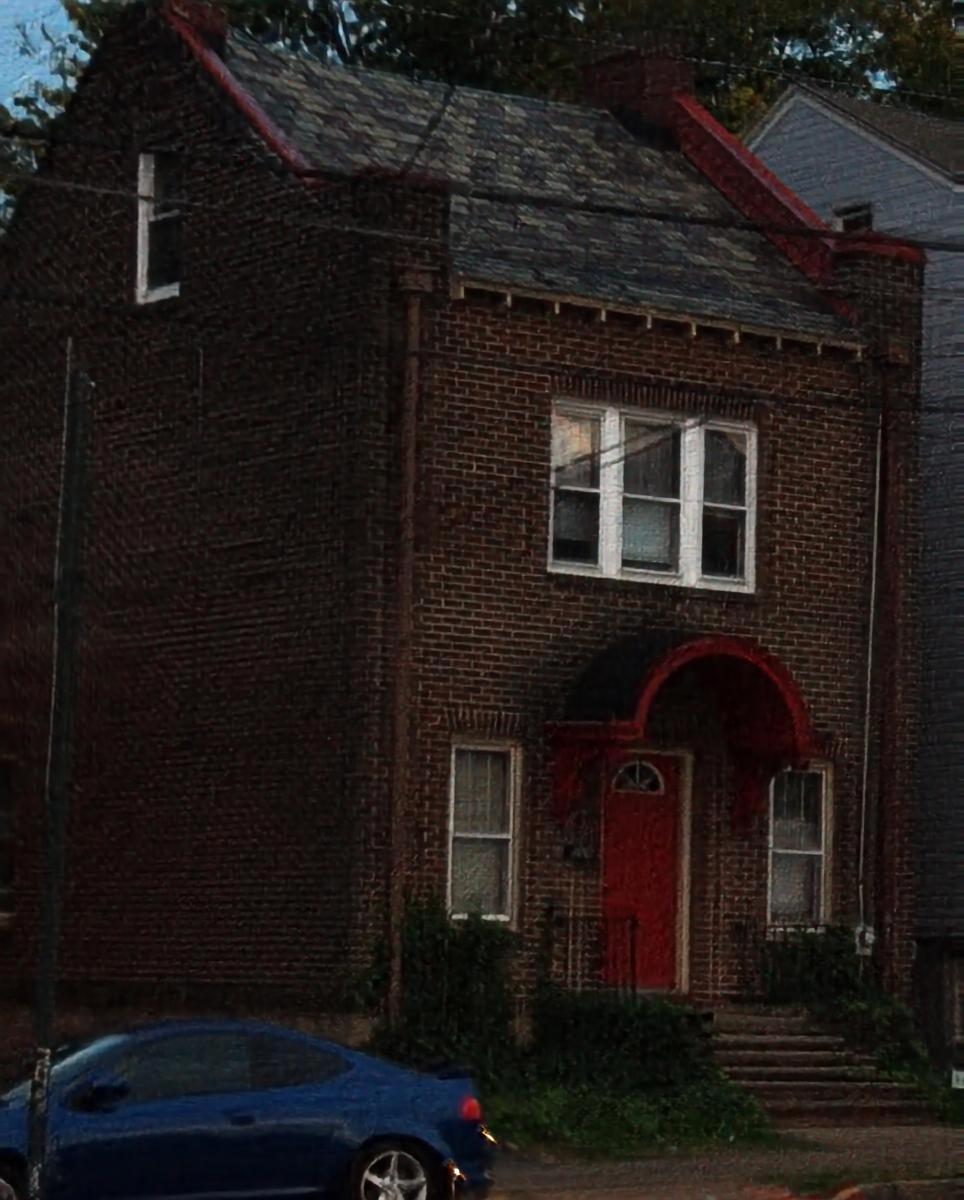}
    &
    \includegraphics[trim=200 0 40 540, clip,width=0.19\textwidth]{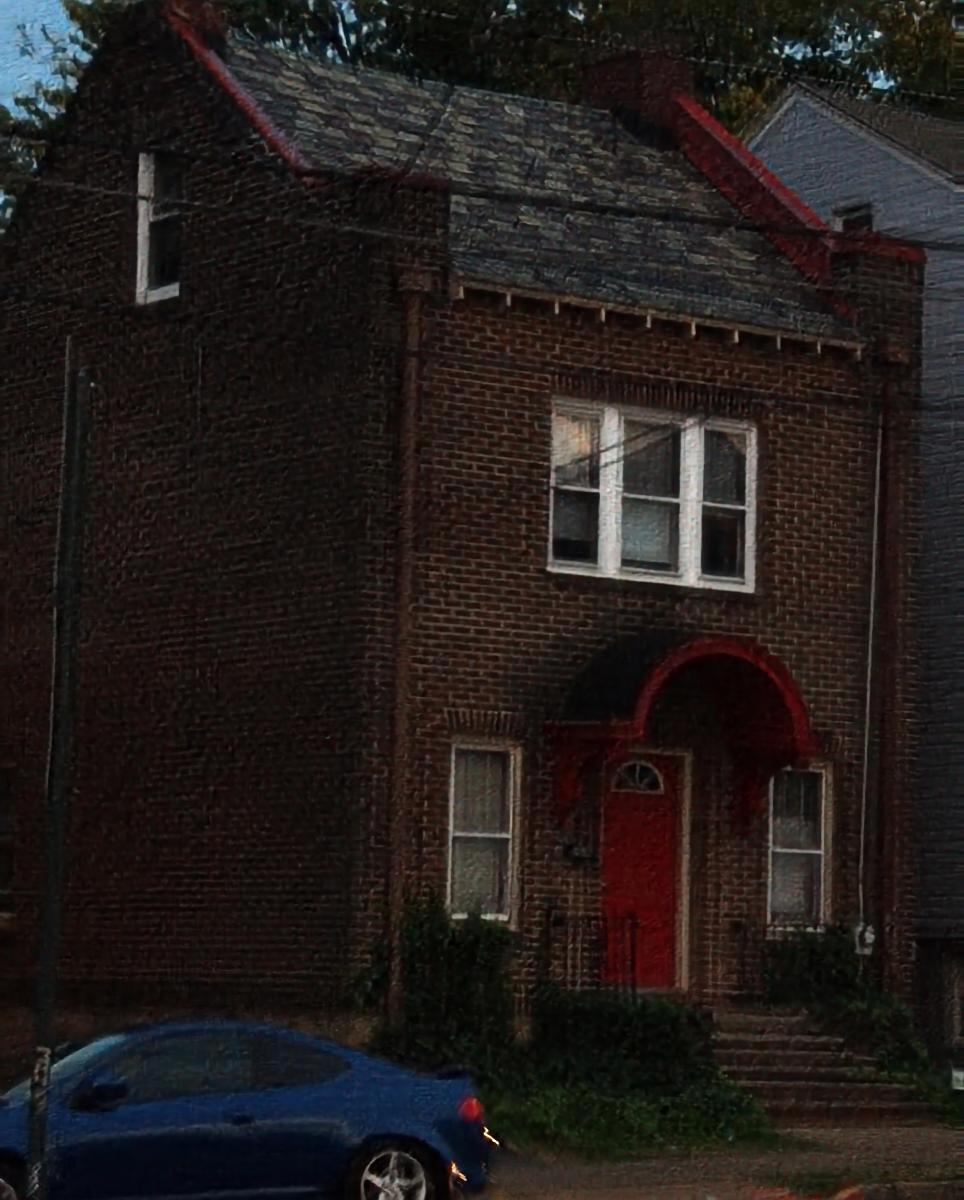}   \\
    
  \end{tabular}
  \caption{The primary factor behind the poor generalization of networks trained on the GoPro dataset is not the utilization of discontinuous kernels. When employing ``dotted'' kernels (8 or 15 points in the trajectory) during training, the obtained results closely resemble those achieved with continuous kernels (1000 points). All the cases yield superior results than training with the original GoPro dataset. Despite being one of the networks that generalize better when training with the GoPro dataset, SRN faces challenges when handling images containing saturated pixels. This limitation may arise from the fact that the blurry frames are generated through the averaging of saturated frames. In contrast, when the convolution model generates blurry frames, it demonstrates improved performance on saturated pixels, even without the explicit modeling of saturation. Best viewed in electronic format.  }
  \label{fig:discontinuous_kernel_effect}
\end{figure*}

We generated different training sets of uniformly blurred images by subsampling the 3D motion trajectories with different numbers of points. We denote the datasets \emph{SBDD\_U} followed by the number of trajectory points. We trained an SRN network with each set. As shown in \cref{tab:discointinuos}, the results on the GoPro test set were better when training with discontinuous kernels, while on the K\"{o}hler benchmark results remained almost unchanged. The former is due to the presence of ``dotted'' kernels in the GoPro dataset. The latter suggests that contrary to popular belief, the ``dotted'' kernels may not be the leading cause for the lack of generalization of the networks trained on the GoPro dataset. In fact, all the networks trained in this experiment generalize better than those trained on the original GoPro dataset. This claim is supported quantitatively on the K\"{o}hler test set, and some examples from Lai's dataset \citep{lai2016comparative} are shown in \cref{fig:discontinuous_kernel_effect}. More examples are also provided in the Supplementary Material.

\begin{table}[t]
    \centering
    \caption{Influence of the number of motion trajectory sample points. Results correspond to PSNR/SSIM metrics. Discontinuous kernels yield similar or better results than continuous (1000 points) trajectories. Best \textbf{cross-dataset} results indicated in \textbf{bold}. We alert with \textcolor{red}{red} when the testing set is synthesized like the training set. }
    \setlength{\tabcolsep}{2pt}
    \begin{tabular}{l|l|cc}
       \toprule 
         \multicolumn{2}{c|}{} & \multicolumn{2}{|c}{Testing Sets} \\
         \midrule 
        Exp &  Training Set  & GoPro & Kohler \\ 
        E0  &   GoPro                      & \textcolor{red}{30.72}/\textcolor{red}{0.907}  & 26.90/0.789\\
        E1  &  \emph{SBDD\_U}(1000 points) & 29.30/0.884 &  28.40/0.817  \\ 
        E2  & \emph{SBDD\_U}(25 points)    & 29.59/0.891 &  \textbf{28.43}/\textbf{0.819}  \\ 
        E3  & \emph{SBDD\_U}(15 points)    & \textbf{29.65}/\textbf{0.892} & 28.41/0.817  \\ 
        E4  & \emph{SBDD\_U}(8 points)    & 29.59/0.890 & 28.41/\textbf{0.819}  \\ 
        \bottomrule 
    \end{tabular}
    \label{tab:discointinuos}
\end{table}

\subsubsection{Uniform vs. non-uniform synthetic blur}

\begin{figure*}[ht!]
  \centering
  \small
  \setlength{\tabcolsep}{1pt}
  \setlength{\fboxrule}{1pt}
  \setlength{\fboxsep}{0pt} %
  \begin{tabular}{*{20}{c}}
    \multicolumn{5}{c}{Blurry} & \multicolumn{5}{c}{Ana-Syn\cite{kaufman2020deblurring}}&  \multicolumn{5}{c}{SRN + \emph{SBDD\_U} (Ours)} &   \multicolumn{5}{c}{SRN + \emph{SBDD\_NU} (Ours)} \\
    \multicolumn{5}{c}{\includegraphics[width=0.22\textwidth]{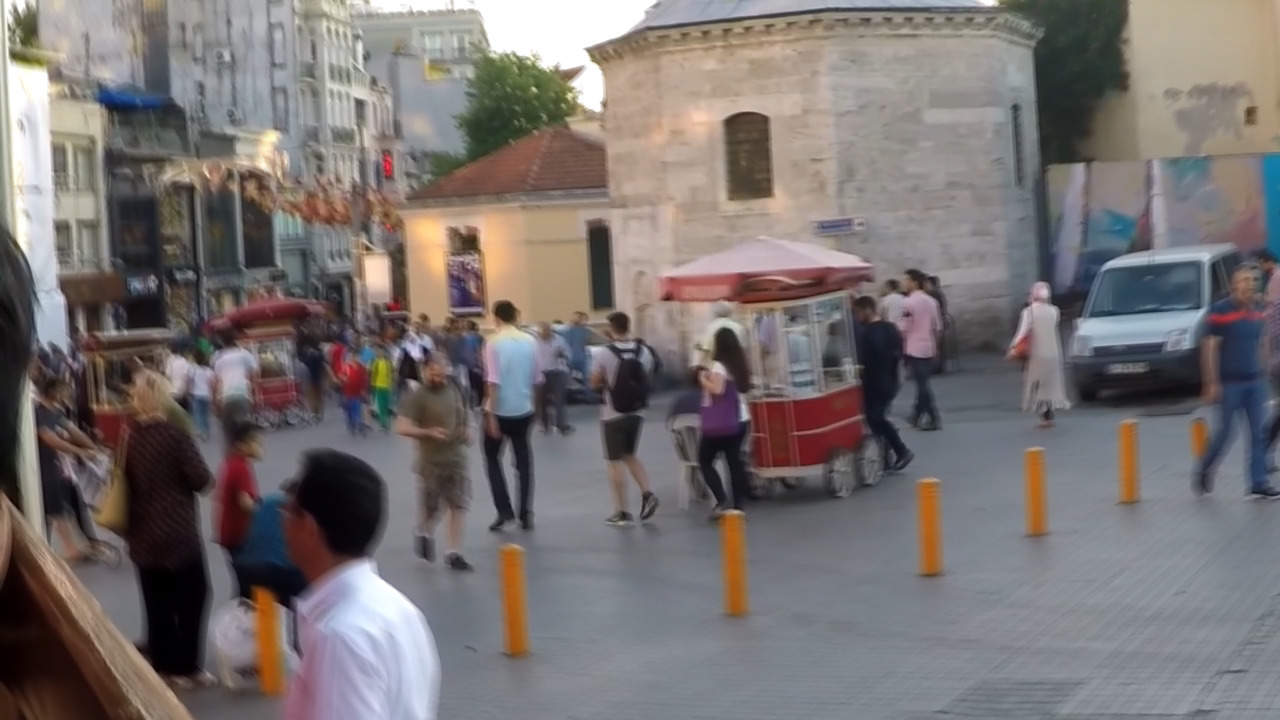} }
    & 
    \multicolumn{5}{c}{\includegraphics[width=0.22\textwidth]{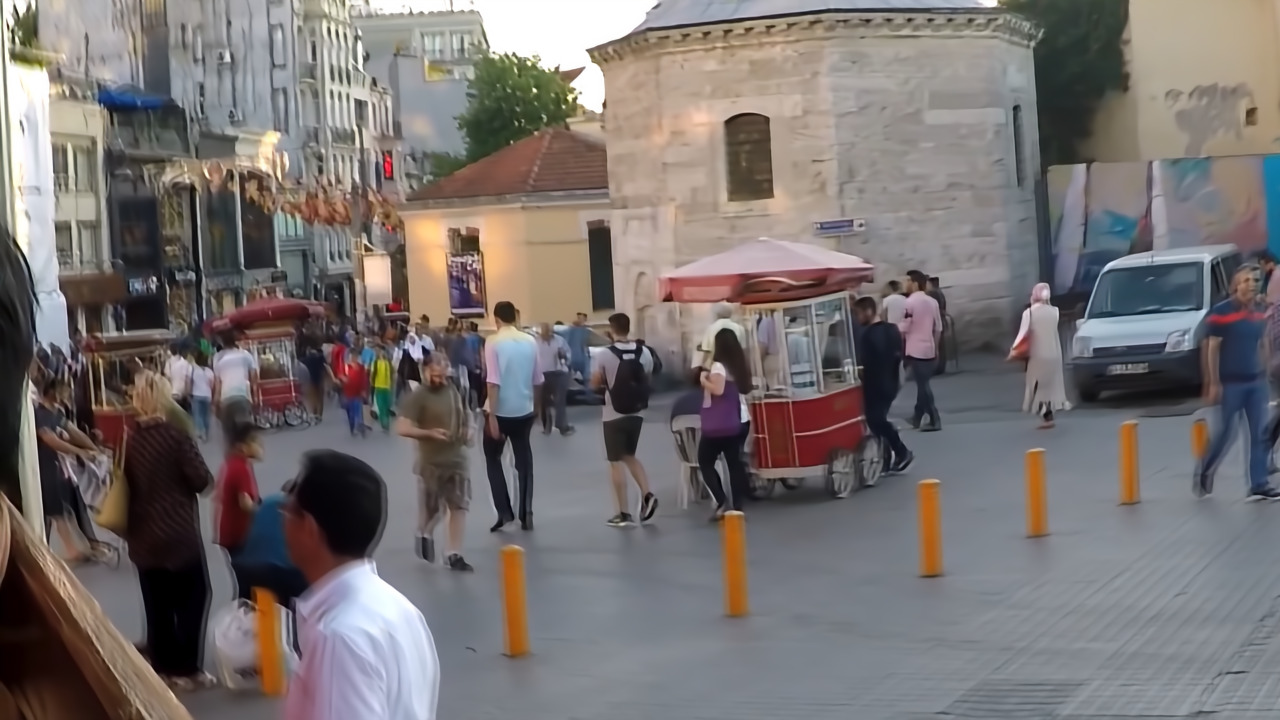}} 
    &
    \multicolumn{5}{c}{\includegraphics[width=0.22\textwidth]{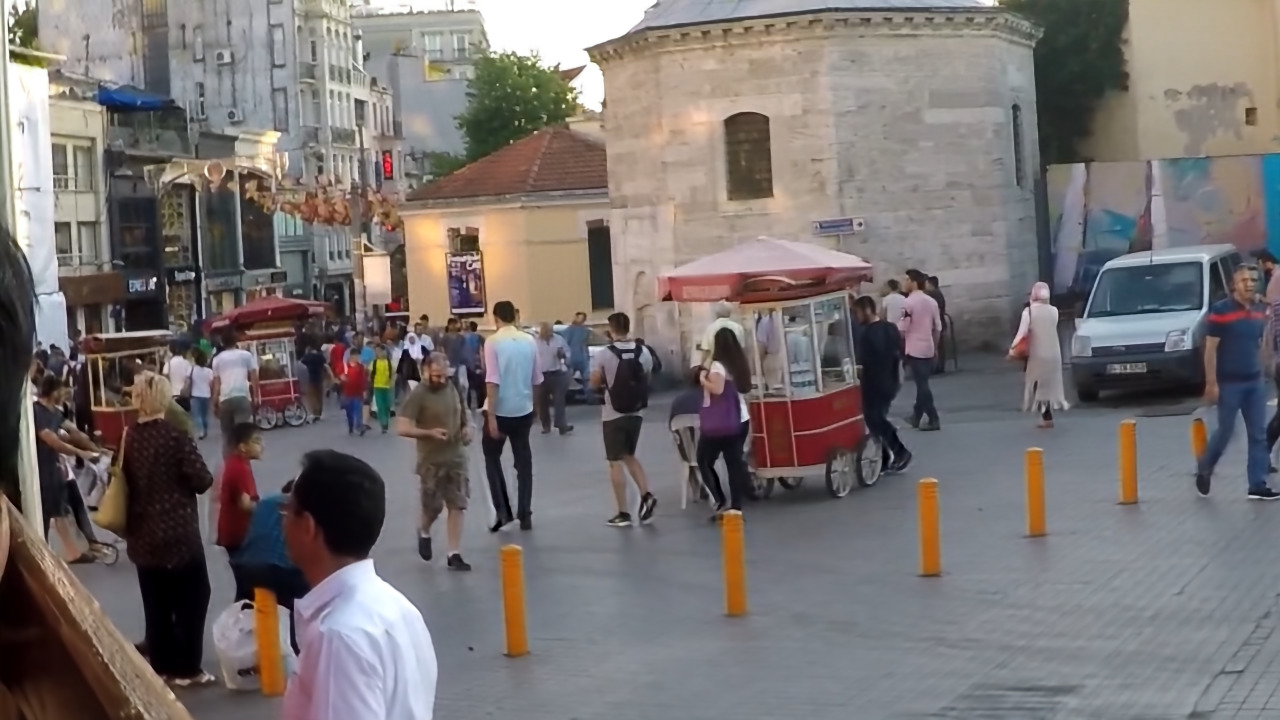}}  &
    \multicolumn{5}{c}{\includegraphics[width=0.22\textwidth]{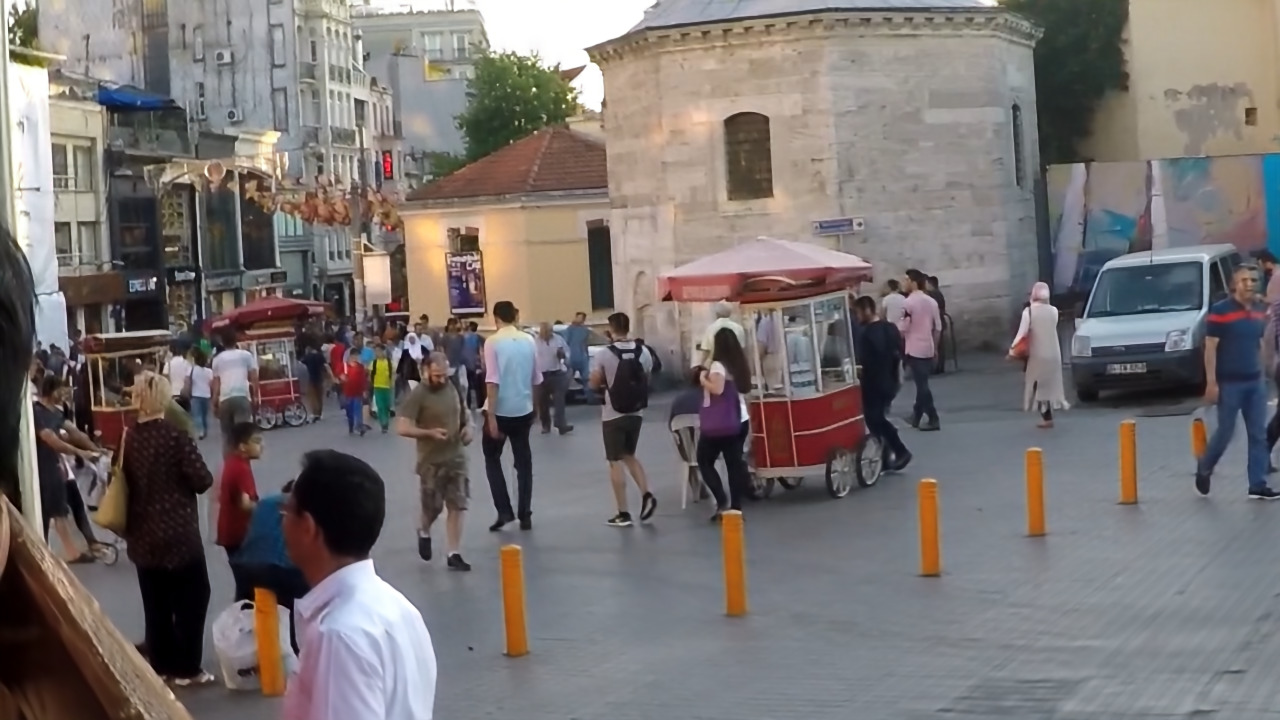}} \\
    \includegraphics[trim=400 140 800 450, clip,height=1.2cm]{imgs/Blurry/GOPR0384_11_00_000057.jpg} & 
    \includegraphics[trim=380 250 800 320, clip,height=1.2cm]{imgs/Blurry/GOPR0384_11_00_000057.jpg}  &
    \includegraphics[trim=470 170 730 420, clip,height=1.2cm]{imgs/Blurry/GOPR0384_11_00_000057.jpg} & 
    \includegraphics[trim=600 180 620 440, clip,height=1.2cm]{imgs/Blurry/GOPR0384_11_00_000057.jpg}  &
    \includegraphics[trim=700 180 540 460, clip,height=1.2cm]{imgs/Blurry/GOPR0384_11_00_000057.jpg} & 
    \includegraphics[trim=400 140 800 450, clip,height=1.2cm]{imgs/AnaSyn/GOPR0384_11_00_000057_deblurred.jpg} &
    \includegraphics[trim=380 250 800 320, clip,height=1.2cm]{imgs/AnaSyn/GOPR0384_11_00_000057_deblurred.jpg} 
    &     \includegraphics[trim=470 170 730 420, clip,height=1.2cm]{imgs/AnaSyn/GOPR0384_11_00_000057_deblurred.jpg} 
    & 
    \includegraphics[trim=600 180 620 440, clip,height=1.2cm]{imgs/AnaSyn/GOPR0384_11_00_000057_deblurred.jpg}  
    &      \includegraphics[trim=700 180 540 460, clip,height=1.2cm]{imgs/AnaSyn/GOPR0384_11_00_000057_deblurred.jpg} 
    & 
    \includegraphics[trim=400 140 800 450, clip,height=1.2cm]{imgs/SRN_with_GoPro_uniform_ks65_texp05_F1000_ill_aug_2up_n10_gf1/GOPR0384_11_00_000057_PSNR_31.67150.jpg} &
    \includegraphics[trim=380 250 800 320, clip,height=1.2cm]{imgs/SRN_with_GoPro_uniform_ks65_texp05_F1000_ill_aug_2up_n10_gf1/GOPR0384_11_00_000057_PSNR_31.67150.jpg} &
    \includegraphics[trim=470 170 730 420, clip,height=1.2cm]{imgs/SRN_with_GoPro_uniform_ks65_texp05_F1000_ill_aug_2up_n10_gf1/GOPR0384_11_00_000057_PSNR_31.67150.jpg}  &
    \includegraphics[trim=600 180 620 440, clip,height=1.2cm]{imgs/SRN_with_GoPro_uniform_ks65_texp05_F1000_ill_aug_2up_n10_gf1/GOPR0384_11_00_000057_PSNR_31.67150.jpg}   &      
    \includegraphics[trim=700 180 540 460, clip,height=1.2cm]{imgs/SRN_with_GoPro_uniform_ks65_texp05_F1000_ill_aug_2up_n10_gf1/GOPR0384_11_00_000057_PSNR_31.67150.jpg} 
    &   
    \includegraphics[trim=400 140 800 450, clip,height=1.2cm]{imgs/SRN_with_GoPro_non_uniform_mob5_ks65_texp05_F1000_ill_aug_2up_n10_gf1/GOPR0384_11_00_000057_PSNR_31.77346.jpg} &
    \includegraphics[trim=380 250 800 320, clip,height=1.2cm]{imgs/SRN_with_GoPro_non_uniform_mob5_ks65_texp05_F1000_ill_aug_2up_n10_gf1/GOPR0384_11_00_000057_PSNR_31.77346.jpg} &
    \includegraphics[trim=470 170 730 420, clip,height=1.2cm]{imgs/SRN_with_GoPro_non_uniform_mob5_ks65_texp05_F1000_ill_aug_2up_n10_gf1/GOPR0384_11_00_000057_PSNR_31.77346.jpg}  & 
    \includegraphics[trim=600 180 620 440, clip,height=1.2cm]{imgs/SRN_with_GoPro_non_uniform_mob5_ks65_texp05_F1000_ill_aug_2up_n10_gf1/GOPR0384_11_00_000057_PSNR_31.77346.jpg}   &      
    \includegraphics[trim=700 180 540 460, clip,height=1.2cm]{imgs/SRN_with_GoPro_non_uniform_mob5_ks65_texp05_F1000_ill_aug_2up_n10_gf1/GOPR0384_11_00_000057_PSNR_31.77346.jpg}  \\
  \end{tabular}
  \caption{Evaluation of synthetic blur datasets in a non-uniformly blurred image from GoPro \citep{Nah_2017_CVPR}. Methods ordered according to performance (PSNR/ SSIM) for this image: Analysis-Synthesis (27.28/0.850), SRN+SBDD\_U (31.67/ 0.922), SRN+SBDD\_NU (31.77/ 0.9195). The suffixes U and NU stand for spatially uniform and non-uniform blur, respectively. Note that training with databases that simulate uniform blur performs well, except at the boundaries of moving objects. 
  \emph{SBDD\_U} outperforms Analysis-Synthesis because, unlike Analysis-Synthesis, SRN does not assume the blur is uniform across the whole image. Training SRN with the proposed \emph{SBDD\_NU} dataset performs better in the boundaries of moving objects, as the network learns from patches where multiple motion blur kernels coexist.}
  \label{fig:non_uniform_eval_GoPro}
\end{figure*}

In this section, we validate our approach and compare its performance to training with uniformly blurred images when it comes to deblurring non-uniformly blurred images. In the experiments, we denote non-uniformly blurred images as \emph{SBDD\_NU}. From now on, we generate continuous trajectories by sampling 1000 points. We employ two benchmark architectures: Analysis-Synthesis \citep{kaufman2020deblurring}, which assumes uniform blur across the entire image and is trained using uniformly blurred images from Open Images, and SRN, trained with both uniformly and non-uniformly blurred images generated following the proposed procedure. For the non-uniformly blurred dataset, we also explore using masks located at random positions, instead of relying on the segmentation masks provided by Mask CNN \citep{wei2018mask}.  We conduct a quantitative evaluation on two datasets: GoPro~\citep{Nah_2017_CVPR} and Lai's non-uniform dataset~\citep{lai2016comparative}. The quantitative results presented in \cref{tab:res_non_uniform} reveal that training with non-uniform blur is beneficial on the GoPro test set. However, training with uniformly blurred images yields slightly better results on Lai's non-uniform test set, characterized by a much smoothly varying blur kernel field. This observation underscores the effectiveness of SRN, a convolutional network, when dealing with scenarios where the blur can be considered locally uniform. It is noteworthy the contrast with  Analysis-Synthesis \citep{kaufman2020deblurring}, which not only undergoes training with uniformly blurred images but also assumes uniform blur across the entire image. On the other hand, training SRN with the GoPro dataset fails to generalize to Lai's non-uniform test set, mirroring the behavior observed with K\"{o}hler's test set. In \cref{fig:non_uniform_eval_GoPro}, we provide a qualitative illustration that showcases how our approach, in comparison to methods synthesizing uniform blur, achieves significantly superior performance when deblurring small moving objects within a dynamic scene sourced from the GoPro dataset.  %

\begin{table}[h!]
    \caption{ Cross-dataset evaluation (PSNR/SSIM) to quantify the capacity of methods to deal with non-uniform blur. 
        SRN trained with a uniform blur dataset can handle some non-uniform images. Analysis-Synthesis~\cite{kaufman2020deblurring} excels on uniform blurry images but struggles on non-uniform blurry images. Best \textbf{cross-dataset} results indicated in \textbf{bold}. We alert with \textcolor{red}{red} when the testing set is synthesized like the training set. }
    \label{tab:res_non_uniform}
    \centering
    \footnotesize  
    \setlength{\tabcolsep}{0.01pt}
    \begin{tabular}{ll|cc}
     \toprule 
        & & \multicolumn{2}{|c}{Testing Sets} \\
      \cline{3-4} 
       Exp & Arch + Training Sets         & GoPro   & Lai (non.u )  \\
       \midrule 
       E1 & SRN + \emph{SBDD\_U}  & 29.30/0.884   & \textbf{22.04}/\textbf{0.702}  \\
       E5 & SRN + \emph{SBDD\_NU} & \textbf{29.81}/\textbf{0.893}  & 21.96/0.703     \\
       E6 & SRN + \emph{SBDD\_NU-rand-masks} & 29.59/0.887  & 21.83/0.694    \\
       E7 & SRN + GoPro (non-u.)  &  \textcolor{red}{30.72}/\textcolor{red}{0.907}  & 21.25/0.668 \\ 
       \midrule 
       E8 & Ana-Syn + Open Images (u.) \cite{kaufman2020deblurring}&    28.02/0.864  &  21.45/0.695   \\
       \bottomrule  
    \end{tabular}
\end{table}

\subsubsection{Kernels' size and shape}

In the previous experiments, we set $65 \times 65$ pixels as the maximum size for the kernels' support. While this choice may appear somewhat arbitrary, it is driven by the objective of utilizing the largest feasible kernel size that can be effectively used in practice. As demonstrated in the results presented in  \cref{tab:F_texp}, there is a noticeable degradation in performance when either increasing or decreasing the maximum kernel size. In addition to size, the shape of the kernels plays an important role. Since exposure times typically do not exceed 0.5 seconds and objects tend to move in relatively straight trajectories, training with less curved kernels (F=1000) is more effective in practice. Although further optimization may lead to better quantitative results for specific datasets, we used F=1000 and texp=0.5 for subsequent experiments. The primary objective is to demonstrate the feasibility of enhancing results by leveraging \textit{a priori}  knowledge to select a customized set of kernels. Replacing the trajectories generator may be convenient for specific settings since it was designed to simulate camera-shake trajectories.

\begin{table}[t]
    \centering
    \footnotesize 
    \setlength{\tabcolsep}{0.5pt}
    \caption{Influence of blur kernel size and shape. Results correspond to PSNR/SSIM metrics. Training with medium size (65) and simple shape kernels generated with 0.5s exposure time and 1000 focal length yields better results than training with smaller (33), larger (99), or more complex kernels (F=500, texp=1).  }
    \begin{tabular}{l|ccc|cc}
    \toprule 
         & \multicolumn{3}{c|}{\emph{SBDD\_U} kernels parameters} & \multicolumn{2}{|c}{Testing Sets} \\
         \cline{2-6} 
        Exp & F  & Kernel support & Exp. time  & GoPro & Kohler \\ 
        E1 & 500 & 65$\times$65 $px^2$     & 1s & 29.30/ 0.884 &  28.40/ 0.817  \\ 
        E9 & 500 &  33$\times$33 $px^2$ &  1s & 28.59/ 0.870 &  28.07/ 0.810  \\  
        E10 & 500 & 99$\times$99 $px^2$  & 1s & 29.27/ 0.884 &  28.27/ 0.814  \\ 
        E11 & 1000 & 65$\times$65 $px^2$  & 0.5s& \textbf{29.61}/ \textbf{0.892} &  \textbf{28.57}/ \textbf{0.820}  \\ 
        E12 & 1000 & 99$\times$99 $px^2$   & 0.5s & 28.02/ 0.810 &  28.18/ 0.861  \\ 
        \bottomrule 
    \end{tabular}
    \label{tab:F_texp}
\end{table}

\subsubsection{Camera Response Function}

While the GoPro and K\"{o}hler datasets rely on a linear Camera Response Function (CRF), the majority of real-world images undergo $\ gamma$ correction. In \cref{tab:gamma_corrected}, our results highlight a notable performance degradation when training with a $\gamma$-corrected dataset on GoPro and K\"{o}hler. Interestingly, for GoPro ($\gamma$=2.2) and RealBlur, which both employ a non-linear CRF, incorporating $\gamma$-correction leads to improved outcomes. Surprisingly, while the vulnerability of deblurring algorithms to $\gamma$-correction  changes has been known for at least one decade \citep{tai2013nonlinear}, as far as we know, this issue has not been addressed in the context of learning-based image restoration. 

\begin{table}[t]
    \centering
    \setlength{\tabcolsep}{0.2pt}
    \footnotesize 
    \caption{Influence of $\gamma$-correction. Results correspond to PSNR/SSIM metrics. Training with a linear CRF is optimal for datasets (K\"{o}hler, GoPro) that do not model the $\gamma$ correction. Conversely, incorporating the CRF in the model is better for the GoPro($\gamma$=2.2) and RealBlur datasets. A CRF mismatch degrades the results severely.  }
    \begin{tabular}{cc|cccc}
    \toprule 
         &  & \multicolumn{4}{|c}{Testing Sets} \\
         \cline{3-6}
        Exp &  CRF & GoPro & GoPro ($\gamma$=2.2) &  Kohler & RealBlur \\ 
        \midrule 
        E11 &  $\gamma$=1.0 & \bf{29.61}/\bf{0.892} &  27.98/0.876 & \bf{28.57}/\bf{0.82} &  29.05/0.883 \\ 
        E13 &  $\gamma$=2.2 & 28.37/0.882 &  \bf{29.67}/\bf{0.896} & 27.54/0.80 & \bf{29.41}/\bf{0.887} 
        \\
    \bottomrule 
    \end{tabular}
    \label{tab:gamma_corrected}
\end{table}

\paragraph{A Single Training for Different CRFs\label{sec:single_training}} 

A simple modification in the training strategy consists in restoring the images in the photon domain by reversing the $\gamma$-correction process before feeding the network with the blurry image. Subsequently, the network's output is $\gamma$-corrected before computing the \textit{loss}. The primary advantage of this procedure lies in its flexibility during inference: the user decides whether to apply the $\gamma$ correction or not. Consequently, the same model can effectively accommodate a wider set of images, which enhances its usefulness. To generalize this procedure to an even wider range of images, it's necessary to reverse the effects of the CRF. This requires either knowing the CRF or being able to estimate it from a single image.\footnote{Some methods have been proposed to estimate a CRF from a single image \cite{tai2013nonlinear, li2017crf, sharma2020single}}

The Supplementary Material illustrates why the CRF mismatch affects the restorations on synthetic patterns and real images using various CRFs. We also show that training in the photon domain is effective for images affected by CRFs of the $\gamma$-family.

\subsubsection{Saturated pixels}

\begin{figure*}[ht]
\centering
\setlength{\tabcolsep}{1pt}

  \begin{tabular}{ccc||c}
    Blurry & No augmentation  & Multiplicative augment.  & Original GoPro \citep{tao2018scale} \\
    \includegraphics[trim=200 600 0 70, clip,width=0.24\textwidth]{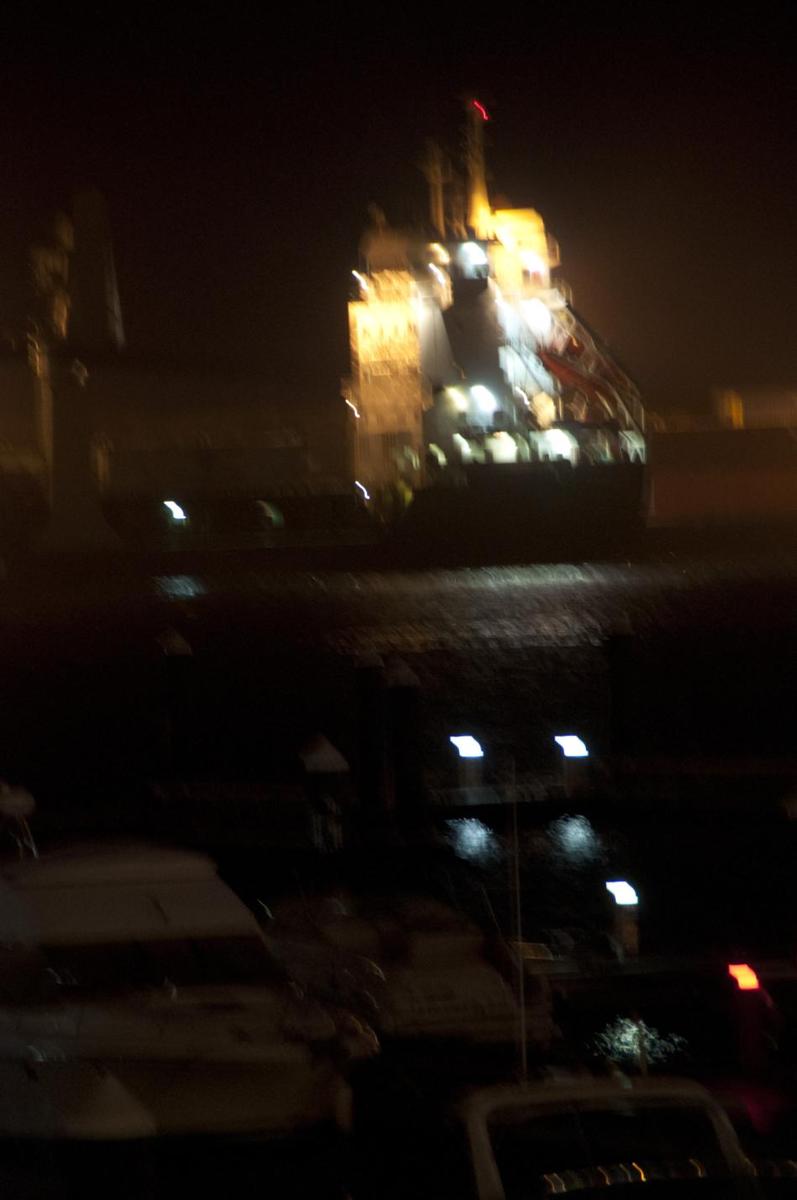} &
    \includegraphics[trim=200 600 0 70,                    clip,width=0.24\textwidth]{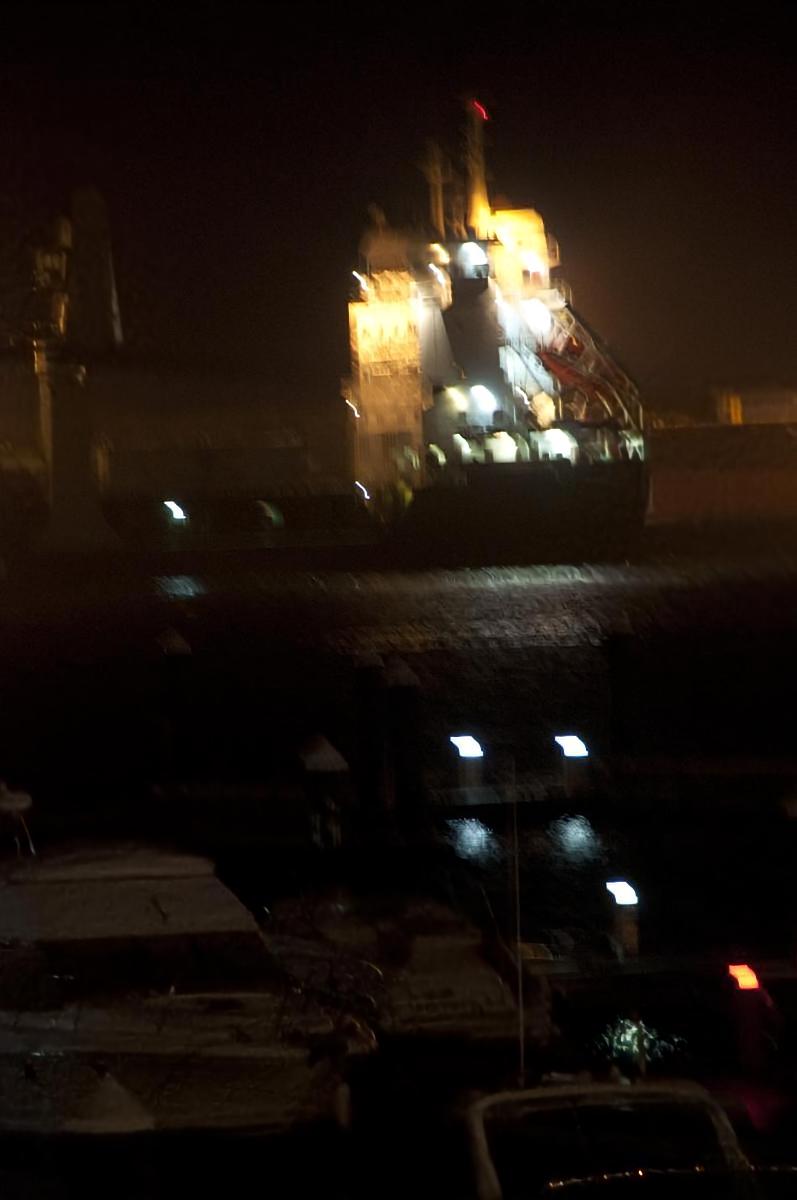}  &
    \includegraphics[trim=200 600 0 70,                    clip,width=0.24\textwidth]{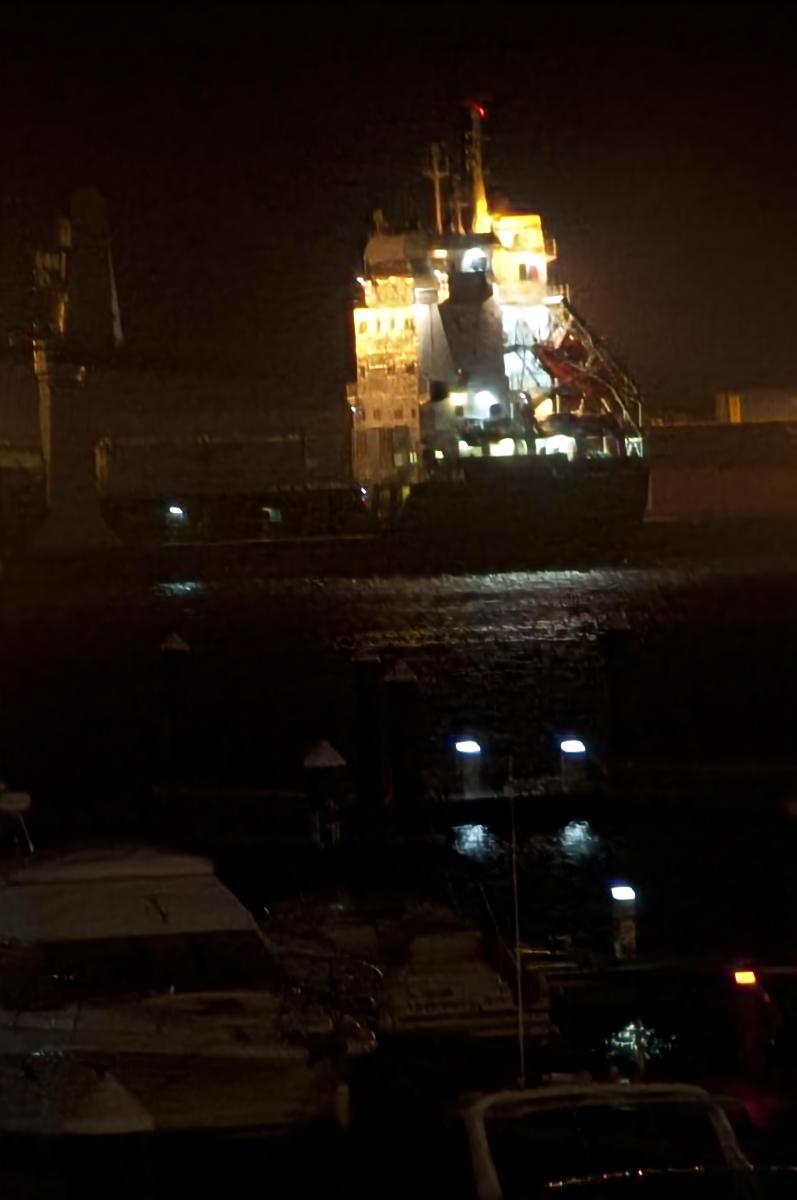}
    &  \includegraphics[trim=200 600 0 70, clip,width=0.24\textwidth]{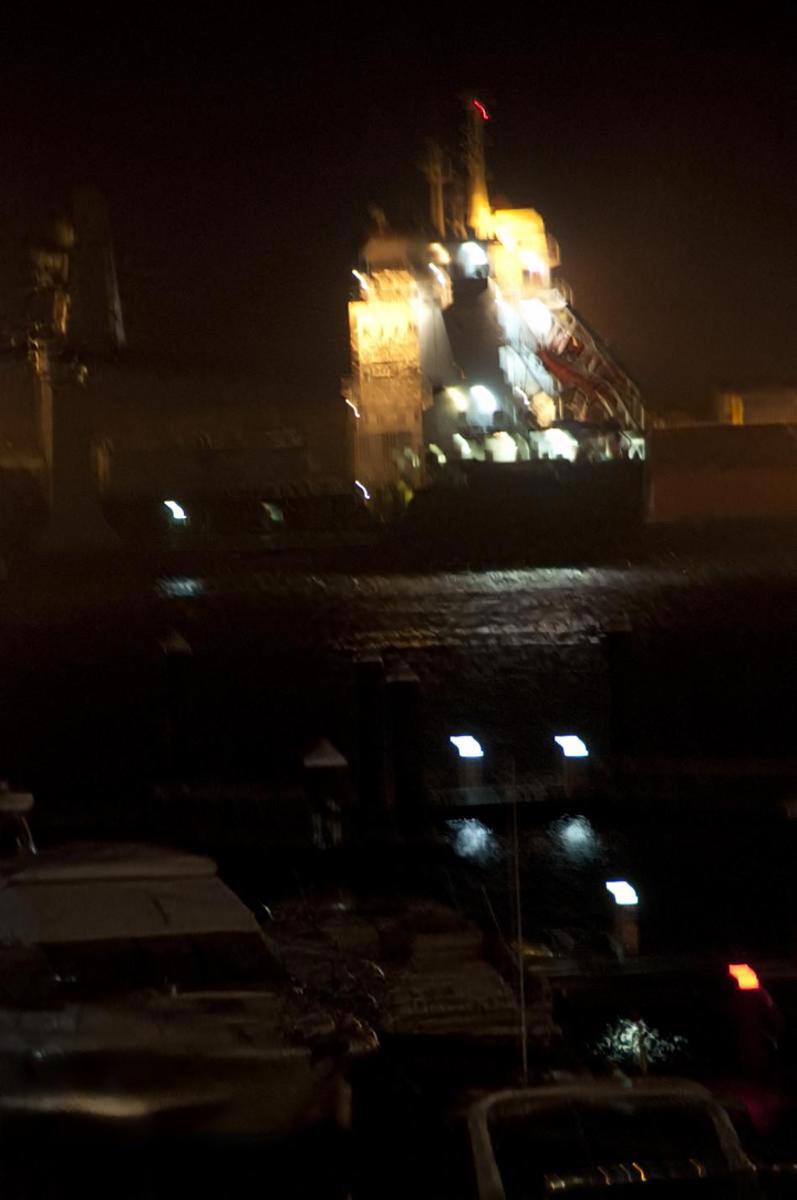}  \\
    Multiplicative clipped & Random streaks &Saturation streaks & RealBlur\_j \citep{rim_2022_ECCV} \\
    \includegraphics[trim=200 600 0 70, clip,width=0.24\textwidth]{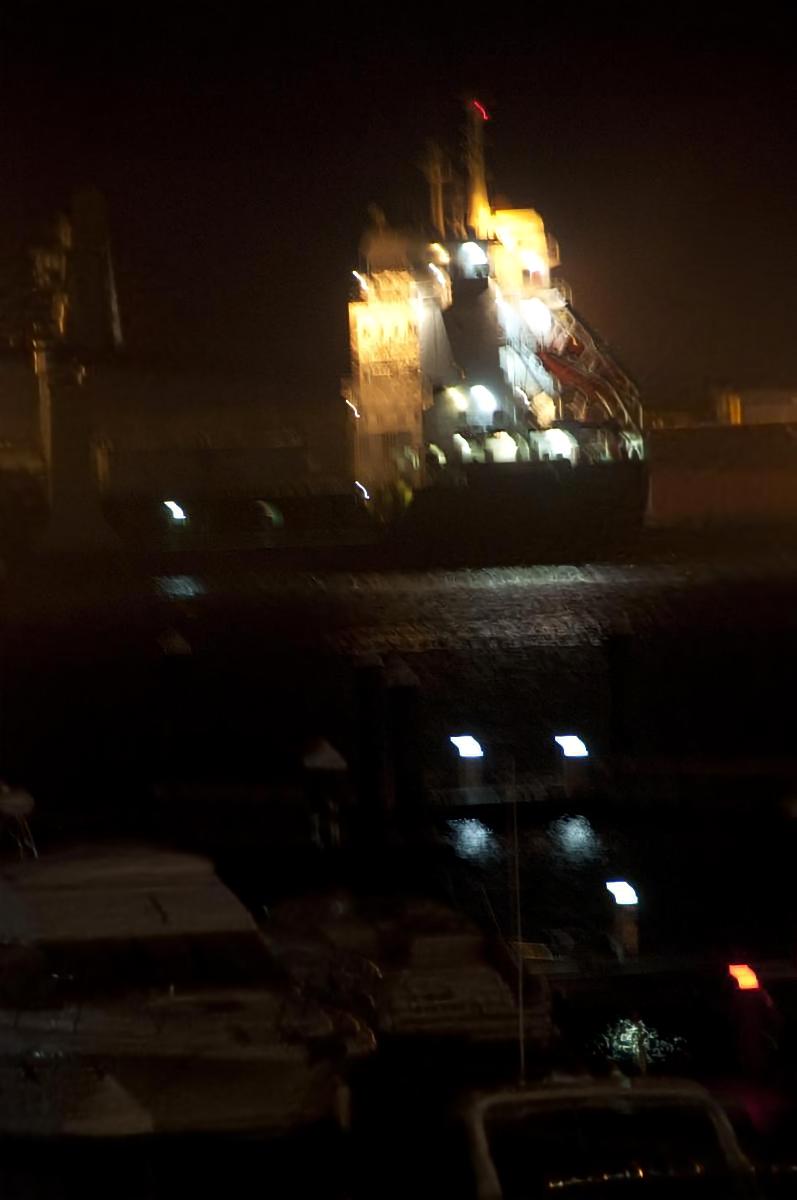} &
    \includegraphics[trim=200 600 0 70,                    clip,width=0.24\textwidth]{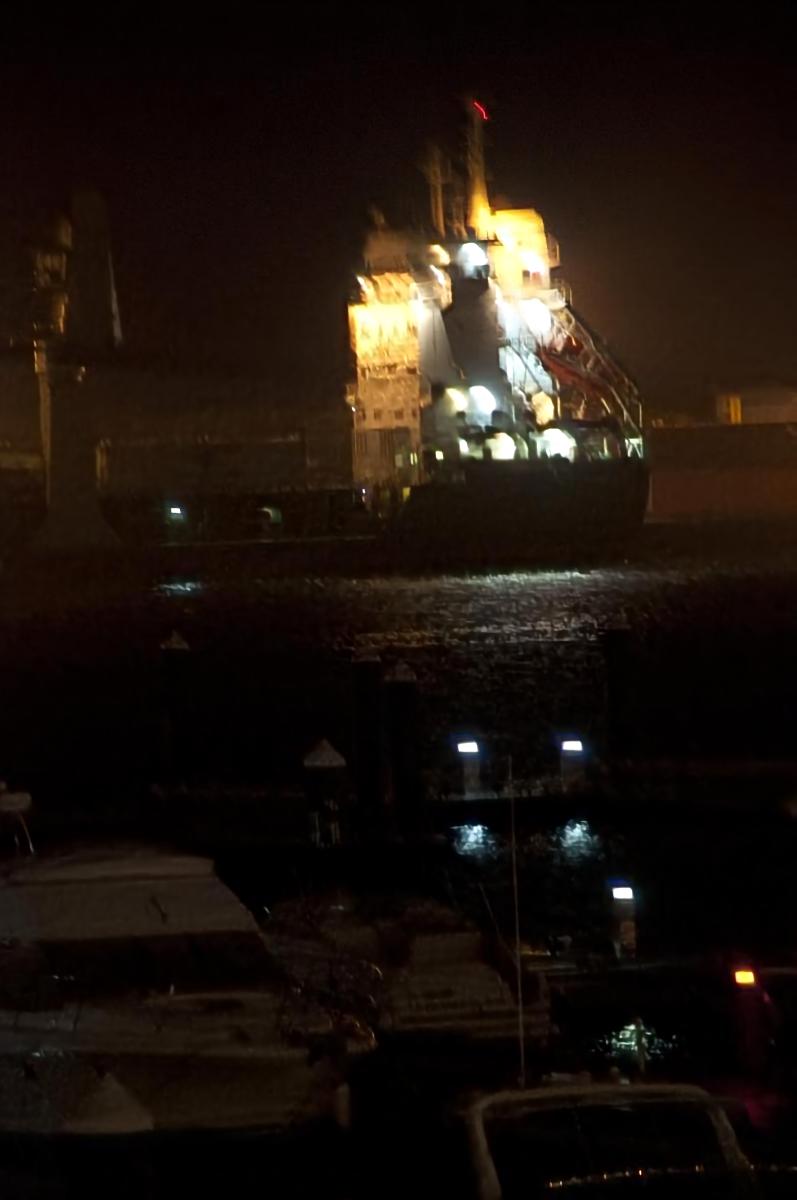}  &
    \includegraphics[trim=200 600 0 70,                    clip,width=0.24\textwidth]{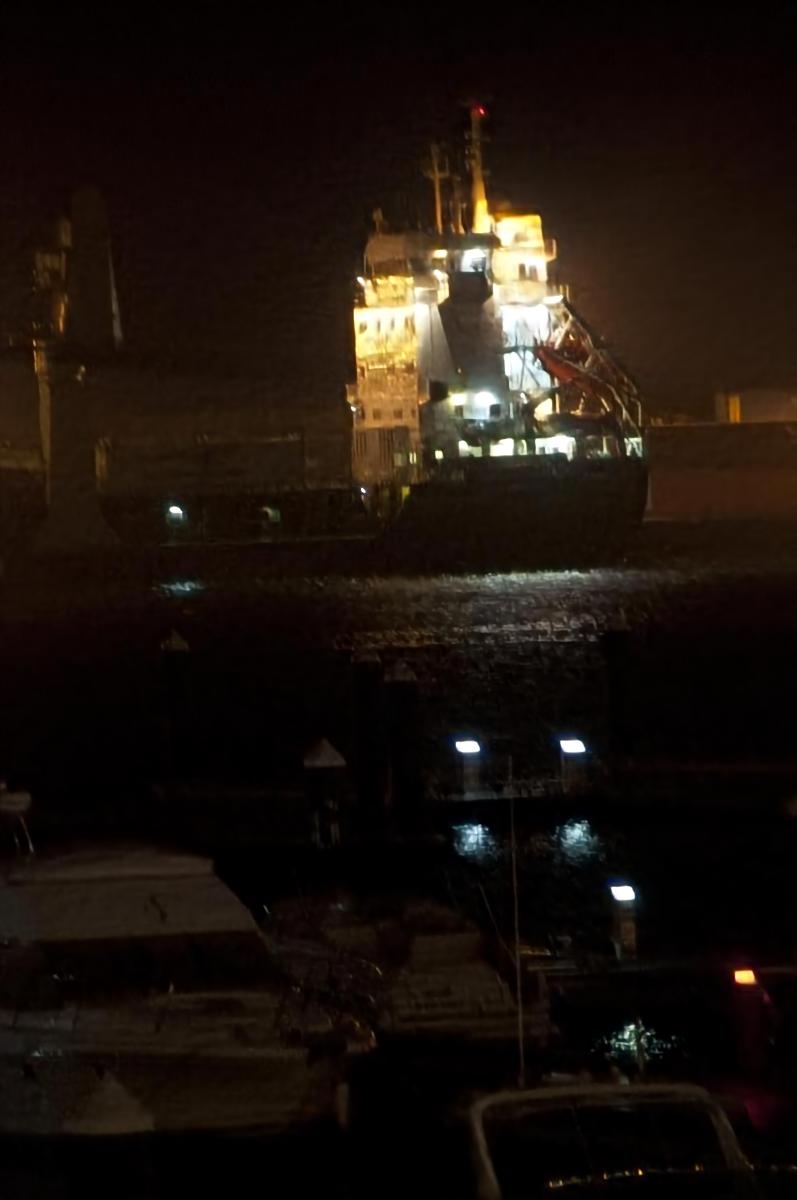} &
    \includegraphics[trim=200 600 0 70, clip,width=0.24\textwidth]{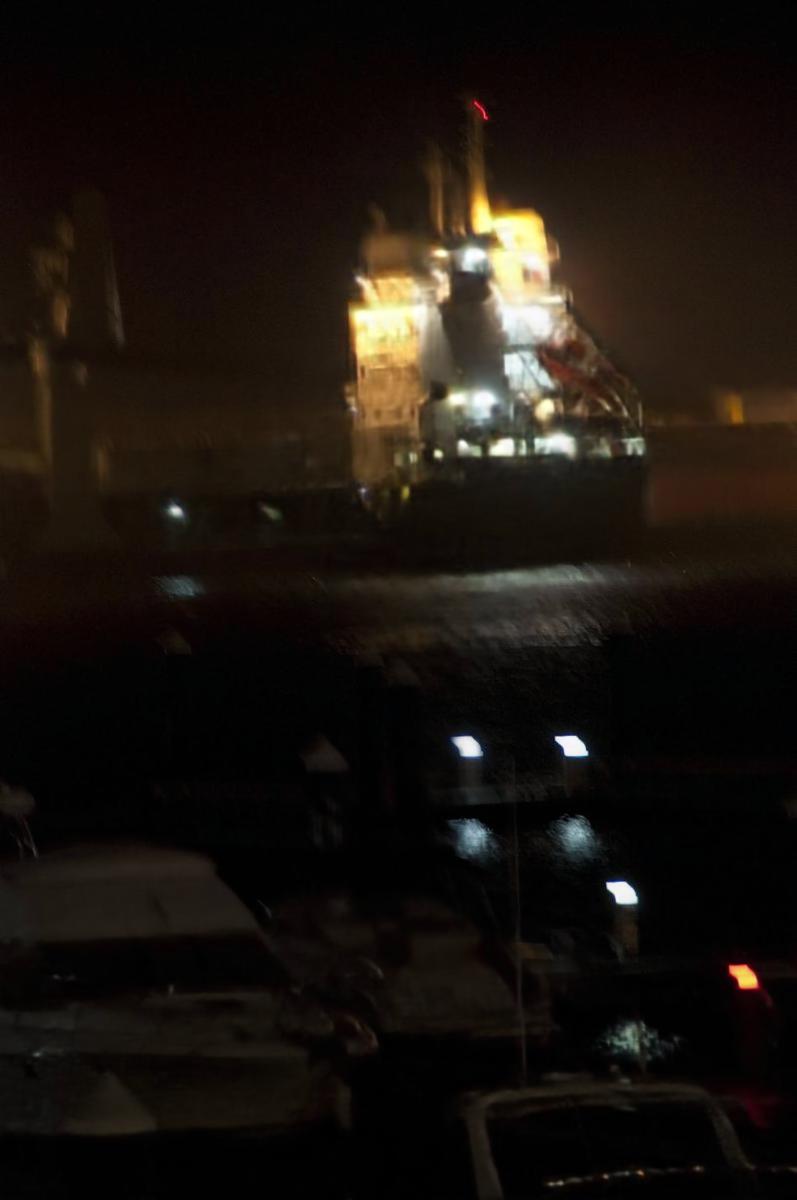}  \\
     \end{tabular}
  \caption{Comparison of the assessed \textit{data augmentation} techniques when deblurring a challenging and common low-light scenario of motion blur with saturated pixels. Additionally, we present the results of the SRN network, as supplied by the authors (trained on GoPro), and the SRN model trained on a real-world dataset consisting of images captured under low-light conditions, as made available by \cite{rim_2022_ECCV}, for reference. Test image sourced from \cite{lai2016comparative}.}
  \label{fig:sat_augmentations}
\end{figure*}

\begin{table}[t]
    \centering
    \setlength{\tabcolsep}{1pt}
    \footnotesize  
    \caption{Influence of illumination augmentation. Results correspond to PSNR/SSIM metrics. Augmenting training images to account for saturated pixels (E14, E16, E17) considerably improves the results on real-world images affected by saturation. Conversely, all the assessed augmentations diminish the performance on the GoPro($\gamma$=2.2) benchmark dataset. The synthetic benchmark dataset is not a good proxy for the performance of real-world images.}
    \begin{tabular}{l|c|cc}
         \toprule  
         & \multicolumn{1}{c|}{\emph{SBDD\_U}} & \multicolumn{2}{|c}{Testing Sets} \\
         \cline{3-4}
        Exp &  Augmentation Method & GoPro ($\gamma=2.2$)  & RealBlur \\ 
        \midrule  
        E13 & No augmentation  &  \bf{29.67}/\bf{0.896}  &  29.41/0.887  \\ 
        E14 &  Multiplicative & 29.39/0.891 &  \bf{30.13}/\bf{0.891}  \\
        E15 &  Mult. clipped & 29.60/0.893 &  29.23/0.881  \\
        E16 &  Random streaks & 29.56/0.893 &  29.67/0.890 \\
        E17 &  Sat. streaks & 29.26/0.89 &  30.03/0.89 \\
        \bottomrule  
    \end{tabular}
    \label{tab:augmentation_results}
\end{table}

As indicated in \cref{tab:augmentation_results}, the sole augmentation method that failed to yield an improvement over \textit{no augmentation} was the \textit{multiplicative clipped augmentation}. Conversely,  \textit{multiplicative augmentation} improved the RealBlur results the most. Compared with \textit{no augmentation}, all the methods obtained worse results on the GoPro dataset, underscoring the distinct characteristics of both datasets. A comparative restoration example of a real-blur image, obtained by training with the considered saturation data augmentation strategies, is shown in \cref{fig:sat_augmentations}. More examples are provided in the Supplementary Material. 

One important consideration when comparing results is that when training the network with a fixed number of iterations and introducing more degradations through augmentation, the performance within a specific subset may deteriorate when confronted with a higher number of degradations. In \cref{tab:more_iterations}
we evaluate the effect of adding more iterations.

\subsection{Generalization performance of the dataset generation methods}

\begin{table*}[ht]
    \caption{Average PSNR/SSIM on common benchmark datasets (RealBlur, K\"{o}hler, GoPro) for SRN \citep{tao2018scale} trained on datasets produced with different generation methods. A Camera Response Function match between training and testing sets is crucial for obtaining favorable outcomes. Non-uniformly blurred training sets outperform their uniform counterparts specifically on the GoPro dataset. Convolution-based approaches exhibit superior \textbf{cross-domain performance}. When the test set is generated following the same procedure as the training set, we show the results in red. The generation procedure contaminates those values and does not reflect the performance of the methods on real images.} 
    \label{tab:res_srn}
    \small 
    \setlength{\tabcolsep}{1pt}
    \centering
    \begin{tabular}{c|l|cccc}
             \multicolumn{2}{c}{ \textbf{}} & \multicolumn{4}{|c}{ Test Sets} \\
      \toprule
       Generation method & \hspace{50pt}    Training Set     &   RealBlur  &   K\"{o}hler & GoPro & GoPro($\gamma=2.2$) \\
      \hline 
      \multirow{8}{*}{{Convolutional}} & SBDD\_U ($\gamma=1.0$)   &  30.0/0.893   &  \textbf{28.55}/0.817  & 29.6/0.892 & 28.5/0.883   \\
       & SBDD\_U ($\gamma=2.2$)  & 30.33/0.894  & 27.23/0.809  & 28.22/0.879 &  29.63/0.896   \\
       & SBDD\_U ($a=5$)  &  \textbf{30.84}/0.901  & 28.04/0.811  & 28.12/0.879  & 28.85/0.888     \\
       & GoPro\_U \citep{rim_2022_ECCV}      & 30.75/\textbf{0.902}   & 27.67/0.809  & 27.1/0.859  & 27.87/0.871 \\
       
        \cline{2-6} 
       
       & SBDD\_NU ($\gamma=1.0$)  & 29.8/0.890 &  28.15/\textbf{0.818}    & \textbf{29.83}/\textbf{0.893} &   28.5/0.883 \\
       & SBDD\_NU ($\gamma=2.2$)  & 30.15/0.891  & 27.2/0.807  & 28.28/0.880 &   \textbf{29.91}/\textbf{0.898}     \\
       & SBDD\_NU ($a=5$)  & 30.66/0.900  & 27.78/0.817  & 28.16/0.881 &  29/0.9       \\
       \hline 
       \multirow{5}{*}{{Frame Averaging}} &  GoPro\_ABME\_aug \citep{rim_2022_ECCV} & 30.33/0.890 & 26.66/0.779 & 27.52/0.862 & 28.66/0.879  \\
       & RSBlur\_syn \citep{rim_2022_ECCV} & 29.32/0.876   & 26.58/0.777  & 27.86/0.867 & 29.48/0.885 \\
       & RSBlur\_syn\_aug \cite{rim_2022_ECCV} & 29.81/0.883   & 26.22/0.780  & 27.65/0.861 &  29.01/0.876 \\
       & GoPro \citep{nah2017deep}-524k    &  28.56/0.867  &  26.9/0.789 &  \textcolor{red}{30.72}/\textcolor{red}{0.907} & 28.53/0.888  \\
        & GoPro($\gamma$)-524k \cite{tao2018scale}    &  28.55/0.863  & 26.38/0.771  & 29.03/0.892  & \textcolor{red}{31.00}/\textcolor{red}{0.911}  \\
       & REDS \citep{nah2019ntire}-450k    &  28.95/0.868  & 25.46/0.741  & 26.84/0.845   &   28.75/0.881 \\
       \hline 
       \multirow{2}{*}{{Beamspliter}} & RSBlur\_real \citep{rim_2022_ECCV} & 29.81/0.880   &  26.45/0.777 & 27.54/0.861 & 28.95/0.877 \\
       & RealBlur\_j \citep{rim_2020_ECCV, rim_2022_ECCV} & \textcolor{red}{30.77}/\textcolor{red}{0.899}   & 26.75/0.789  & 26.89/0.848 & 27.45/0.855 \\
       \bottomrule 
    \end{tabular}
\end{table*}

We generated several instances of the proposed dataset by fixing the following parameters: \textit{multiplicative augmentation} to deal with saturated pixels, continuous kernels generated with F=1000, texp=0.5, and support 65$\times$65. Besides building uniformly and non-uniformly blurred datasets, we also varied the CRF used to generate the blurry images. For example, the training set denoted as SBDD\_U~($\gamma=2.2$) was generated by convolving sharp images from the GoPro training set in the photons domain with a single kernel, assuming a $\gamma$-function with $\gamma=2.2$ as the CRF. Similarly, SBDD\_U~($a=5$) was generated assuming an exponential CRF with $a=5$. By default, our models underwent training for 200 epochs, equivalent to 262,000 iterations.  As a reference, we also present the results obtained with the models provided by \citet{rim_2022_ECCV} and the SRN model \cite{tao2018scale} trained on the GoPro, GoPro~($\gamma$=2.2), and REDS dataset. When the authors trained a model for more iterations than those used by default, we indicated the iterations with a suffix. All the results are reported in \cref{tab:res_srn}.

\begin{figure}
    \centering
    \includegraphics[width=0.5\textwidth]{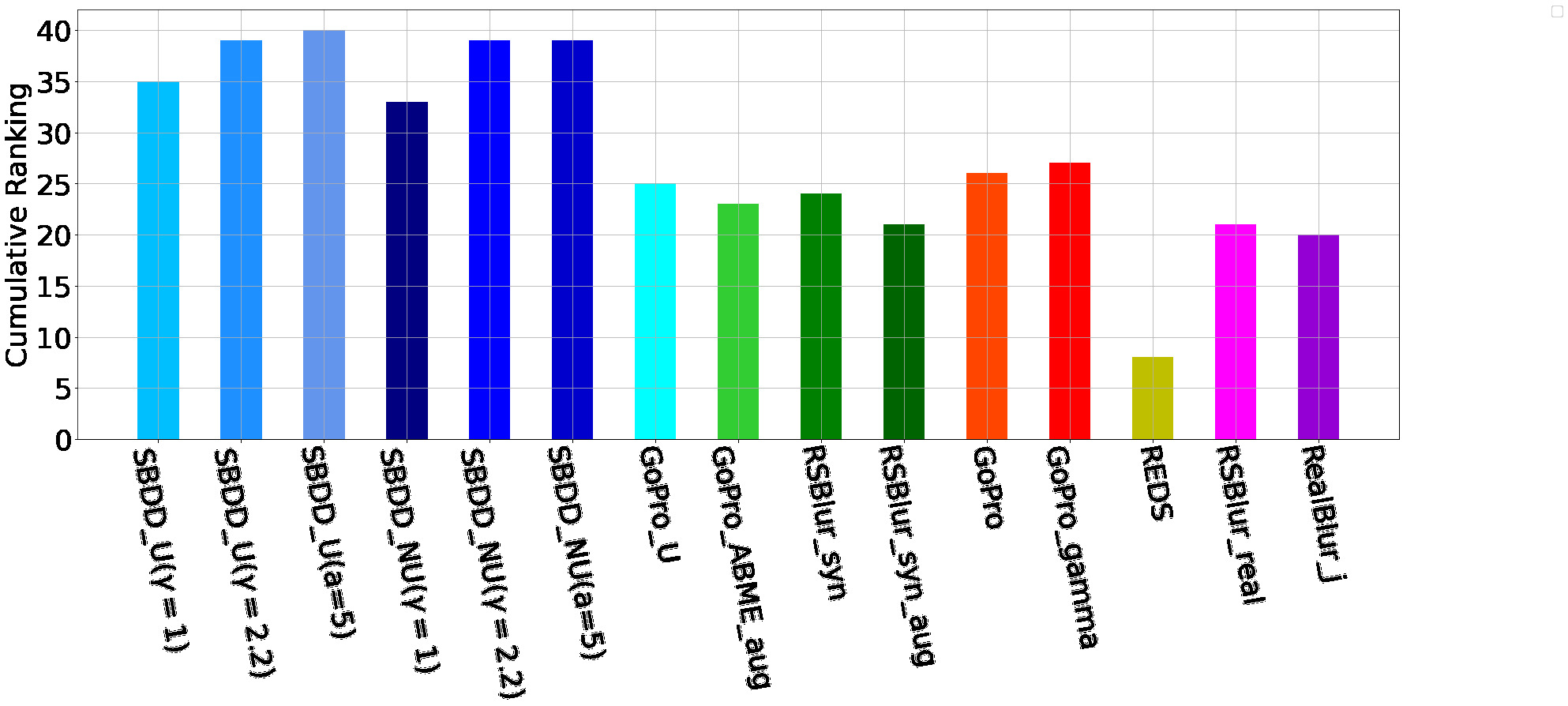}
    \caption{Cumulative ranking of trained models. The 15 training sets are ordered according to the SRN performance on each benchmark dataset (RealBlur, Kohler, GoPro, GoPro($\gamma=2.2$). The best on a evalation dataset accumulates 15 points, the second 14, and so forth. The models trained on datasets generated with the proposed procedure (SBDD) occupy the first six places in the ranking (bluish bars).}
    \label{fig:cumulative}
\end{figure}

\begin{figure*}[t!]
\centering
\small 
\setlength{\tabcolsep}{2pt}

  \begin{tabular}{cccccc}
    Blurry & GoPro \cite{Nah_2017_CVPR}   &  GoPro ($\gamma$) \cite{Nah_2017_CVPR}  & REDS \cite{nah2019ntire}  & RS\_Blur\_syn  \cite{rim_2022_ECCV} \\   %
    \includegraphics[trim=100 250 100 150, clip,width=0.18\textwidth]{imgs/Blurry/Blurry2_1.jpg} &
    \includegraphics[trim=100 250 100 150, clip,width=0.18\textwidth]{imgs/SRN/Blurry2_1.jpg} &
    \includegraphics[trim=100 250 100 150,                    clip,width=0.18\textwidth]{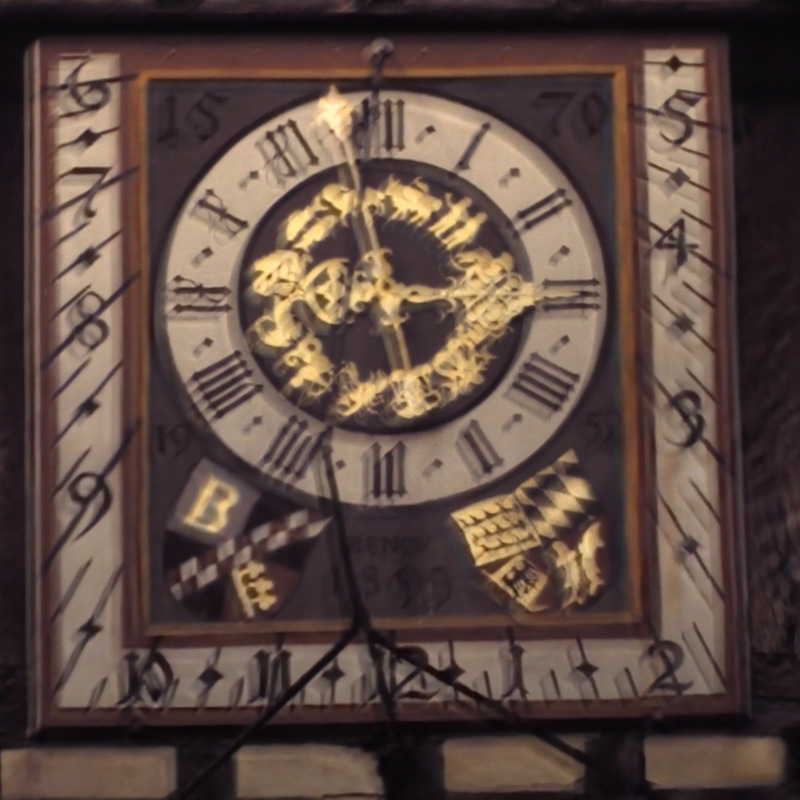}   &
    \includegraphics[trim=100 250 100 150,                    clip,width=0.18\textwidth]{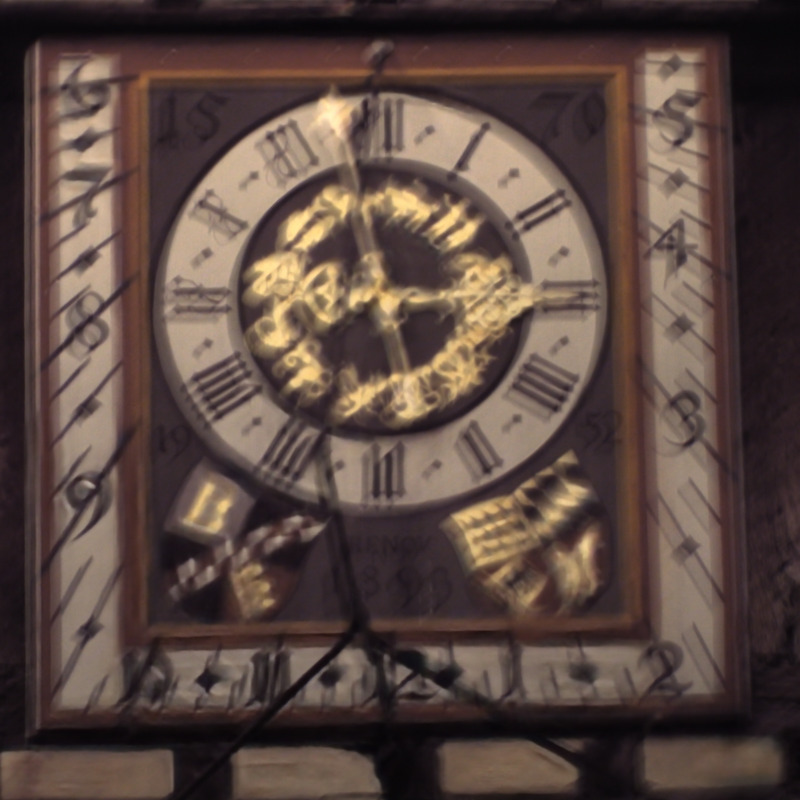} &
    \includegraphics[trim=100 250 100 150,                    clip,width=0.18\textwidth]{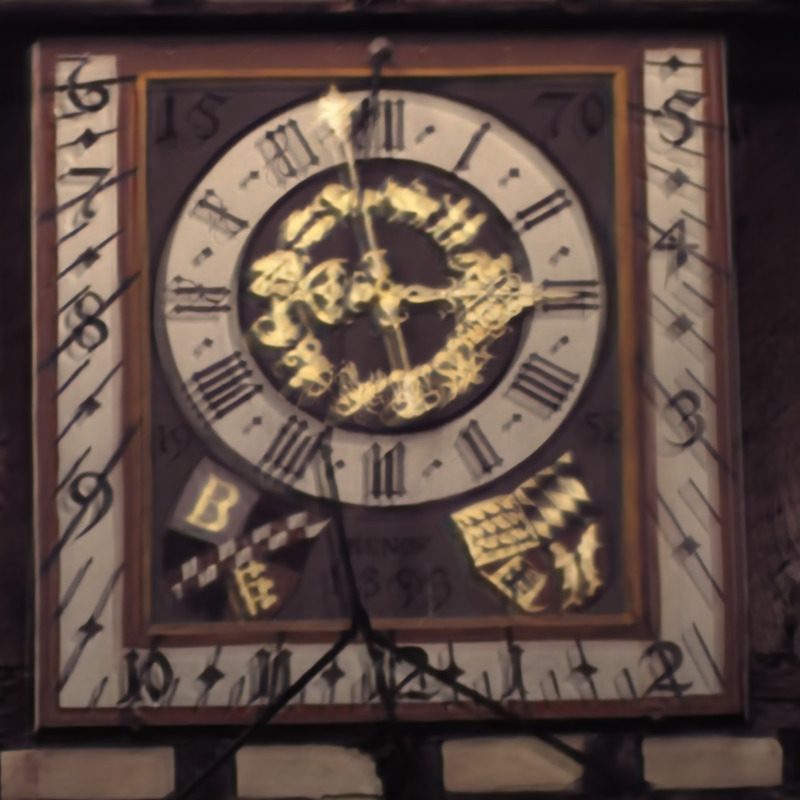} \\

    \includegraphics[trim=0 100 0 0, clip,width=0.18\textwidth]{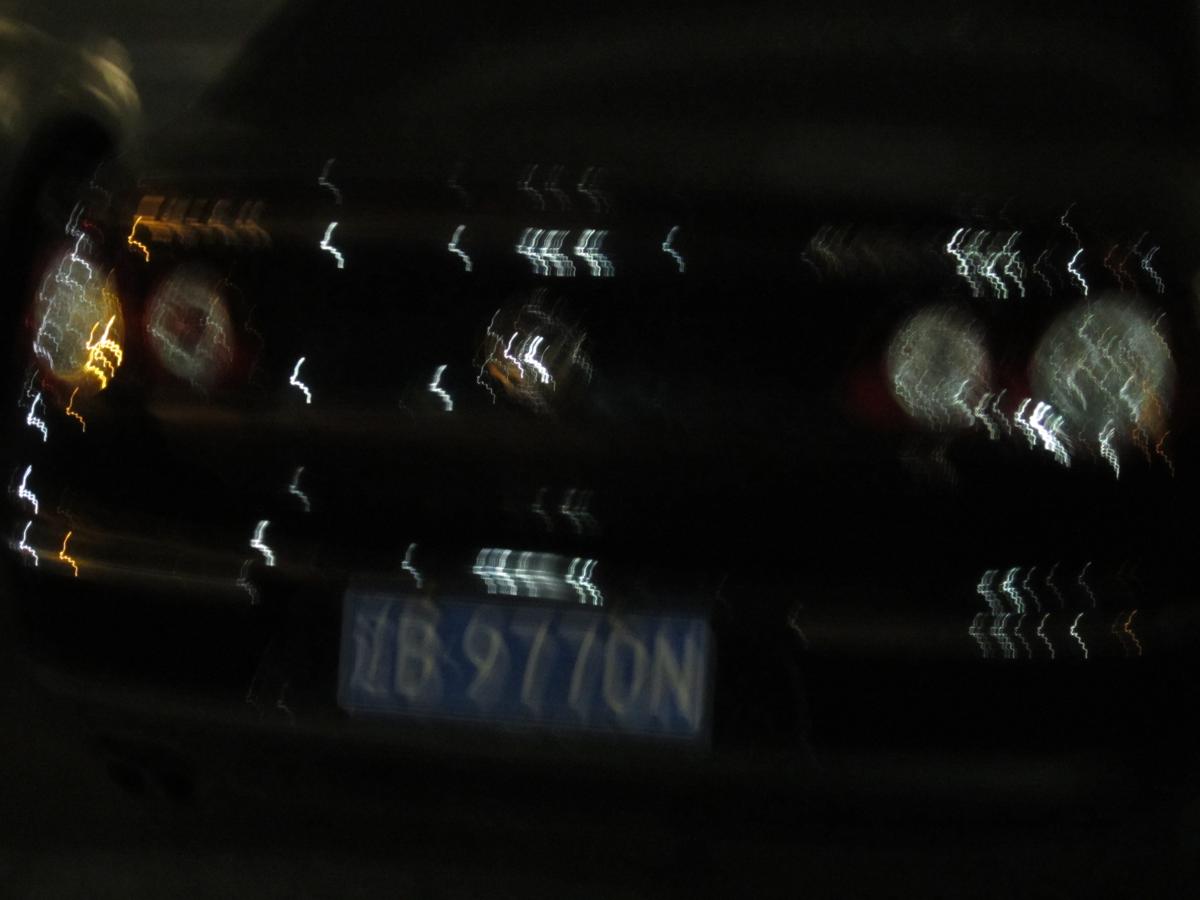} &
    \includegraphics[trim=0 100 0 0, clip,width=0.18\textwidth]{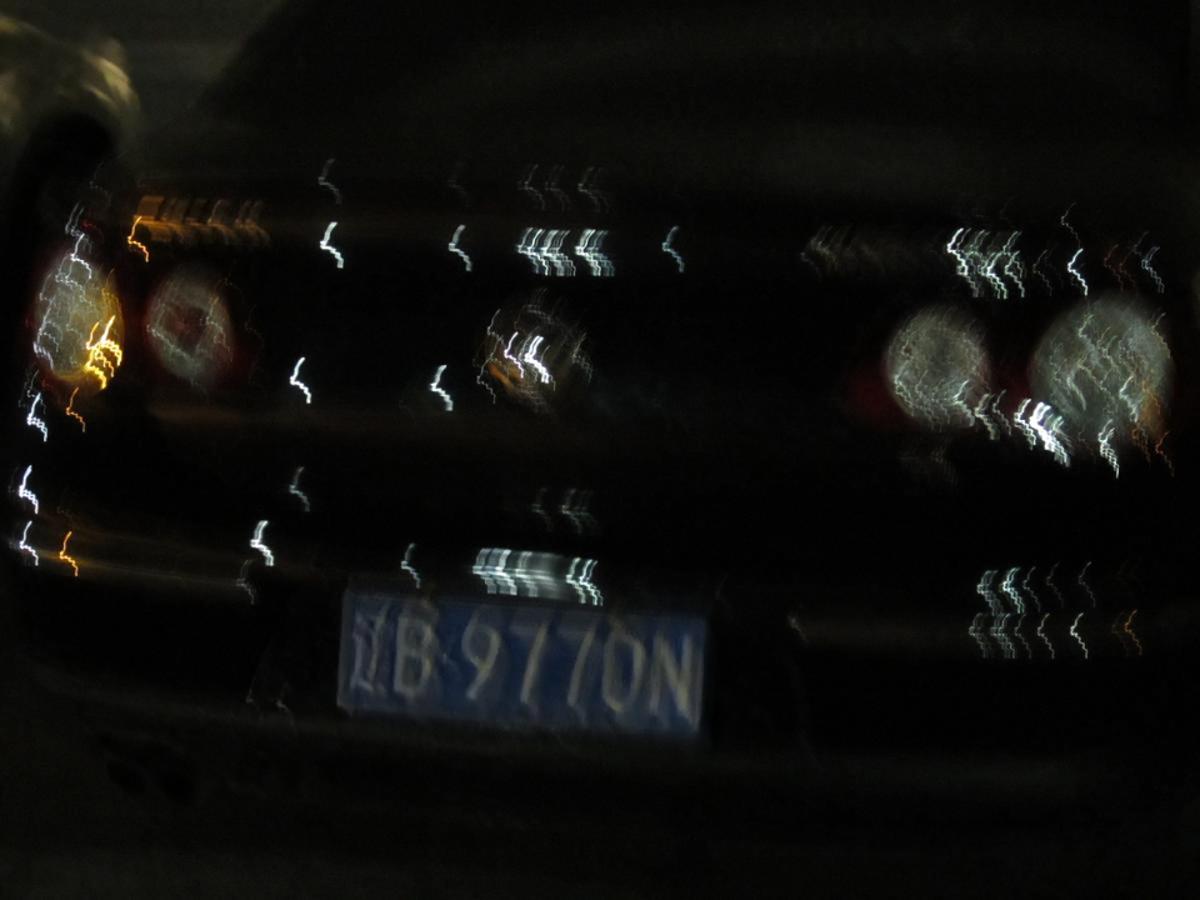} &
    \includegraphics[trim=0 100 0 0,                    clip,width=0.18\textwidth]{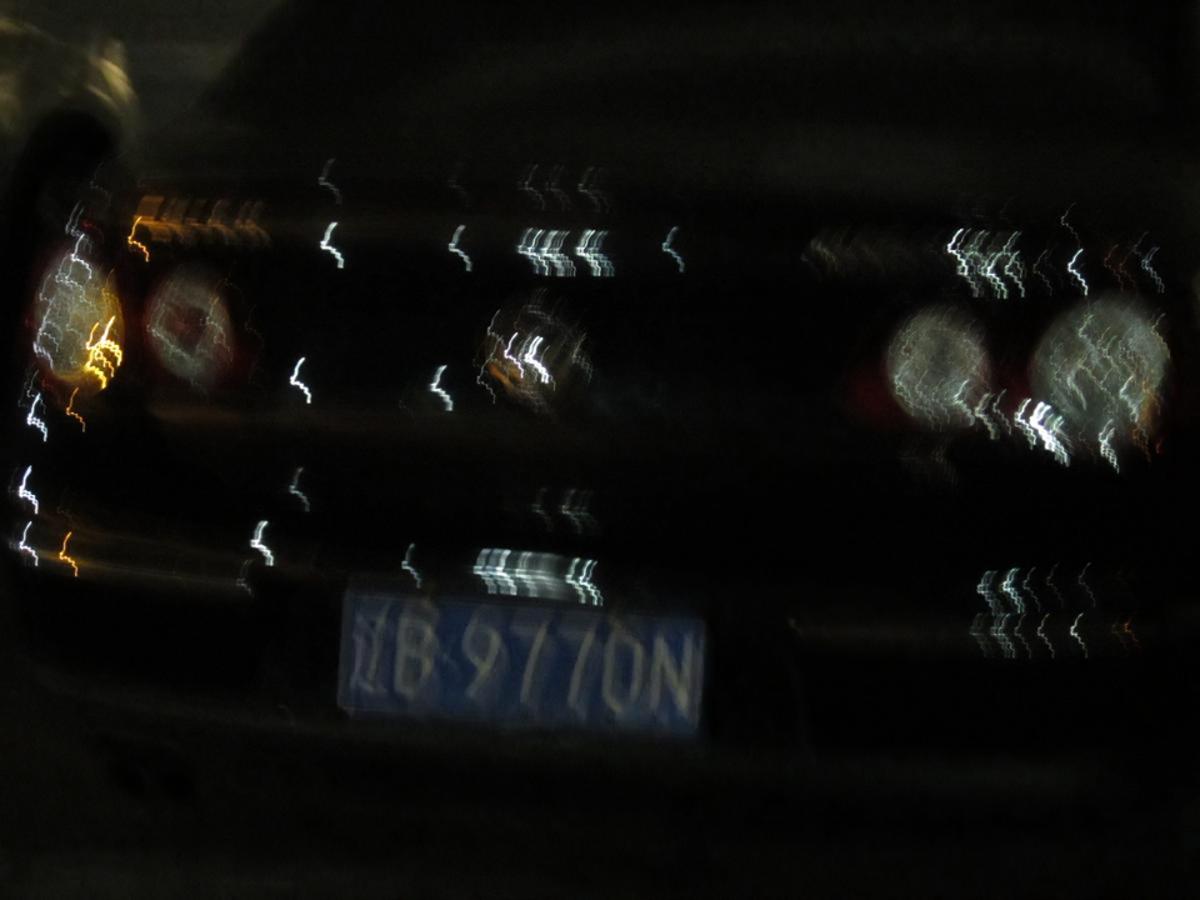}   &
    \includegraphics[trim=0 100 0 0,                    clip,width=0.18\textwidth]{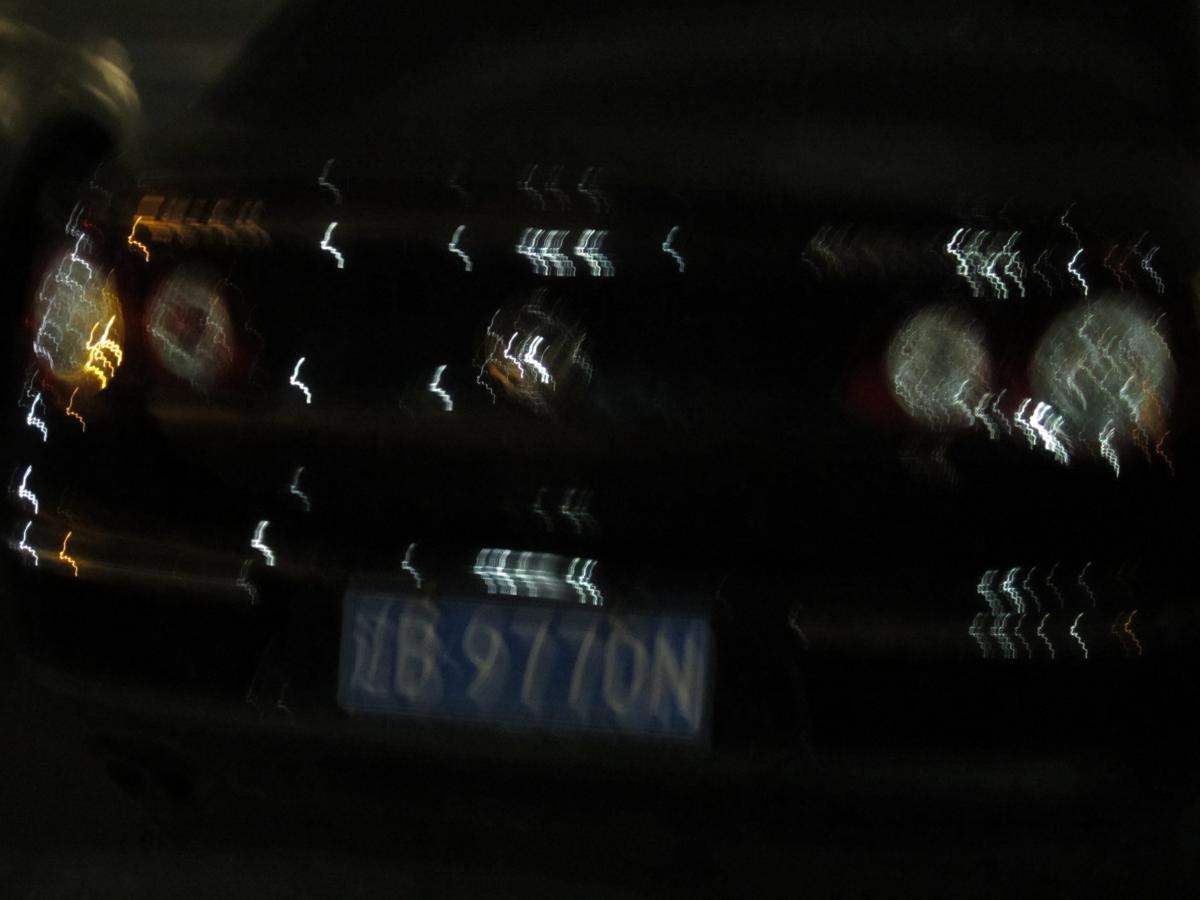}
     &
    \includegraphics[trim=0 100 0 0,                    clip,width=0.18\textwidth]{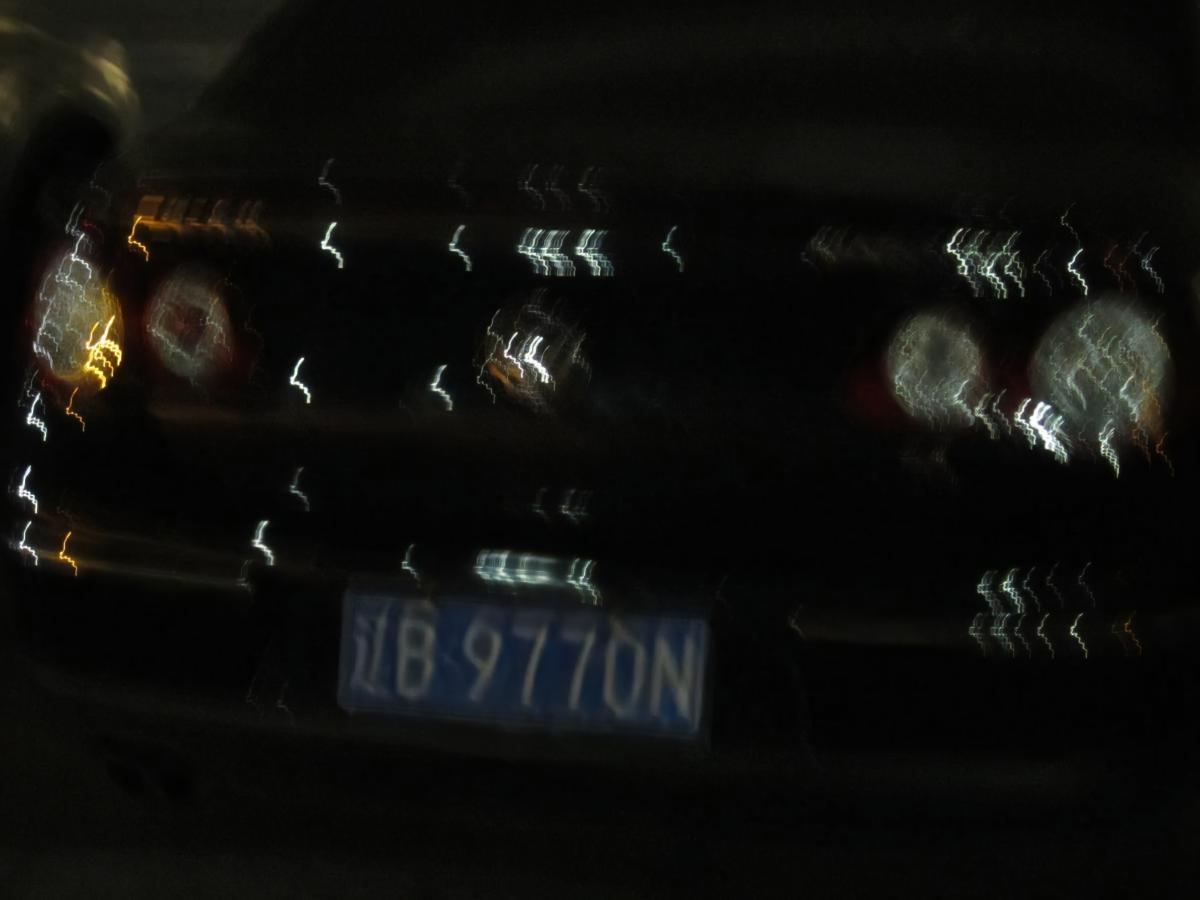} \\
    
     RSBlur \cite{rim_2022_ECCV} &    RealBlur \cite{rim_2020_ECCV}& SBDD\_NU($\gamma=1$)& SBDD\_NU($\gamma=2.2$) & SBDD\_NU($a=5$)  \\

    \includegraphics[trim=100 250 100 150,                    clip,width=0.18\textwidth]{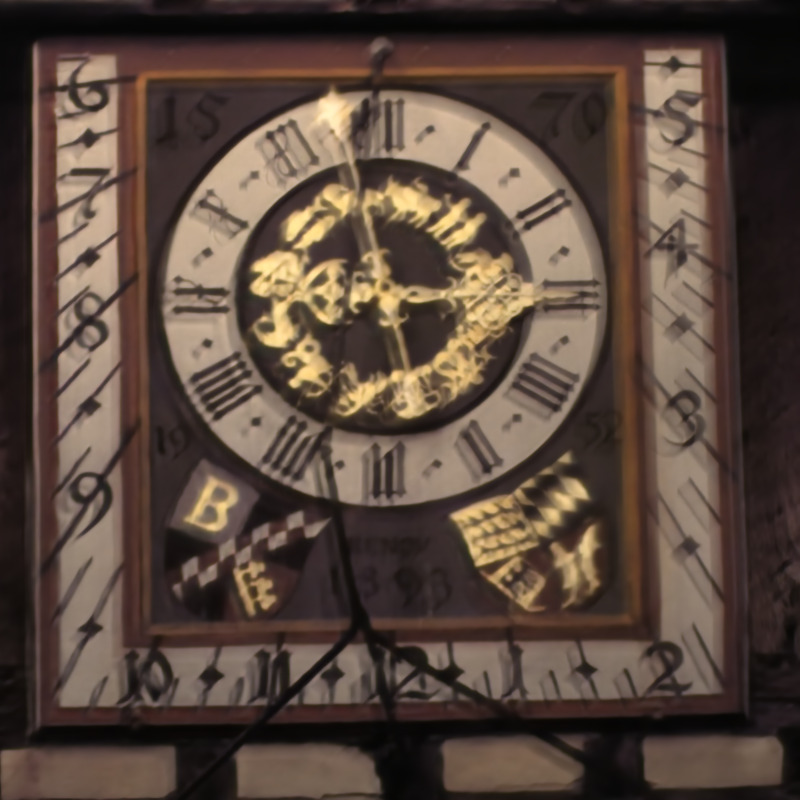} &
    \includegraphics[trim=100 250 100 150,                    clip,width=0.18\textwidth]{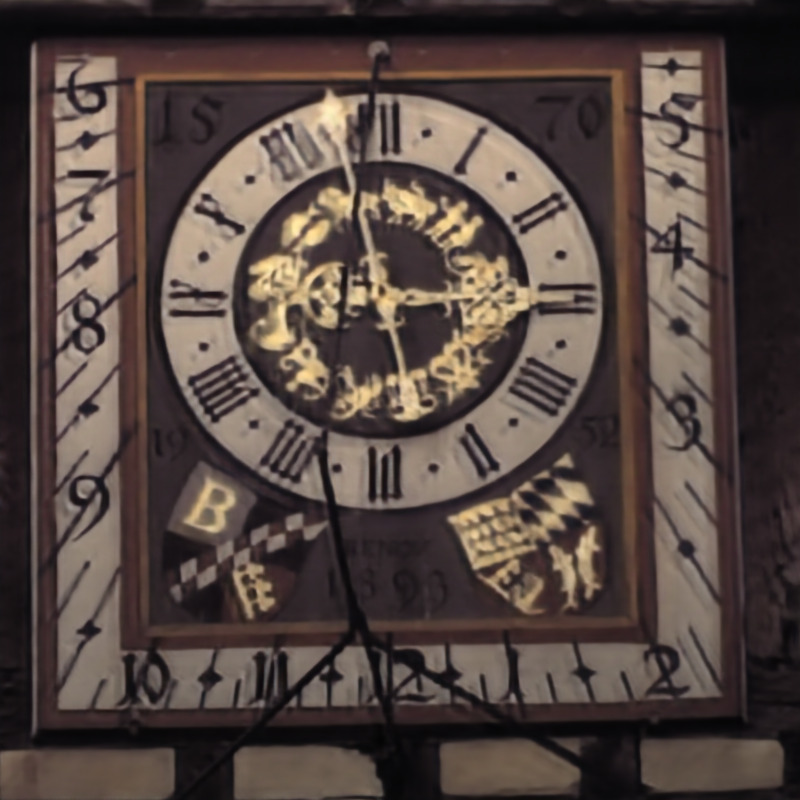} &
    \includegraphics[trim=100 250 100 150,                    clip,width=0.18\textwidth]{imgs/SRN_with_GoPro_non_uniform_mob5_ks65_texp05_F1000_ill_aug_2up_n10_gf1/Blurry2_1.jpg}  &
    \includegraphics[trim=100 250 100 150,                    clip,width=0.18\textwidth]{imgs/SRN_with_GoPro_non_uniform_mob5_ks65_texp05_F1000_ill_aug_2up_n10_gf1/Blurry2_1}  &
    \includegraphics[trim=100 250 100 150,                    clip,width=0.18\textwidth]{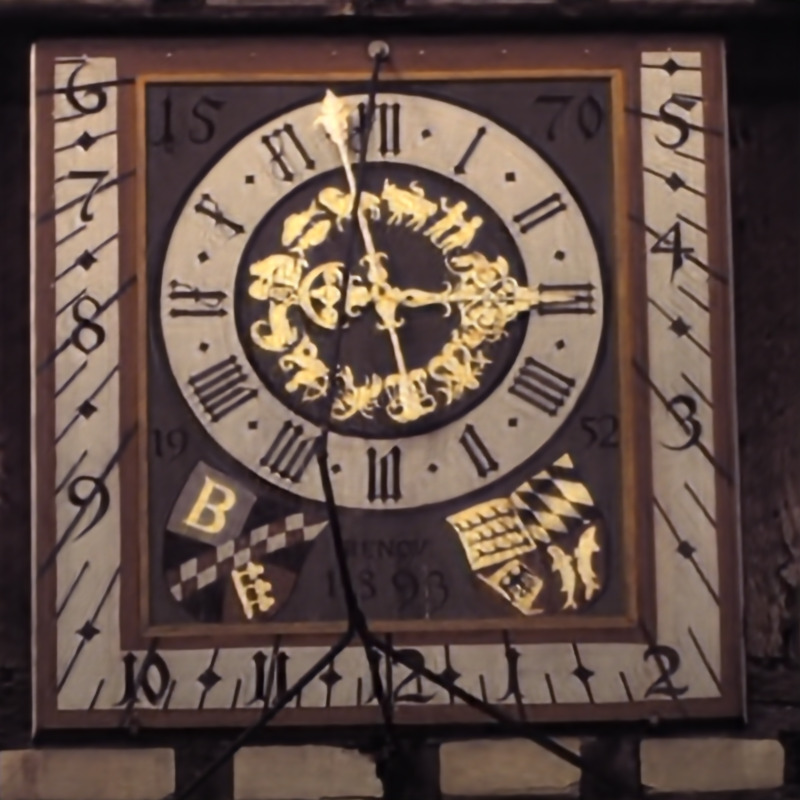} \\

    \includegraphics[trim=0 100 0 0,                    clip,width=0.18\textwidth]{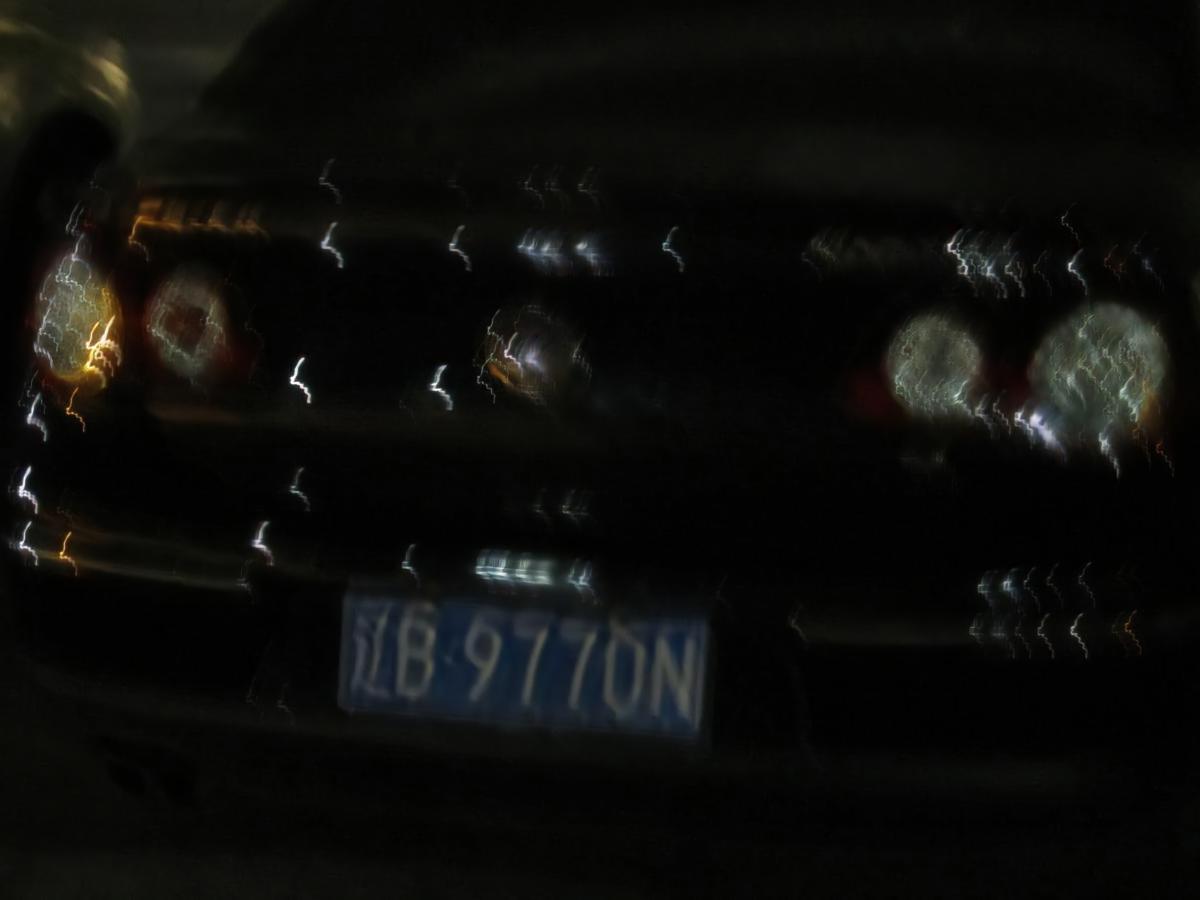}  &
    \includegraphics[trim=0 100 0 0,                    clip,width=0.18\textwidth]{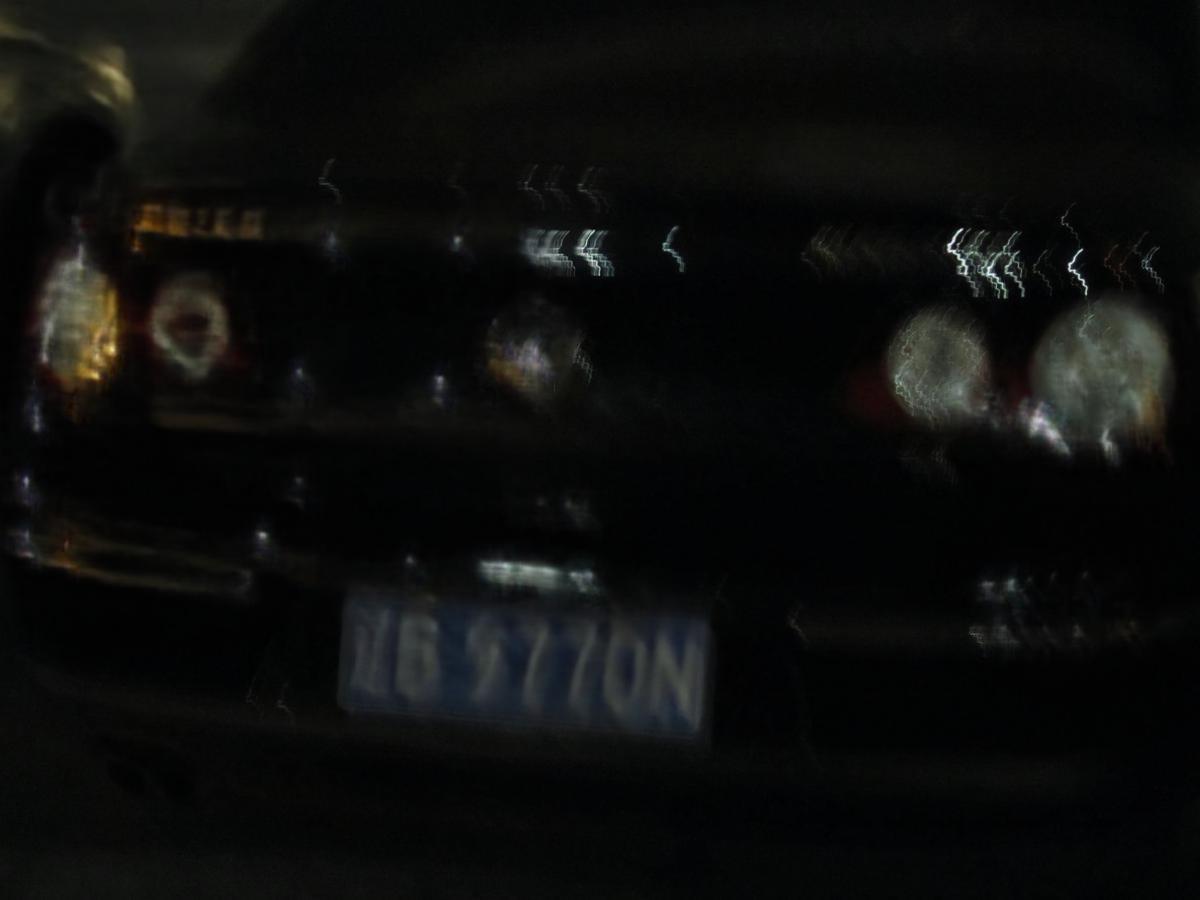} &
    \includegraphics[trim=0 100 0 0,                    clip,width=0.18\textwidth]{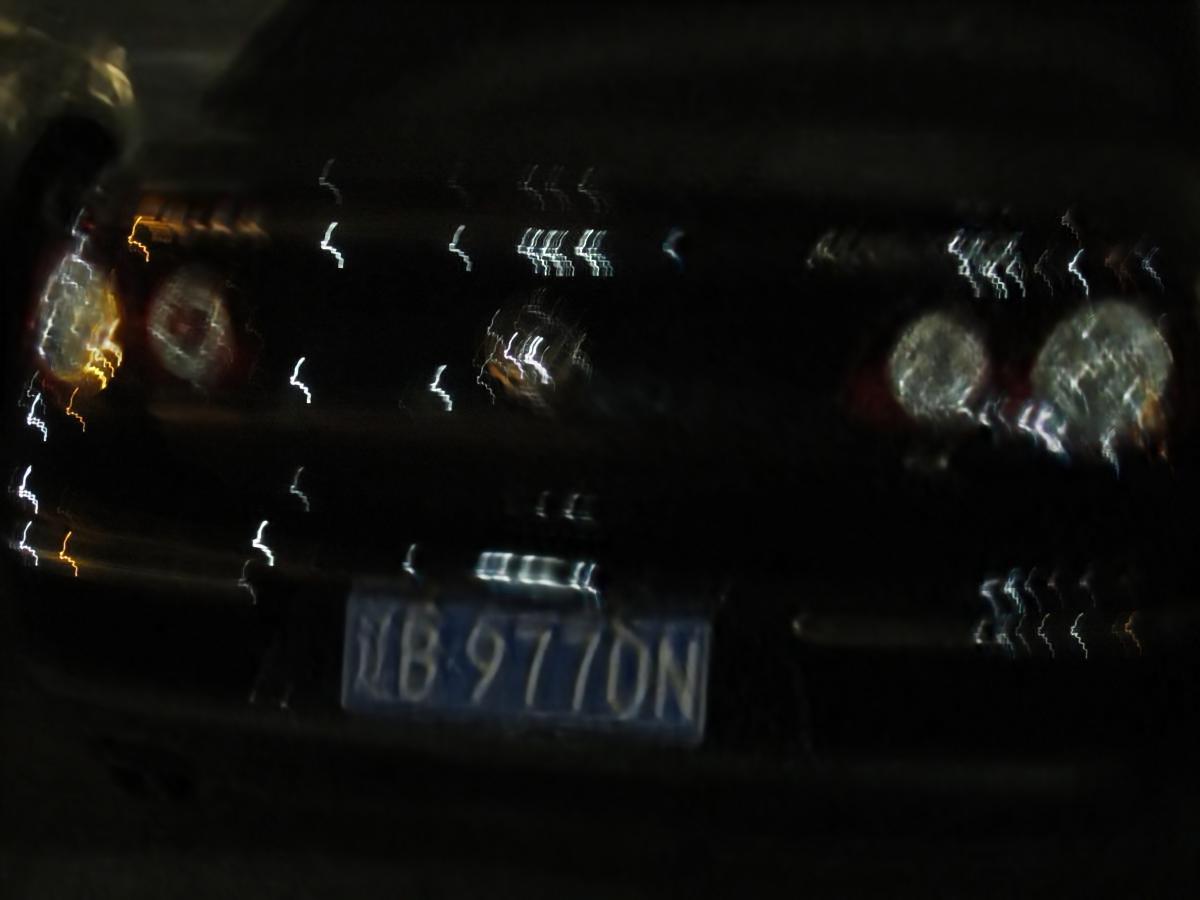}  &
    \includegraphics[trim=0 100 0 0,                    clip,width=0.18\textwidth]{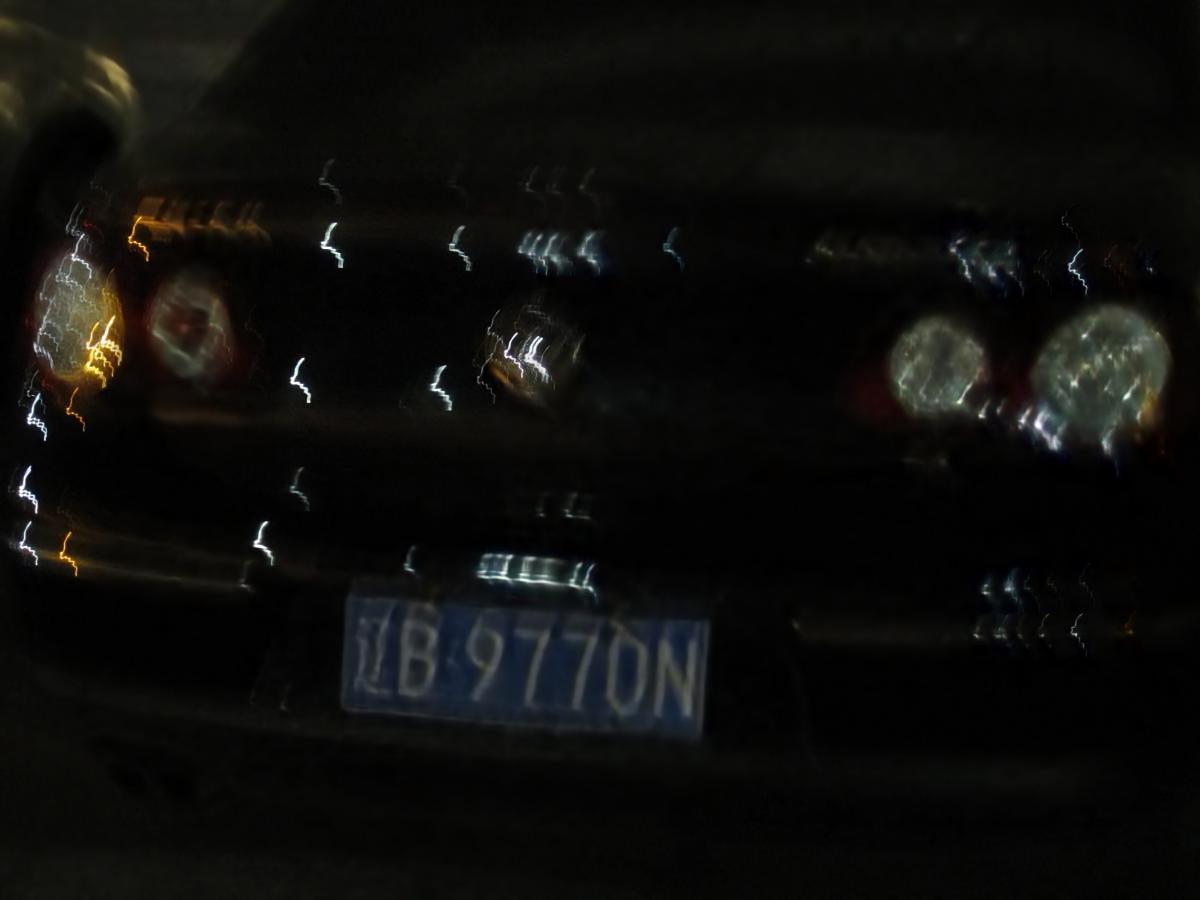}  &
    \includegraphics[trim=0 100 0 0,                    clip,width=0.18\textwidth]{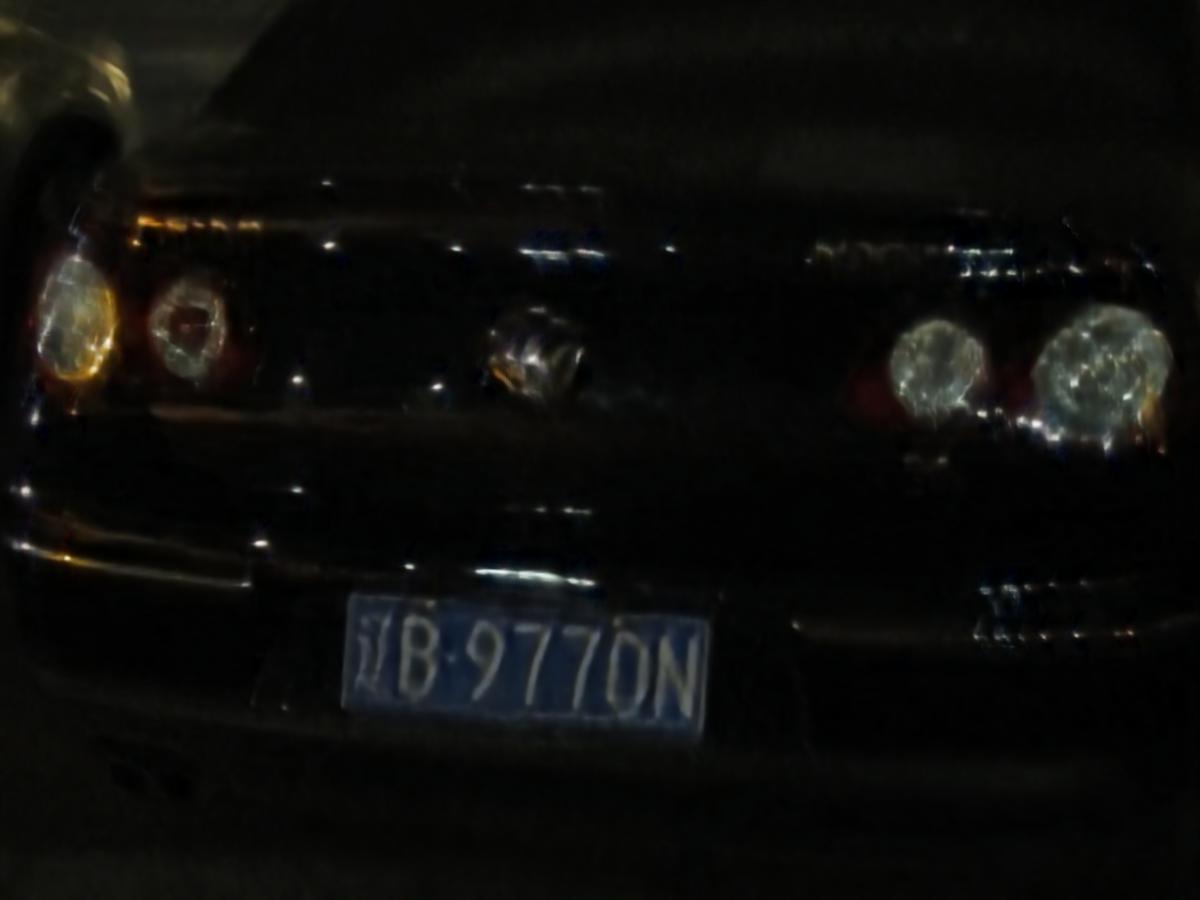}  \\
  \end{tabular}
  \caption{Restoration of real blurry images using the SRN network \citep{tao2018scale} trained on different training sets, specified above each restored image. \textbf{1$^{st}$ and 3$^{rd}$ rows:} image from \citet{kohler2012recording} dataset. \textbf{2$^{nd}$ and 4$^{th}$ rows:} image from \citet{lai2016comparative}.}
  \label{fig:srn_results}
\end{figure*}

The cross-dataset evaluation highlights the pivotal role of the CRF in image deblurring. Models trained with $\gamma=1$ yield the best results on K\"{o}hler's dataset, characterized by a linear CRF. Also, they produce the best results on GoPro. Conversely, models trained with $\gamma=2.2$ achieve the best results on GoPro($\gamma$=2.2). Training with an exponential CRF proves most effective for the RealBlur dataset. While these results align with the expectations, it is noteworthy that a mismatched CRF severely degrades performance. For instance, a model trained on the RealBlur dataset and tested on the RealBlur test set yields excellent results within its training domain but exhibits subpar cross-dataset performance, indicating an inability to grasp the concept of blurriness. Our model trained with an exponential CRF delivers similar results on the RealBlur dataset, but its cross-dataset performance surpasses that of models trained with real images. We argue that the better cross-dataset performance can be attributed to the convolution-based model underpinning the synthesis procedure. Among the models provided by \cite{rim_2022_ECCV}, the best-performing model on the RealBlur dataset is the only convolution-based model, the GoPro\_U. Compared with our best-performing model on RealBlur, the SBDD\_U~($a=5$), our model demonstrates superior generalization across all cross-dataset evaluations. Notably, models trained with the Realistic Blur Synthesis (RSBlur) procedure proposed by \cite{rim_2022_ECCV}  exhibit poor generalization across the benchmark datasets, possibly due to their propensity to learn capture-device-specific transformations and the loss of information caused by the sensor saturations. \cref{fig:cumulative} shows the cumulative ranking of the methods. Notice that the models trained with the proposed procedure occupy the top six positions in the ranking. \cref{fig:srn_results} illustrates how the SRN network benefits from training with our proposed dataset compared to other training sets. More examples are provided in the Supplementary Material.

Training with non-uniformly blurred datasets consistently improves the results on the GoPro dataset compared to training with uniformly blurred images. This can be attributed to multiple moving objects in the GoPro datasets. Training with uniformly blurred images is more effective for smoothly varying blur, such as the K\"{o}hler and RealBlur datasets. 

\subsubsection{Augmenting for the CRF and adding iterations \label{tab:more_iterations}}

We generated some dataset instances as the concatenation of datasets. For example, SBDD\_U~($\gamma=1.0$, $\gamma=2.2$, $a=5$) is the union of SBDD\_U~($\gamma=1.0$), SBDD\_U~($\gamma=2.2$), and SBDD\_U~($a=5$), therefore it has the triplet of images. The number of iterations used to train with the dataset instance is denoted with a suffix. The results are summarized in \cref{tab:res_srn_concat}. 

It is observed that increasing the number of iterations during training consistently leads to better results across all datasets. This indicates that the network is learning to deblur and is not just memorizing the specific characteristics of a particular dataset. Creating a dataset by combining multiple CRFs is advantageous for the overall generalization performance of the model. However, if a CRF that is very different from the test set is included in the training dataset, it can lead to a decline in the accuracy of the results. For instance, adding SBDD\_NU($a=5$) is not beneficial for the K\"{o}hler test set.

\begin{table*}[ht]
    \caption{Compared with results presented in \cref{tab:res_srn}, training with more iterations improves overall performance. Using a training set with several CRFs is also beneficial for overall performance but may degrade the performance on some test sets.}
    \label{tab:res_srn_concat}
    \small 
    \centering
    \begin{tabular}{l|cccc}
             \multicolumn{1}{c}{ \textbf{}} & \multicolumn{4}{|c}{ Test Sets} \\
      \toprule
       \hspace{50pt}    Training Set     &   RealBlur  &   K\"{o}hler & GoPro & GoPro($\gamma=2.2$) \\
      \hline 
      SBDD\_NU ($\gamma=2.2$)-262k  & 30.15/ 0.891  & 27.2/ 0.807  & 28.28/ 0.880 &   29.91/ 0.898     \\
       SBDD\_NU ($\gamma=2.2$)-524k  & 30.26/ 0.894 & 27.73/ 0.814  & 28.59/0.891 &   30.28/ 0.906     \\
        \cline{1-5} 
       SBDD\_NU ($\gamma=1.0$, $\gamma=2.2$)-524k  & 30.45/ 0.902  & 28.78 0.826  & 29.85/ 0.902 &  30.05/ 0.904       \\
       \cline{1-5} 
       SBDD\_U ($\gamma=1.0$,$\gamma=2.2$,$a=5$)-262k  & 30.57/ 0.896 & 28.31/ 0.815  & 29.24/ 0.890 & 29.39/ 0.892   \\
       SBDD\_U ($\gamma=1.0$,$\gamma=2.2$,$a=5$)-786k  & 30.68/ 0.901  & 28.66/ 0.823  &  29.45/ 0.895 & 29.62/ 0.898  \\
       
        \cline{1-5} 
       SBDD\_NU ($\gamma=1.0$, $\gamma=2.2$, $a=5$)-600k  & 30.37/0.890  & 27.91/ 0.814  & 29.24/ 0.889 &  29.57/0.892       \\
       SBDD\_NU ($\gamma=1.0$, $\gamma=2.2$, $a=5$)-786k  & 30.64/ 0.9  & 28.35/ 0.826  & 29.66/0.8989 &  29.86/ 0.9       \\
       \bottomrule 
    \end{tabular}
\end{table*}

\subsection{On the generalization capability induced by the training dataset on other motion deblurring networks}

\begin{table}[ht]
    \caption{Generalization performance to real blurred images of deep deblurring models trained with different datasets. Since K\"ohler uses a linear response function we generate a version of our dataset with $\gamma=1$ that highlights the importance of $\gamma$-correction. \textbf{The best cross-dataset performance is indicated in bold.} }
    \centering
    \footnotesize   
    \setlength{\tabcolsep}{1pt}
    \begin{tabular}{l|cc}
    \toprule 
     & \multicolumn{2}{c}{Test Sets} \\
          Model + Training Set  &  \multicolumn{2}{c}{RealBlur}  \hspace{3em}  K\"{o}hler \\
          \midrule 
       DeepDeblur + GoPro ($\gamma$) & 28.06/0.855 &  25.28/0.743  \\   
       DeepDeblur + REDS  & 27.96/0.860  & 26.03/0.763  \\   
       DeepDeblur + NU($\gamma=2.2$)  &  \textbf{29.0/0.866} &  \textbf{26.91/0.792} \\
       \midrule 
       MIMO\_UNet+ + GoPro  & 27.64/0.836 & 25.05/0.746  \\     
       MIMO\_UNet+  + NU ($\gamma=1.0$)  &  29.13/\textbf{0.877}   &  \textbf{28.54/0.828} \\
       MIMO\_UNet+  + NU ($\gamma=2.2$)  &  \textbf{29.14}/0.872    & 27.13/0.811  \\
       MIMO\_UNet+ + NU ($a=5$)  & 28.79/0.868    &  27.2/0.816 \\
       \midrule 
       NAFNet + GoPro ($\gamma$) & 28.32/0.857 & 26.36/0.767   \\   
       NAFNet + REDS  & 29.40/0.882  & 26.38/0.775   \\ 
       NAFNet + NU ($\gamma=1.0$)  &   29.42/0.885    & \textbf{29.53/0.849}     \\
       NAFNet + NU ($\gamma=2.2$)  &   \textbf{29.66}/0.884    & 27.95/0.831     \\
       NAFNet + NU ($a=5$)  &   29.38/\textbf{0.888}    & 28.85/0.845     \\
       \bottomrule 
    \end{tabular}
    \label{tab:tab:res_srn_real}
\end{table}

To assess how the synthesis procedure generalizes to other architectures, we considered the classical DeepDeblur network \citep{nah2017deep}, the more recent MIMO-UNet+ \citep{cho2021rethinking}, and NAFNet \citep{chen2022simple}. We compared the models provided by the authors with models trained with the instances of our synthesis procedure on two sets of real images with available ground truth: RealBlur \citep{rim_2020_ECCV}, and K\"{o}hler \citep{kohler2012recording}. Quantitative results are presented in~\cref{tab:tab:res_srn_real}. For all the networks, training with the proposed datasets yields better results than the provided models. 

\begin{figure*}[t!]
\centering
\small 
\setlength{\tabcolsep}{2pt}

  \begin{tabular}{cccc}
  Blurry & NAFNet (GoPro) & NAFNet(REDS) & NAFNet(SBDD\_NU($\gamma=2.2$)) \\
      \includegraphics[trim=100 50 50 300,                    clip,width=0.24\textwidth]{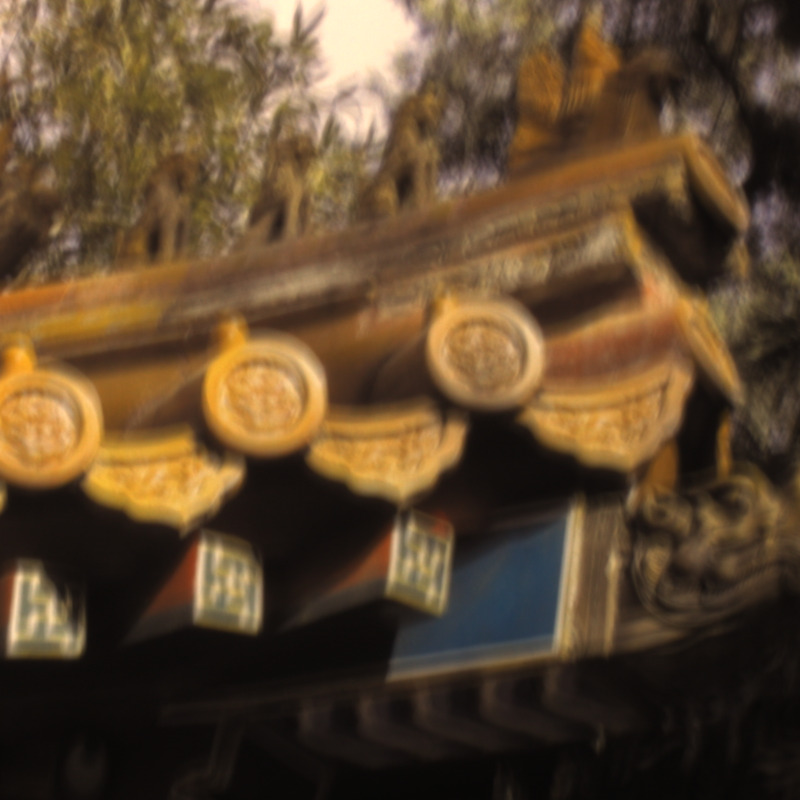} &
    \includegraphics[trim=100 50 50 300 ,                   clip,width=0.24\textwidth]{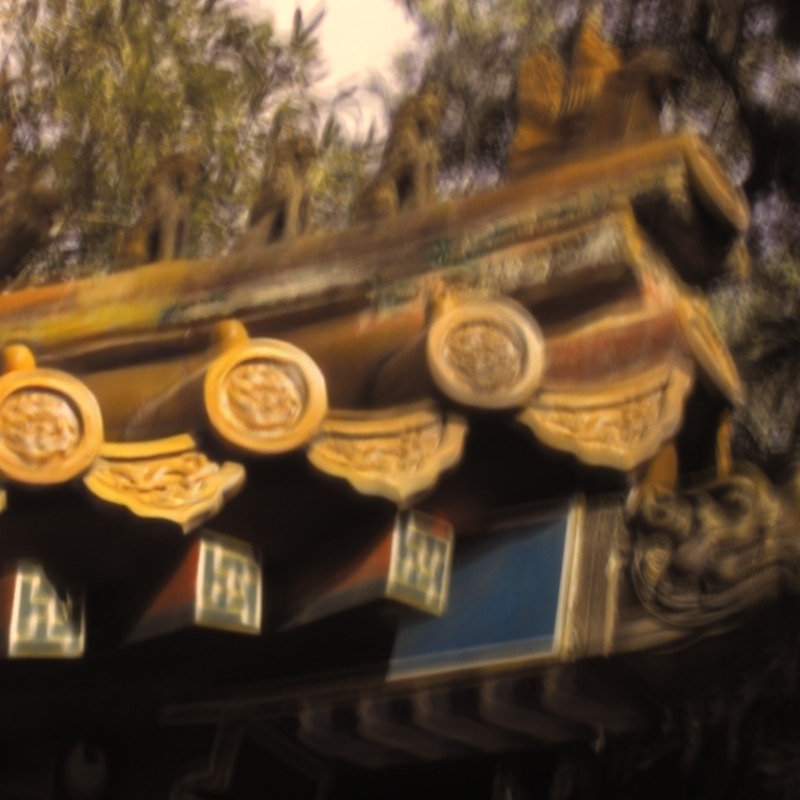} &
    \includegraphics[trim=100 50 50 300 ,                   clip,width=0.24\textwidth]{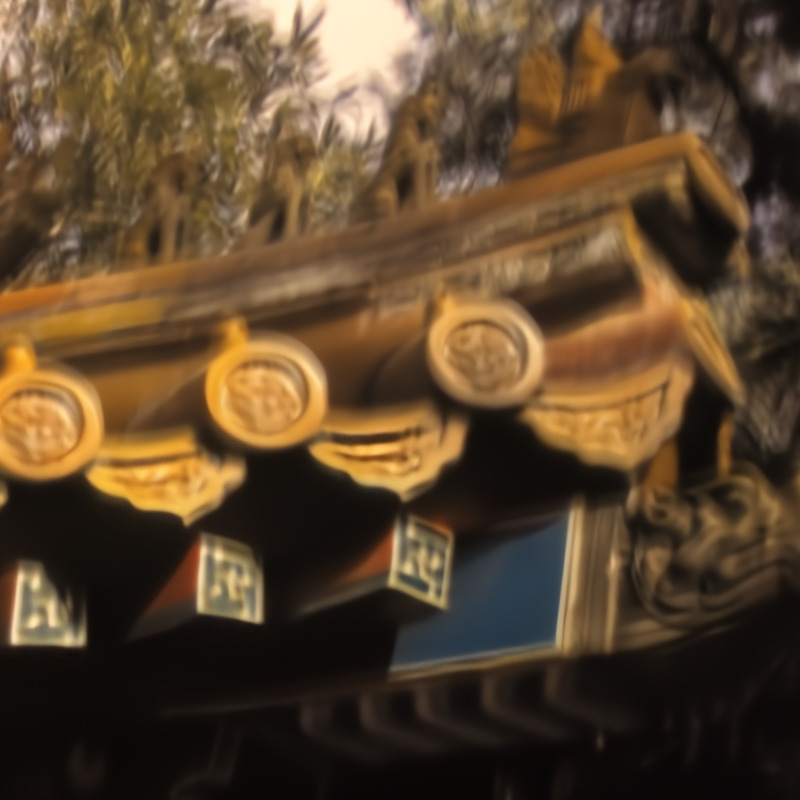}   &
    \includegraphics[trim=100 50 50 300  ,                  clip,width=0.24\textwidth]{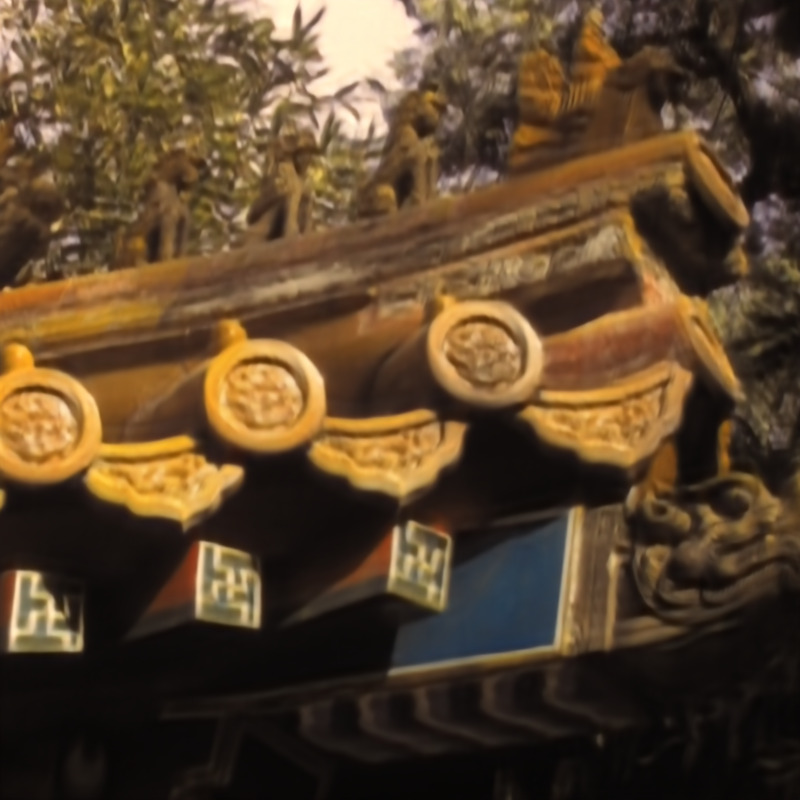} \\
    \includegraphics[trim=0 20 0 110,                    clip,width=0.24\textwidth]{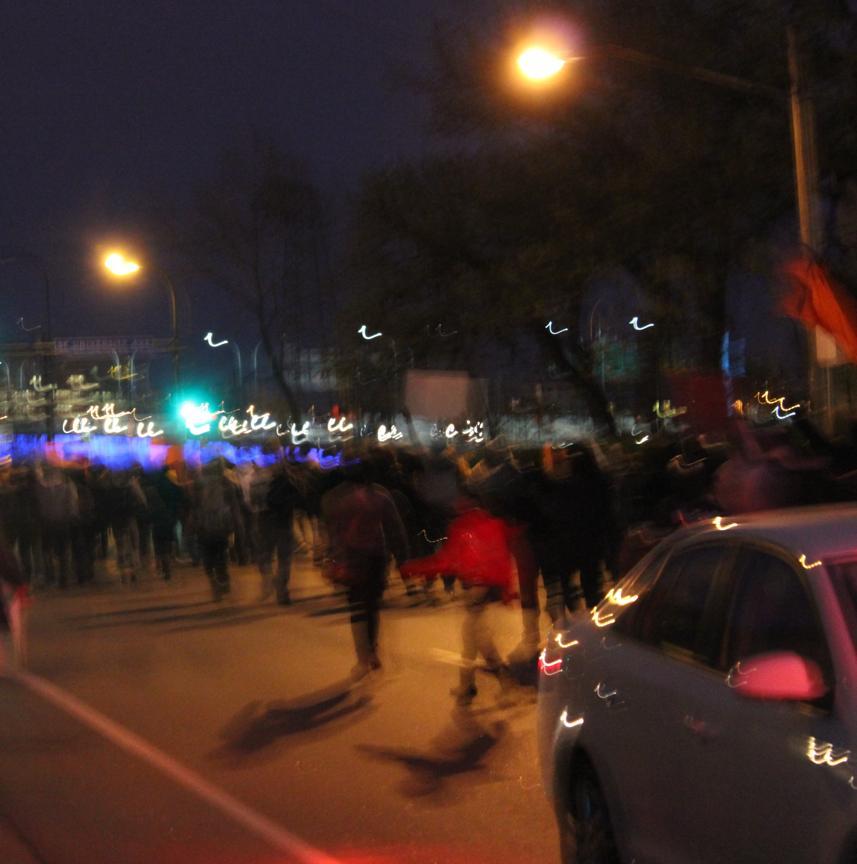} &
    \includegraphics[trim=0 20 0 110 ,                   clip,width=0.24\textwidth]{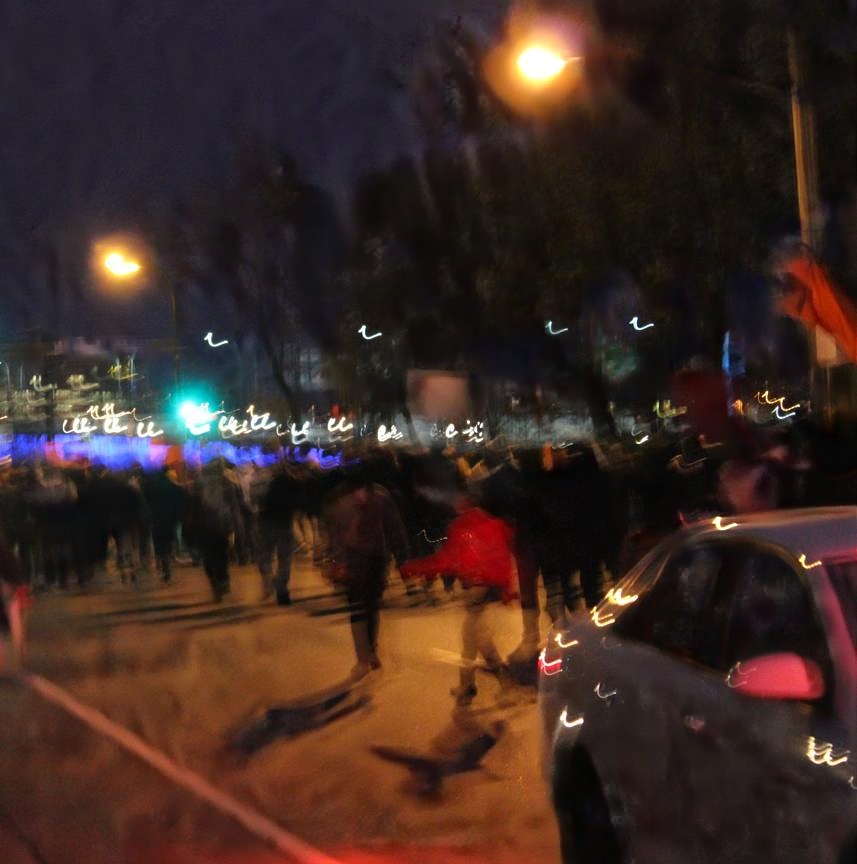} &
    \includegraphics[trim=0 20 0 110 ,                   clip,width=0.24\textwidth]{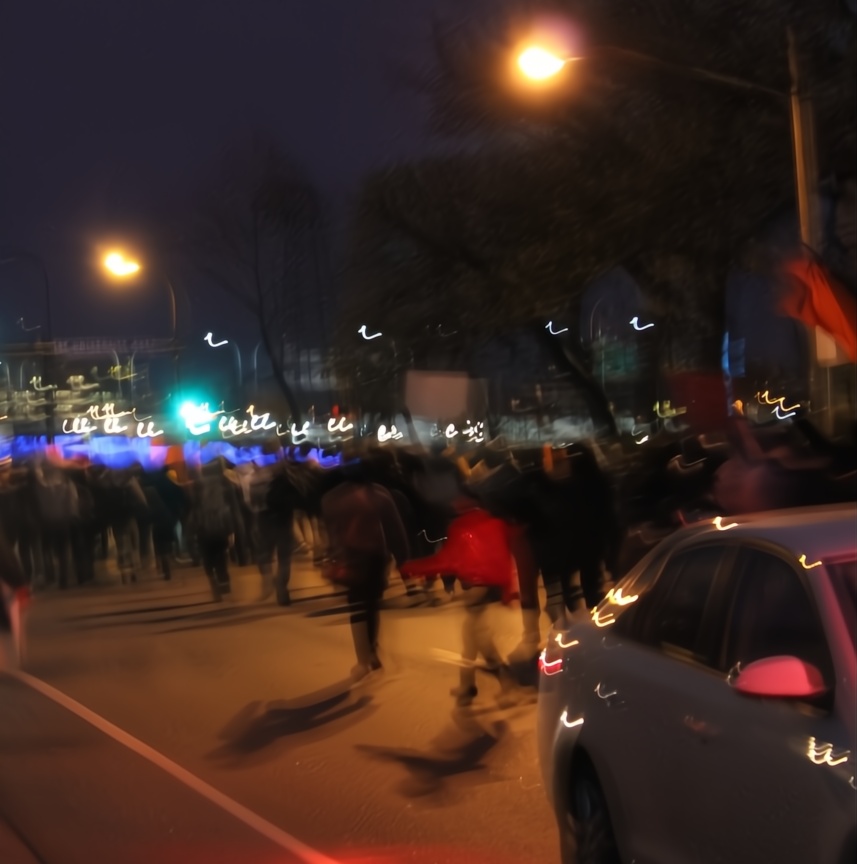}   &
    \includegraphics[trim=0 20 0 110  ,                  clip,width=0.24\textwidth]{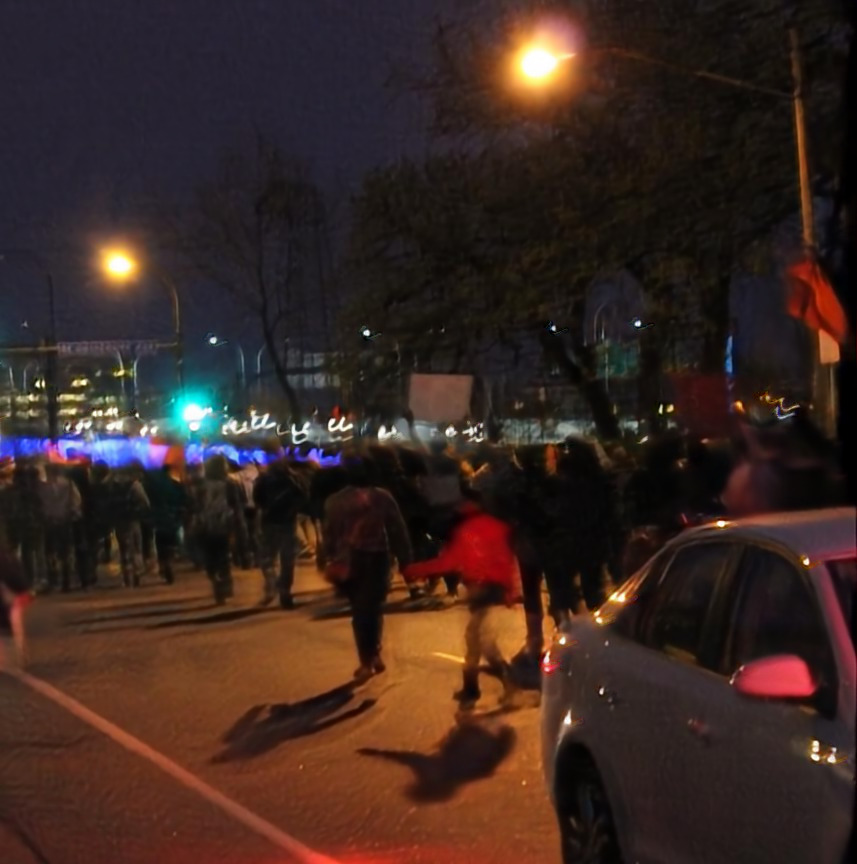} \\
    \includegraphics[trim=0 0 0 0,                    clip,width=0.24\textwidth]{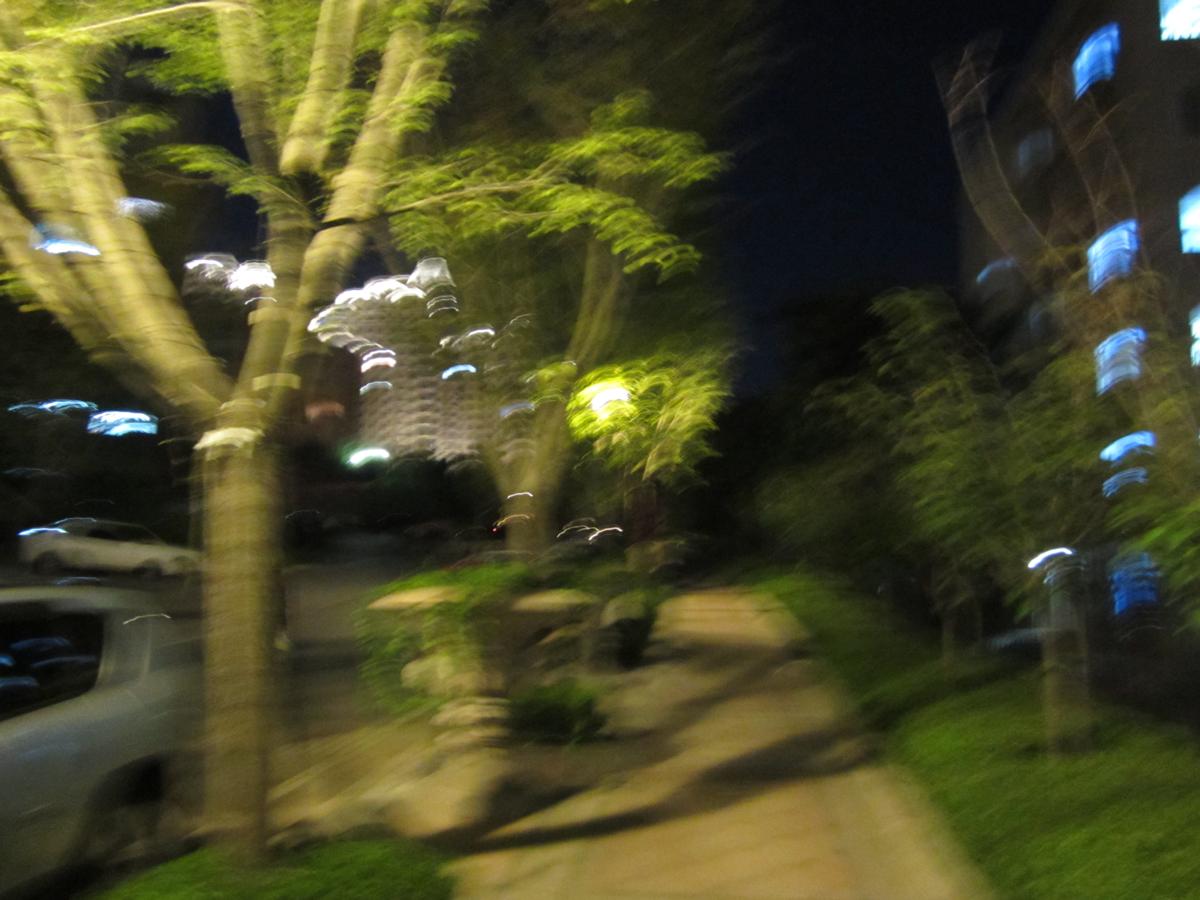} &
    \includegraphics[trim=0 0 0 0 ,                   clip,width=0.24\textwidth]{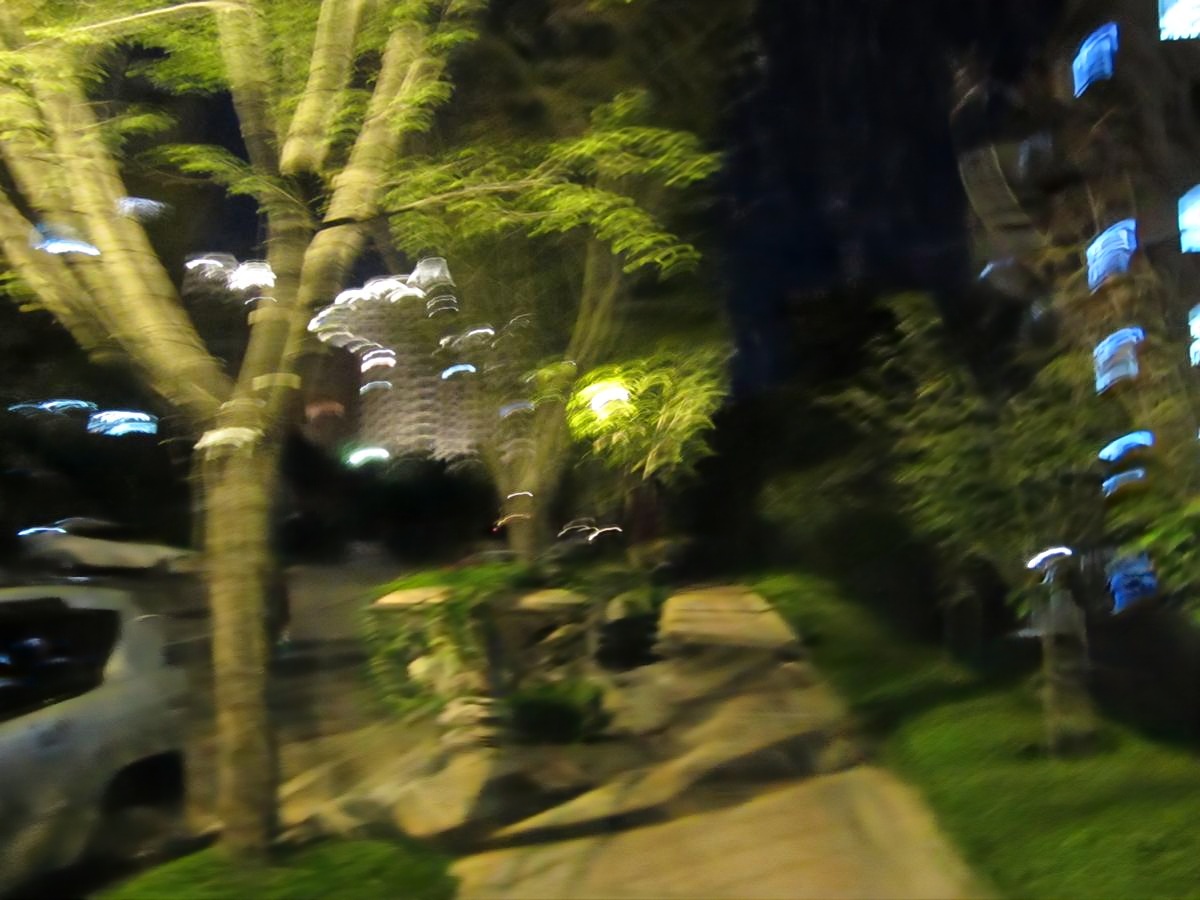} &
    \includegraphics[trim=0 0 0 0 ,                   clip,width=0.24\textwidth]{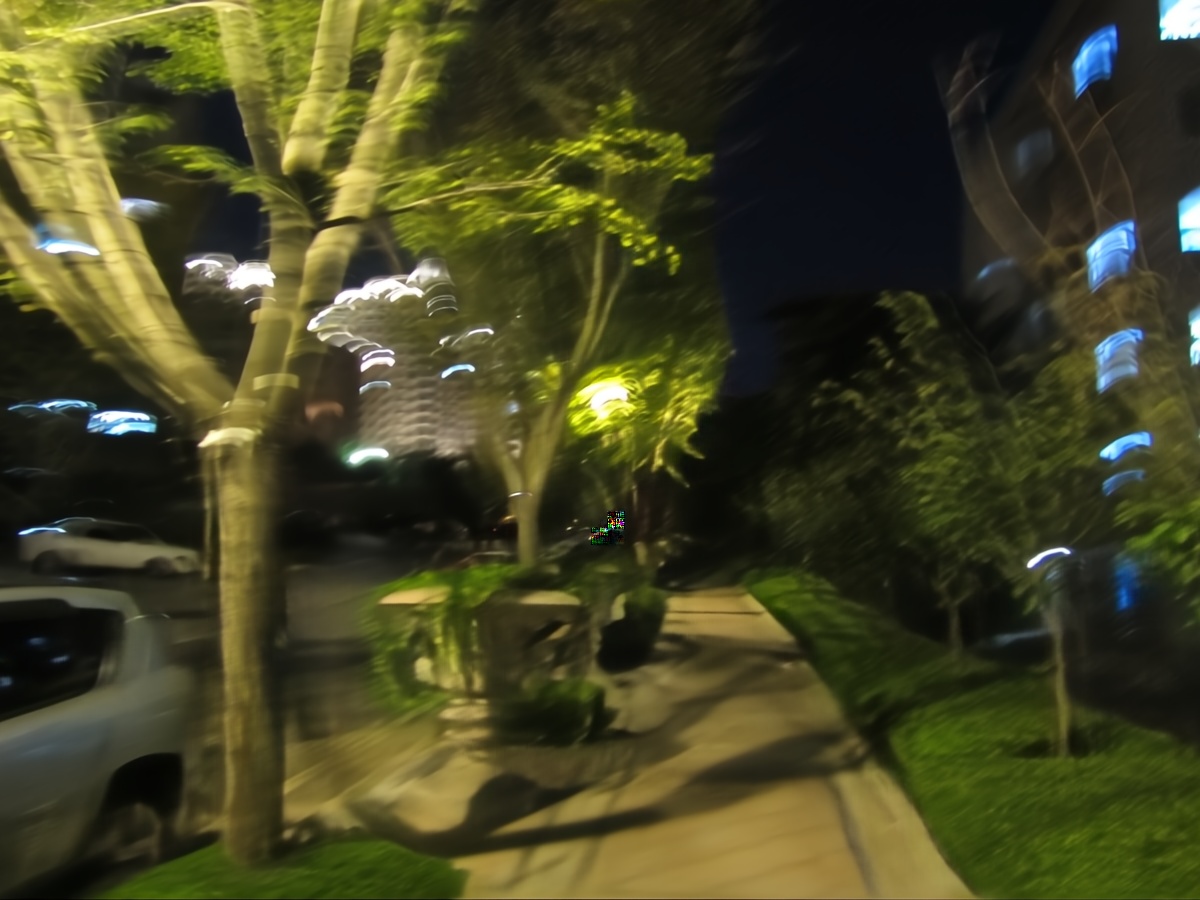}   &
    \includegraphics[trim=0 0 0 0  ,                  clip,width=0.24\textwidth]{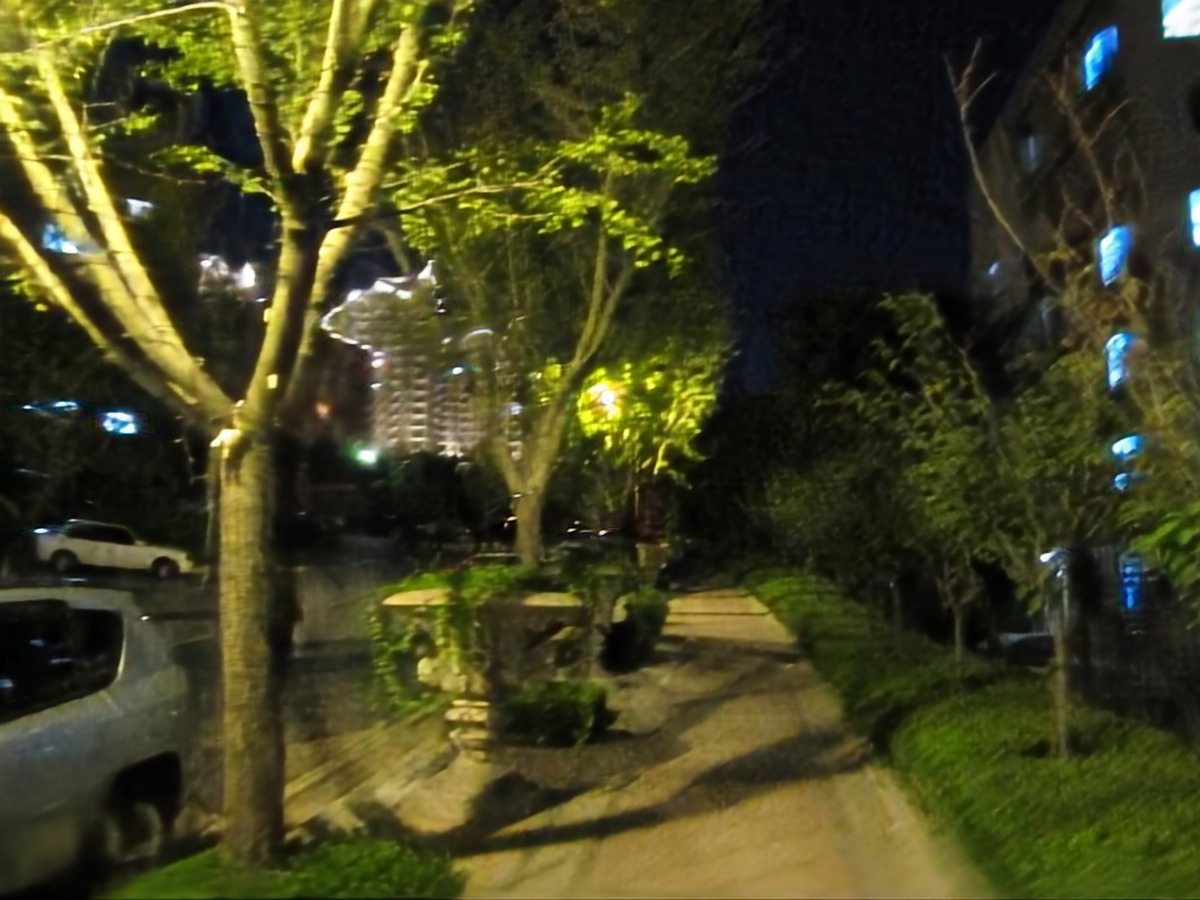} \\

    \includegraphics[trim=0 0 0 0,                    clip,width=0.24\textwidth]{imgs/Blurry/coke.jpg} &
    \includegraphics[trim=0 0 0 0 ,                   clip,width=0.24\textwidth]{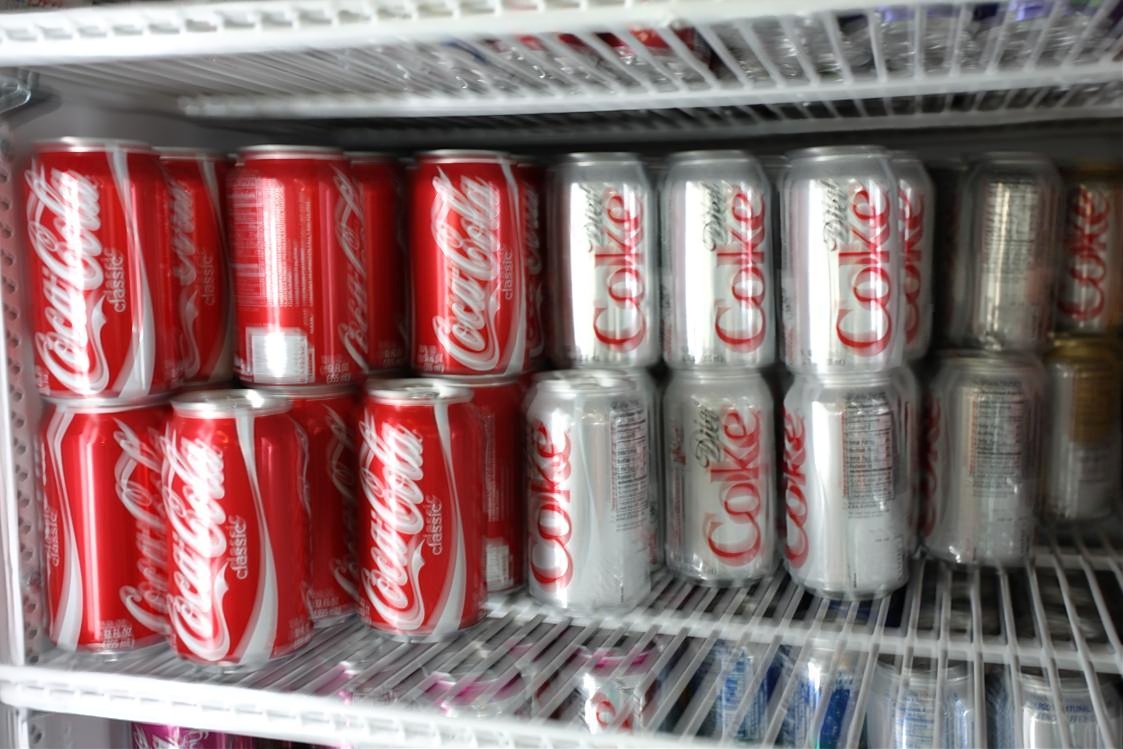} &
    \includegraphics[trim=0 0 0 0 ,                   clip,width=0.24\textwidth]{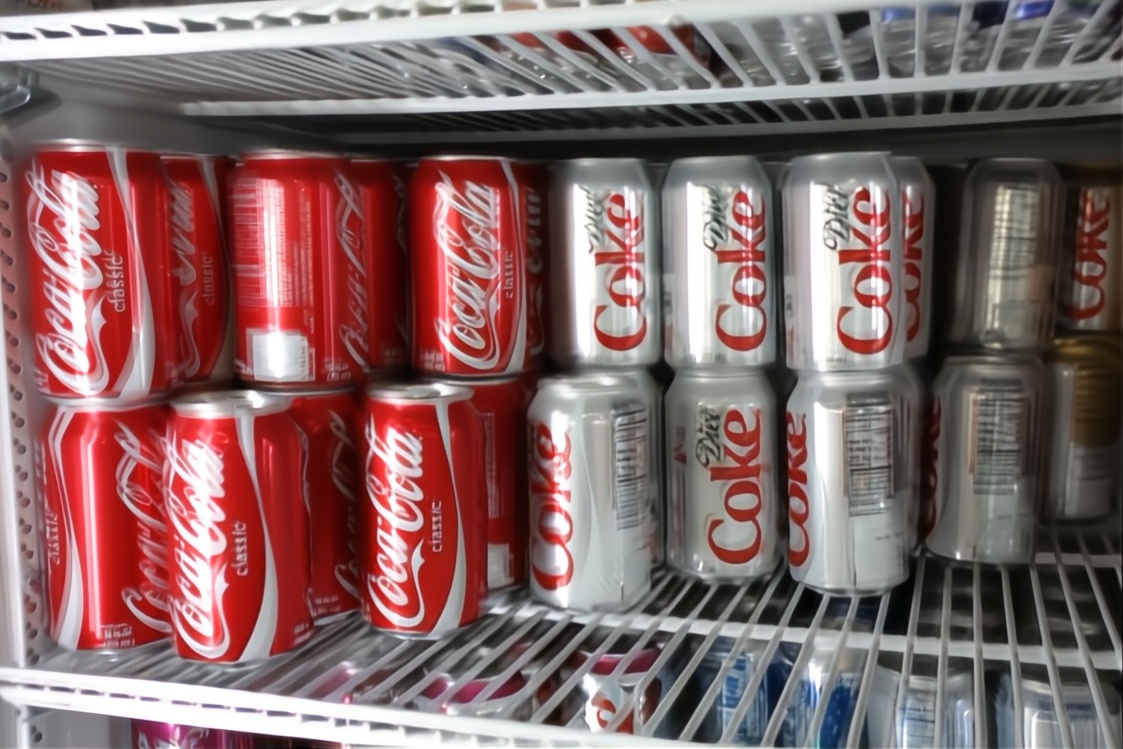}   &
    \includegraphics[trim=0 0 0 0  ,                  clip,width=0.24\textwidth]{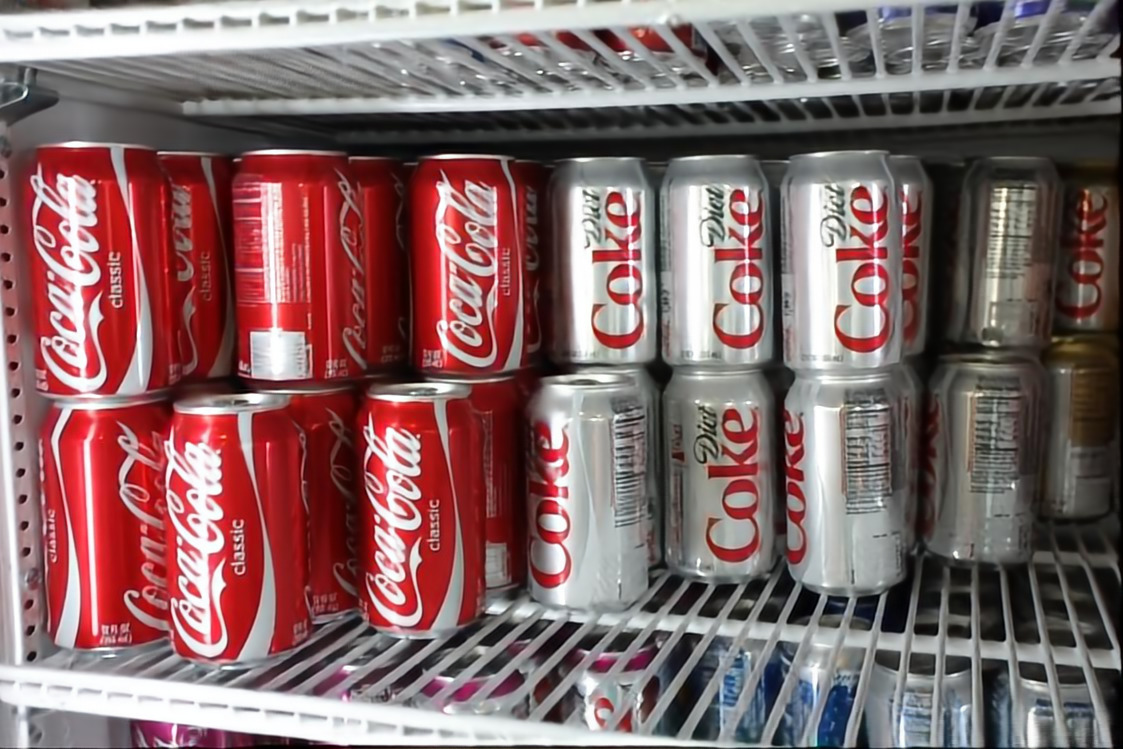} \\

    \includegraphics[trim=0 0 0 50,                    clip,width=0.24\textwidth]{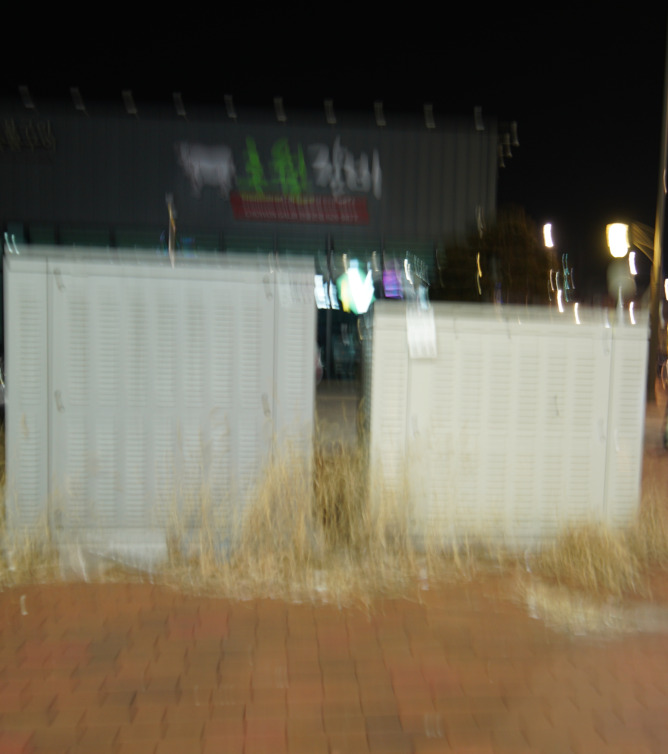} &
    \includegraphics[trim=0 0 0 50 ,                   clip,width=0.24\textwidth]{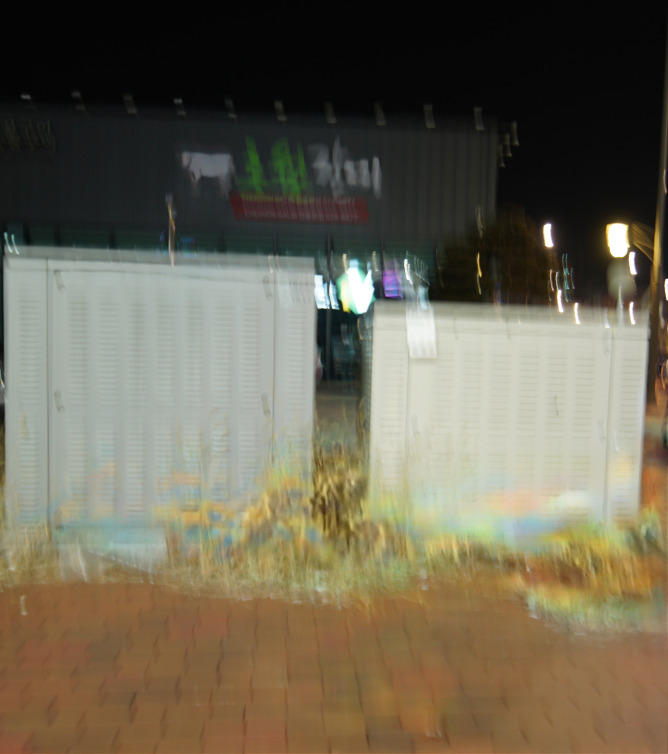} &
    \includegraphics[trim=0 0 0 50 ,                   clip,width=0.24\textwidth]{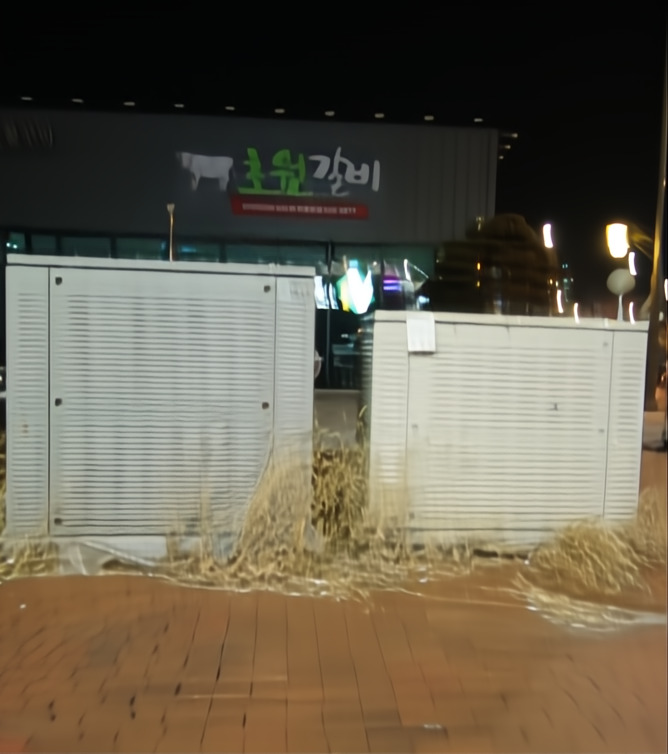}   &
    \includegraphics[trim=0 0 0 50  ,                  clip,width=0.24\textwidth]{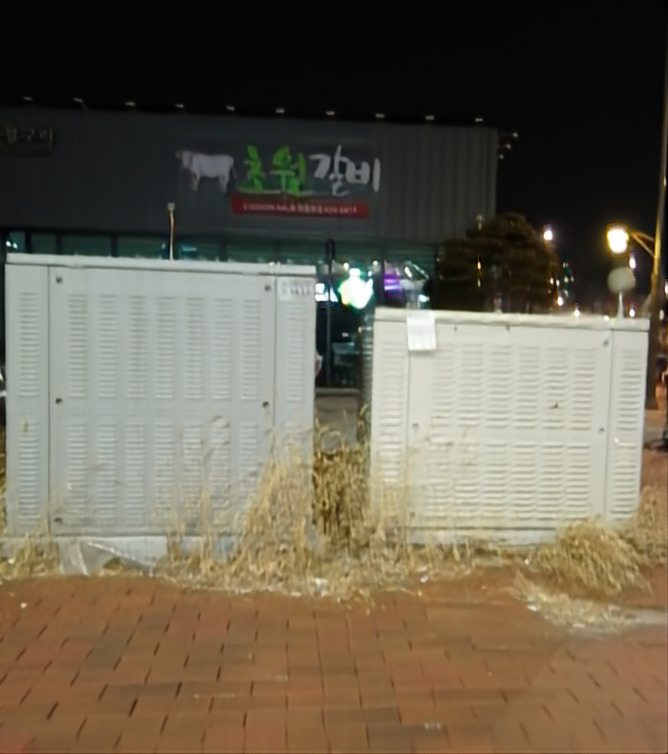} \\
    
  \end{tabular}
  \caption{ Restoration of real blurry images using the NAFNet architecture \citep{chen2022simple} trained on different training sets. \textbf{1$^{st}$ row:} image from \cite{kohler2012recording} dataset. \textbf{2$^{nd}$ to 4$^{th}$ row:} image from \cite{lai2016comparative}. \textbf{5$^{th}$ row:} image from the RealBlur dataset \citep{rim_2020_ECCV} \label{fig:NAFNetResults}. Training with SBDD yields better results on images taken in low-light conditions with saturated pixels despite not having seen this type of image during training. Best viewed in electronic format.} 
\end{figure*}

Deblurring examples on real blurred images for the NAFNet architecture are shown in \cref{fig:NAFNetResults}. Further qualitative comparisons of the results for the different architectures can be found in the Supplementary Material.

\section{Conclusions}

Deep deblurring networks show an impressive capacity to restore image information after heavy degradation from motion blur on benchmark datasets. However, only a limited number of works have explored their cross-dataset performance, or how well they generalize to real motion-blurred images. 

Following extensive experimentation, we first confirm that modern deblurring networks tend to overfit the particularities of training datasets, and their in-distribution performance is not indicative of real deblurring ability in out-of-distribution blurred images. Furthermore, this experimental analysis allows us to identify the factors that most limit the generalization of motion deblurring methods to real motion-blurred images. 

Secondly, building on the previous analysis, we propose a simple methodology for generating training pairs, by simulating motion blur (either uniform or non-uniform) under different conditions (saturated or unsaturated scenes). This way, an arbitrarily large training set can be generated, allowing a significant increase in the generalization performance of existing deblurring networks, particularly on real motion-blurred photographs. 

Interestingly, for the same number of training epochs, a straightforward convolution synthesis method, when combined with appropriate data augmentation, outperforms the more complex setups in generalization. We observe that a simple uniform blur gives the best results in real test datasets where the blur varies slowly. In contrast, non-uniform blur generation achieves the best performance in datasets containing dynamic scenes or significant depth disparity.

\backmatter

\bmhead{Supplementary information}

This article has Supplementary Material.

\bmhead{Acknowledgements}

This work was partially supported by Agencia Nacional de Investigación e Innovación (ANII, Uruguay) grant POS\_FCE\_2018\_1\_1007783. The experiments presented in this paper were carried out using ClusterUY (https://cluster.uy).

\section*{CRediT authorship contribution statement}

\textbf{G. Carbajal:} Conceptualization, Methodology, Investigation, Writing –original draft \& editing, Software. \textbf{P. Vitoria:} Conceptualization, Methodology, Writing - review \& editing. \textbf{P. Mus\'e:} Conceptualization, Methodology, Writing
– original draft and editing, Funding acquisition.
\textbf{J. Lezama:} Conceptualization, Methodology, Writing - review \& editing.

\section*{Declaration of competing interest}
The authors declare that they have no known competing financial interests or personal relationships that could have appeared to
influence the work reported in this paper.

\FloatBarrier

\bibliography{refs}%

\end{document}